\documentclass[journal]{IEEEtran}

\usepackage{cite}

\usepackage{float}
\usepackage{graphicx}
\usepackage{subfig}
\usepackage{overpic}

\usepackage{amsthm,amsmath,amssymb}
\usepackage{bm}

\usepackage{algorithmic}
\usepackage[ruled,vlined,linesnumbered]{algorithm2e}

\usepackage{array}
\usepackage{booktabs}
\usepackage{enumerate}

\usepackage{url}
\usepackage{color}
\usepackage{colortbl}
\usepackage{booktabs}
\definecolor{tabcolor}{rgb}{0,0,1}

\begin{document}
\title{Design and Control of the ``TransBoat": A Transformable Unmanned Surface Vehicle for Overwater Construction}
\author{Lianxin~Zhang, Xiaoqiang~Ji,
		Yang~Jiao, Yihan~Huang,
        and~Huihuan~Qian % <-this % stops a space
\thanks{This paper is partially supported by Projects U1813217 and U1613226 from NSFC, Project AC01202101027 from the Shenzhen Institute of Artificial Intelligence and Robotics for Society (AIRS), Project KQJSCX20180330165912672 from the Science, Technology and Innovation Commission of Shenzhen Municipality, and PF.01.000143 from The Chinese University of Hong Kong, Shenzhen, China.}
\thanks{Lianxin Zhang, Xiaoqiang Ji, and Huihuan Qian are with Shenzhen Institute of Artificial Intelligence and Robotics for Society (AIRS), The Chinese University of Hong Kong, Shenzhen (CUHK-Shenzhen), Shenzhen, Guangdong, 518129, China. }
\thanks{Lianxin Zhang, Yang Jiao, Yihan Huang, and Huihuan Qian are also with School of Science and Engineering (SSE), The Chinese University of Hong Kong, Shenzhen (CUHK-Shenzhen), Shenzhen, Guangdong 518172, China.}
\thanks{\emph{*Corresponding author is Huihuan Qian (e-mail: hhqian@cuhk.edu.cn).}}
}

\markboth{Transactions on Mechatronics} %,~Vol.~14, No.~8, August~2015}%
{Shell \MakeLowercase{\textit{et al.}}: Bare Demo of IEEEtran.cls for IEEE Journals}

\maketitle

\begin{abstract}
This paper presents the TransBoat, a novel omnidirectional unmanned surface vehicle (USV) with a magnet-based docking system for overwater construction with wave disturbances. 
This is the first such USV that can build overwater structures by transporting modules.
The TransBoat incorporates two features designed to reject wave disturbances.
First, the TransBoat’s expandable body structure can actively transform from a mono-hull into a multi-hull for stabilization in turbulent environments by extending its four outrigger hulls.
Second, a real-time nonlinear model predictive control (NMPC) scheme is proposed for all shapes of the TransBoat to enhance its maneuverability and resist disturbance to its movement, based on a nonlinear dynamic model.  
An experimental approach is proposed to identify the parameters of the dynamic model, and a subsequent trajectory tracking test validates the dynamics, NMPC controller and system mobility.
Further, docking experiments identify improved performance in the expanded form of the TransBoat compared with the contracted form, including an increased success rate (of \bm{$\sim 10\%$}) and reduced docking time (of \bm{$\sim 40$} s on average). 
Finally, a bridge construction test verifies our system design and the NMPC control method.

\end{abstract}


% \fi
\IEEEpeerreviewmaketitle

\section{Introduction}

\begin{figure} [tbp] 
	\centering
	\includegraphics[width=1\linewidth]{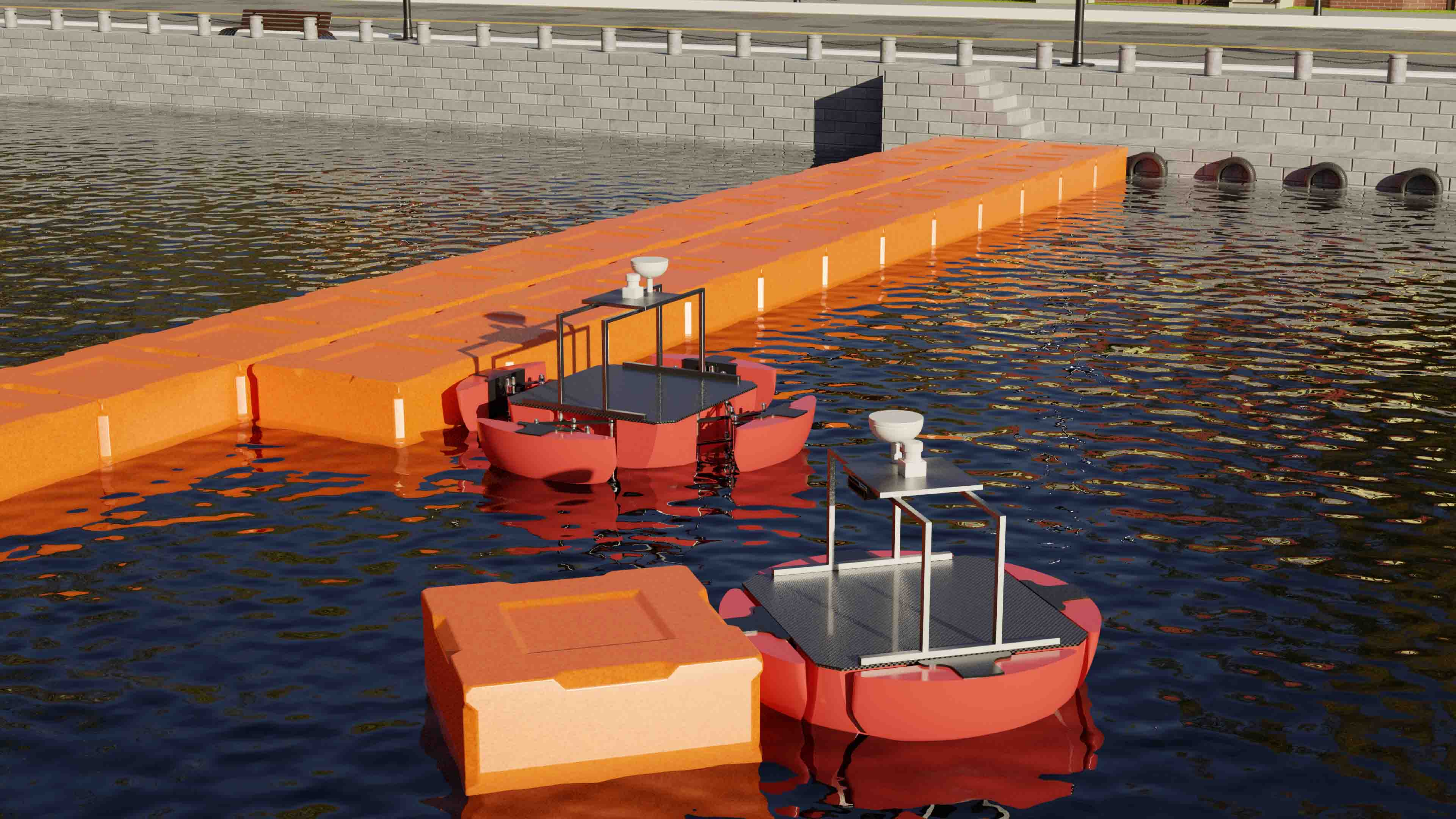}
	\caption{A fleet of TransBoats can build a temporary bridge using prefabricated building blocks. }
	\label{fig:design_idea}
\end{figure}

\IEEEPARstart{O}{verwater} construction denotes the building of stationary infrastructure or dynamic floating structures on a water surface, which provides a critical solution to the urban land shortage problem induced by the population explosion and global sea-level rise \cite{remmers1998mobile, wang2015large}.
Recently, the United Nations has initiated efforts to support floating cities \cite{UN-habitat}.
With the incorporation of robots to assist or completely replace human labor,
autonomous overwater construction has attracted growing attention in recent years due to its high construction efficiency, low accident rate, and ability to perform complicated tasks \cite{o2014self, paulos2015automated, kayacan2019learning}. 
Promisingly, unmanned surface vehicles (USVs) are set to play a significant role in future overwater construction, such as building rescue platforms during floods, forming floating aprons for aerial vehicle landings \cite{seo2013assembly}, and constructing on-demand bridges or parking lots in coastal/riverside cities \cite{wang2018design, zhang2021efficient}, as depicted in Fig. \ref{fig:design_idea}.

Inspired by collective intelligence in nature such as mound-building termites \cite{werfel2014designing}, our main interest in this work is autonomous overwater construction by the Assembly-based Construction Robot System (ACRoS)
which can assemble prefabricated substructures. 
Compared with other technologies such as additive manufacturing and human-robot collaboration, the benefits of ACRoS to robot-assisted building processes include greater construction speed, efficiency, and error tolerance \cite{werfel2011distributed, bier2018robotic}.
Based on substructure type, the ACRoS can be classified into two categories: the ACRoS-R systems with modular robots and the ACRoS-M systems with passive building modules/blocks/bricks (hereinafter referred to as modules).
In the ACRoS-R systems, a fleet of modular robots (aerial/ground/aquatic) automatically assemble to form the desired structure, which is called self-assembly \cite{saldana2018modquad}. During construction, each robot can independently move to its target with collision avoidance and maintain its position under environmental disturbances. This system has been proved to be effective and efficient on a water surface \cite{o2014self, paulos2015automated, kayacan2019learning}. 
However, it is highly expensive and complicated to control for large-scale construction, which limits its application scope.
In the ACRoS-M systems, unactuated modules, rather than robots, are transported to build a predetermined structure \cite{werfel2008three}. 
Robots with grippers or docking systems serve to pick up and deliver the modules, and then assemble them \cite{lindsey2012construction, werfel2014designing, wagner2021smac}.
Researchers have proposed some generalized planning algorithms for the ACRoS-Ms in two dimensions \cite{werfel2007collective, bignell2019open}, which can be applied on water surfaces. 
However, there has been no regard for the effect of wave disturbances on robot movements.
Moreover, no hardware research and validation of the ACRoS-Ms for overwater construction has been documented.

Position drift and attitude instability caused by unpredictable wave disturbances are major hurdles for robotic overwater construction. 
Wave disturbances can be resisted by two design features, namely, a more stable structure and a robust motion controller.
One effective way to improve stability on waves is by widening the hull; however, this reduces mobility, as the aquatic friction increases.
This challenge can be overcome through a transformable design, in which the USV widens its hull when picking up/attaching the modules and then contracts it during their delivery.
In terms of the motion control, a number of approaches have been studied, such as the PID control, the adaptive control \cite{shin2017adaptive}, and the sliding mode method \cite{wang2021coordinated}. 
As the model predictive control (MPC) method has proved to be efficient and robust with minimal tuning of the controller gains \cite{wang2020neurodynamics}, given the nonlinearity of the USV model, a real-time nonlinear model predictive control (NMPC) method can be applied here.

On these grounds, we present the TransBoat, a novel omnidirectional USV with an expandable structure, equipped with an instant docking system based on a switchable magnet \cite{tavakoli2015switchable} that can conduct overwater construction in the ACRoS-M systems through the automatic assembly of the building modules.
The TransBoat is designed as a multi-hull USV of symmetrical structure.
Its shape is approximately spherical, composed of a main hull with four outrigger hulls. 
The full actuation and compact structure endow the TransBoat with maneuverability in dense construction sites.
For the first step, we develop a scale (approximately 1:2) prototype of the TransBoat to facilitate the examination of its construction autonomy and wave resistance.

The main contributions of this work are as follows:
\begin{enumerate} [1)]
	\item A novel transformable omnidirectional USV, the TransBoat, is proposed for overwater construction under wave disturbances. Concurrently, an automated, self-centering, and instant docking system based on switchable magnets is designed to facilitate the assembly process between the TransBoat and the building modules.
	\item We propose a modeling approach for the USV movement with system parameter identification, develop and experimentally validate a real-time NMPC method for all forms of TransBoat used in turbulent environments.
	\item A functional prototype and the validation of overwater construction are implemented through a rapid bridge building experiment, incorporating automated docking, transportation, and assembly on water surfaces.
\end{enumerate}

The remainder of the paper is organized as follows. 
In Section \ref{sect:system}, all components and subsystems of the TransBoat are described in detail. 
Section \ref{sect:control} presents the layout of the dynamical modeling, grey-box identification and NMPC control.
The results of the identification test and validation experiments including tracking, docking, and bridging are illustrated and discussed in Section \ref{sect:experiment}.
Finally, Section \ref{sect:conclusion} concludes the paper.

\begin{figure*} [htpb]
	\centering
	\includegraphics[width=0.9\textwidth]{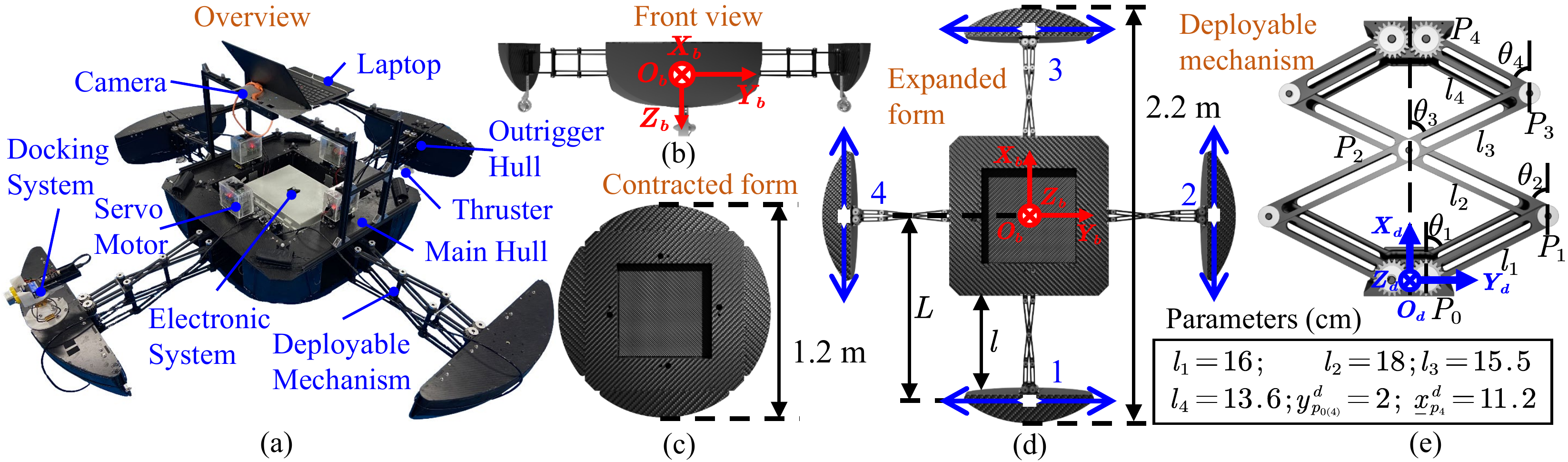}
	\caption{(a) System overview of the TransBoat, (b) front view, (c) contracted form, (d) expanded form, (e) deployable mechanism and design parameters. }
	\label{fig:body}
\end{figure*}

\section{DESIGN OF TRANSBOAT}
\label{sect:system}

Overwater construction under wave disturbances requires robots with great maneuverability and stability, equipped with an automatic docking mechanism, to pick up and release the building modules.
Fig. \ref{fig:body} (a) shows the main components of the TransBoat design, including the mechanical structure, the electronic system, and the docking system, each of which is detailedly described in this section.

\subsection{Mechanical Design}

\subsubsection{Body}

For the fabrication of the TransBoat, ethylene-vinyl acetate (EVA) with a density of $160\ \rm kg/m^3$ is chosen as the major buoyancy material, as it is lightweight, processable, and water-resistant even when damaged.
To enhance the structural strength and stiffness, carbon fiber reinforced polymer (CFRP) plates are cut and assembled into the deck and the deployable mechanisms.

We use an expandable structure to improve the boat's stability in both roll and pitch directions.
As shown in Fig. \ref{fig:body}, the body consists of one main hull and four outrigger hulls connected by the deployable mechanisms. 
The robot length can be increased from $1.2\ \rm m$ to $2.2\ \rm m$.
The consequent stability improvement is shown by experiments in Section \ref{sect:experiment}.
The boat shape is approximately spherical when all hulls are contracted as shown in Fig. \ref{fig:body} (c) and can be changed by mechanism expansion from a mono-hull to a pentamaran.

For control simplicity in the first step, the extension lengths of the four outrigger hulls are kept the same, although independent. 
Due to the symmetrical five-hull design, the lateral and longitudinal hydrodynamics of the TransBoat are identical, which significantly simplifies the control of omnidirectional movements.
As for the overall configuration, a thruster is installed beneath each of the four outrigger hulls, and the docking systems can be fixed on top of each. The electronic system and battery packs are located at the center of the main hull.

\subsubsection{Deployable Mechanism}	
The deployable mechanism is a double-decked scissor-like element (SLE) with excellent shape-transforming performance, e.g., small storage space and large workspace \cite{yang2019deployable}.
Fig. \ref{fig:body} (e) presents the mechanism coordinate system ($O_d \text{-} X_d  Y_d Z_d$), 
where $l_i$ and $\theta_i \ (i=1,2,3,4)$ denote the lengths of the rods and the angles between the rods and the $X_d$-axis, respectively.
We represent the joints with $\mathbf{P}_i = (x^{d}_{P_i}, y^{d}_{P_i}, z^{d}_{P_i})$, and their positions can be calculated as
\begin{equation} \label{eq:joints}
\mathbf{P}_i=
\left[ \begin{array}{c}
\sum_{j=1}^i{l_j\cos \theta _j}\\
\sum_{j=1}^i{l_j\sin \theta _j}+y^{d}_{P_0}\\
0\\
\end{array} \right], 
\end{equation}
where $\theta_1 \ (0 \leq \theta_1 \textless \pi/2)$ is the input rotation actuated by the servo motor.
According to the symmetry and the geometry relations in Fig. \ref{fig:body} (d), the coordinates $y^{d}_{P_i}$ and angles $\theta_i$ satisfy
\begin{equation} \label{eq:relation}
\begin{aligned}
y^{d}_{P_1} &= l_1\sin \theta _1 + y^{d}_{P_0} = - l_2\sin \theta _2,\\
y^{d}_{P_3} &= l_3\sin \theta _3 = - l_4\sin \theta _4 +y^{d}_{P_4}, \\
\theta _2 &= - \theta _3,
\end{aligned}
\end{equation}
where $y^{d}_{P_0}$ and $y^{d}_{P_4}$ are the $ Y_d $-axis values of joint $\mathbf{P}_0$ and $\mathbf{P}_4$, respectively, both of which are design parameters of the mechanism.
From Eq. (\ref{eq:relation}), $\theta_2$ and $\theta_4$ can be represented as functions of  $\theta_1$, that is,
\begin{equation} \label{eq:theta2}
\begin{aligned}
\theta_2 &= \Theta _2\left( \theta_1 \right) = -\text{arc}\sin \left( \frac{l_1}{l_2}\sin \theta_1+\frac{y_{P_0}^{d}}{l_2} \right), \\
\theta_4 &= \Theta _4\left( \theta_1 \right) = \text{arc}\sin \left( -\frac{l_1l_3}{l_2l_4}\sin \theta _1+\frac{y_{P_0}^{d}}{l_2}+\frac{y_{P_4}^{d}}{l_4} \right). 
\end{aligned}
\end{equation}

Next, we substitute Eq. (\ref{eq:theta2}) into Eq. (\ref{eq:joints}), and get
\begin{equation} \label{eq:xp4}
x^{d}_{P_4} = l_1\cos \theta _1+\left( l_2+l_3 \right) \cos \Theta _2\left( \theta_1 \right) + l_4 \cos \Theta_4 \left( \theta_1 \right) .
\end{equation}
Constrained by the geometry, the deployable mechanism cannot be contracted to its limit position.
Let $\underline{x}^{d}_{P_4}$ denote the minimum position to which the mechanism can retract.
Therefore, the expansion length of the deployable mechanisms $l$ controlled by the input $\theta _1$ is
\begin{equation} \label{eq:xE}
l = x^{d}_{P_4} - \underline{x}^{d}_{P_4}.
\end{equation}

Fig. \ref{fig:body} (e) shows the design parameters of the deployable mechanisms, which are driven by a servo motor with a maximum torque of $18 \rm N \cdot m$ in our implementation. 
Therefore, combining Eqs. (\ref{eq:xE}) and (\ref{eq:xp4}), we can expand the mechanism to any desired length by referring to a precomputed table.

\subsubsection{Propulsion Subsystem}
Four thrusters are installed below the outrigger hulls in a ``+" shaped configuration to provide holonomic propulsion, which is more efficient than other configurations such as an ``X" shaped actuator configuration \cite{wang2018design}.
The body coordinate system ($O_b \text{-} X_b  Y_b Z_b$) and thruster layout are shown in Fig. \ref{fig:body} (d), and all thrusters can generate both forward and backward forces. 
Then, using $\mathbf{u} = \left[ f_1 \,\, f_2 \,\, f_3 \,\, f_4 \right]^\mathrm{T}$ to denote the propulsion vector generated by the four thrusters with $f_i \ (i=1,2,3,4)$ representing each propulsion and $L$ denoting the distance from each thruster to the USV body center, the applied force and moment vector  $\mathbf{F}$ in the plane can be computed as
\begin{equation} \label{eq:forces}
\mathbf{F} = \mathbf{Eu}
= \left[ \begin{matrix}
0&		1&		0&		1\\
1&		0&		1&		0\\
-L&	   -L&	    L&	    L\\
\end{matrix} \right] \left[ \begin{array}{c}
f_1\\
f_2\\
f_3\\
f_4\\ 
\end{array} \right], 
\end{equation}
where $L$ is linearly correlated with the expansion length $l$ by $L = l + 0.4435$.

The propulsion forces $f_i$ of each thruster are generated and controlled by inputting pulse-width modulation (PWM) signals $h^{pwm}_i$ with different duty cycles (approximately $6.1\% \sim 8.8\%$) to the regulator. According to this relation, the propulsion model can be written in the form
\begin{equation} \label{eq:propulsion}
f_i = \zeta_i(h^{pwm}_i),
\end{equation}
which is measured by a dynamometer, yielding a propulsion table mapped to all PWM commands.
In each control cycle, the hardware first computes the desired force and then sends the corresponding commands to the propulsion system by referring to this table.

\subsection{Electronics}

\begin{figure} [htbp] 
	\centering
	\includegraphics[width=0.9\linewidth]{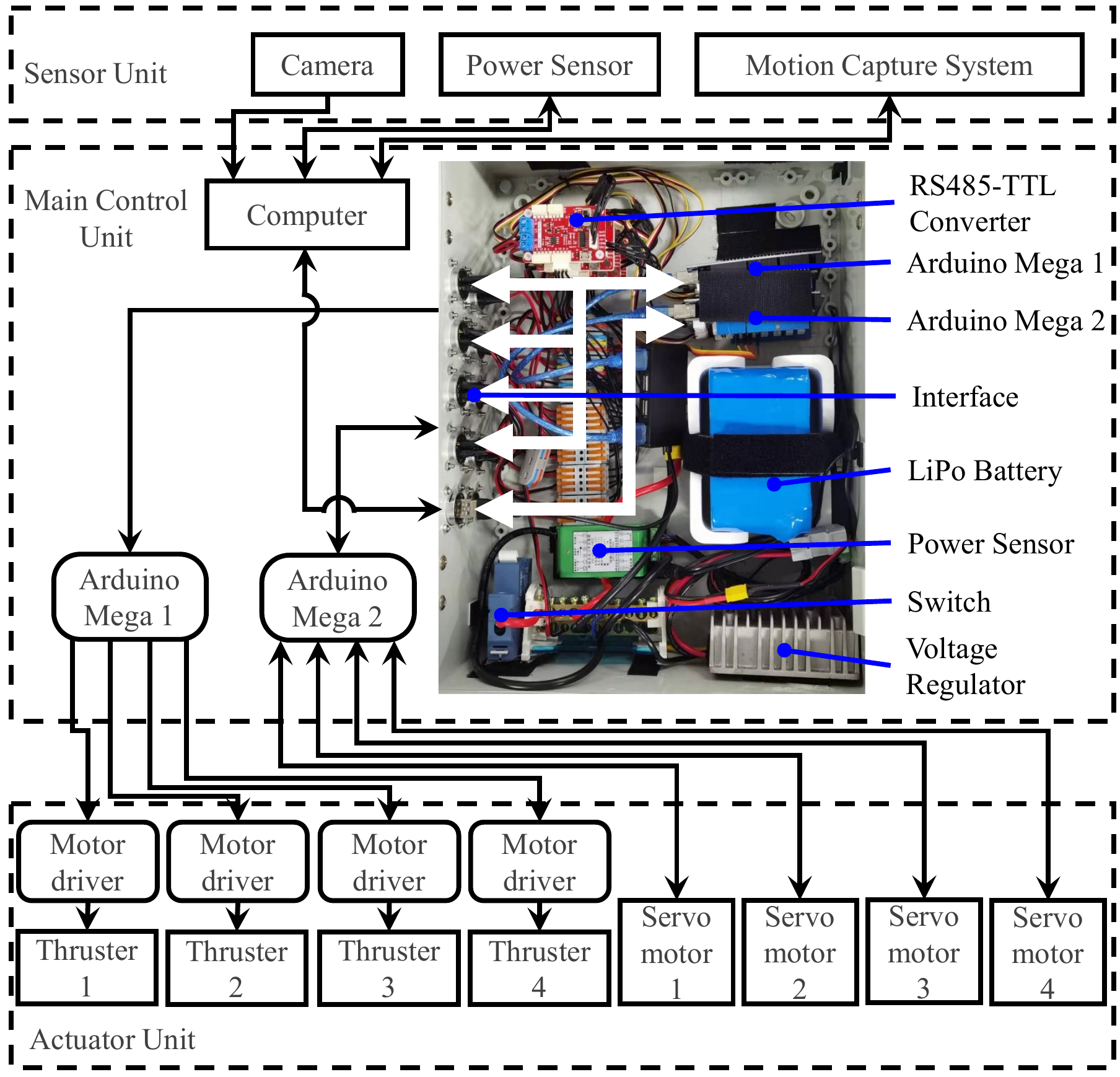}
	\caption{Architecture of the electronic system.}
	\label{fig:electronic_system}
\end{figure}

The architecture of the electronic subsystem is shown in Fig. \ref{fig:electronic_system}. 
It can be seen that the main electronic components are the navigation sensors, the main control unit, and the actuators. 
Several sensors are used for navigation and environmental perception, including an OptiTrack motion capture system, a power sensor, and a camera.
Specifically, the motion capture system is connected with the TransBoat through a local area network (LAN) to provide millimeter-precision positions and the corresponding orientations.
To avoid collision during movements, a binocular camera (MYNT EYE D) is installed on the top of the USV body to detect obstacles or targets.
The processing unit to execute the navigation and control algorithms is a laptop (1.6 GHz Intel Core i5-10210U, 8G LPDDR3). 
Two Arduino Mega 2560 boards, which directly control the four servo motors of the deployable mechanisms and the four thrusters, receive and conduct the control commands from the laptop so that the processing unit can focus on communication and control.

\subsection{Docking Subsystem}

\begin{figure} [htbp] 
	\centering
	\begin{minipage}[b]{.5\linewidth}
		\centering
		\subfloat[]{
		\includegraphics[width=1\linewidth]{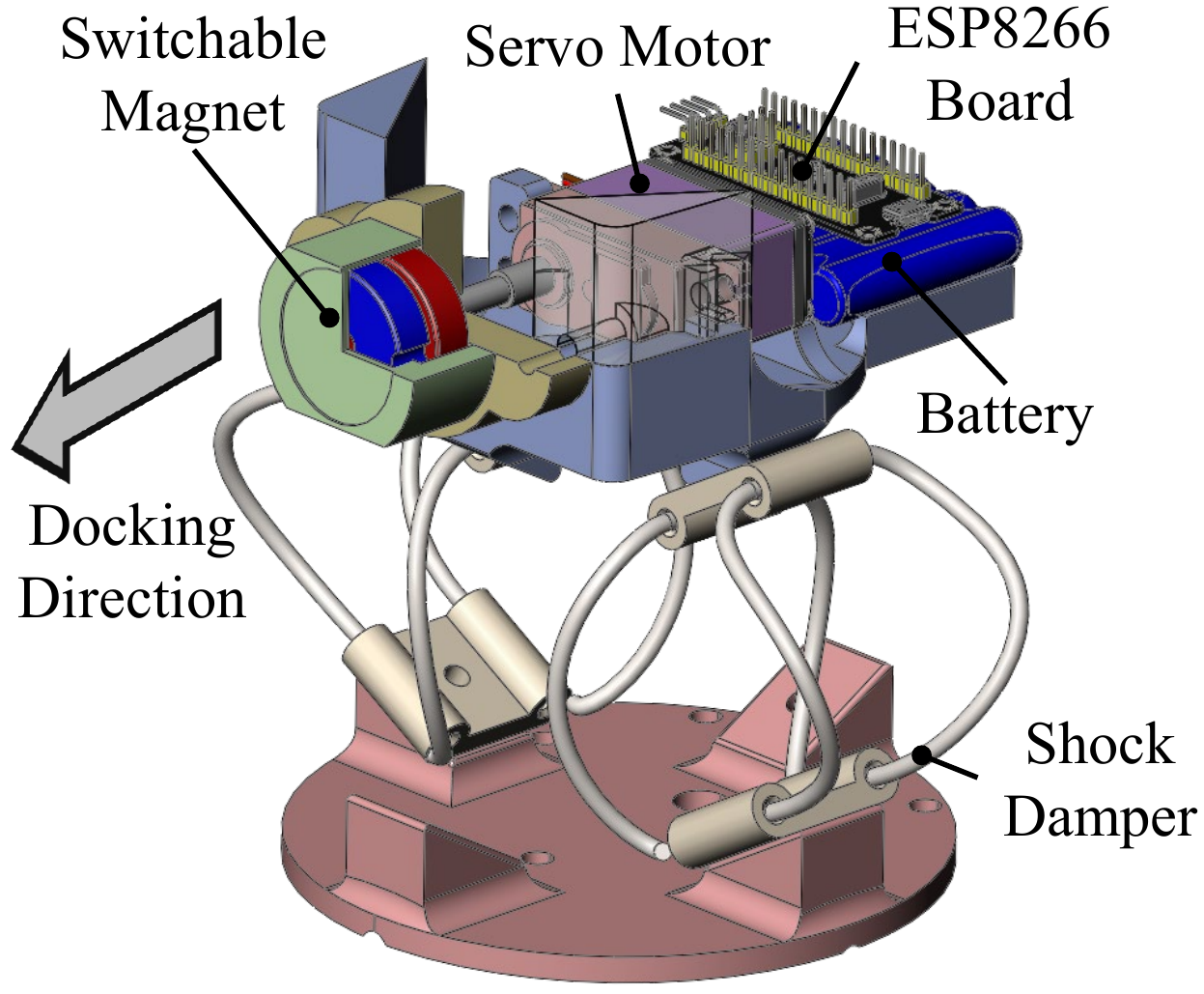}
		}
	\end{minipage}
	\begin{minipage}[b]{.4\linewidth}
		\centering
		\subfloat[]{
			\begin{overpic}[width=1\textwidth]{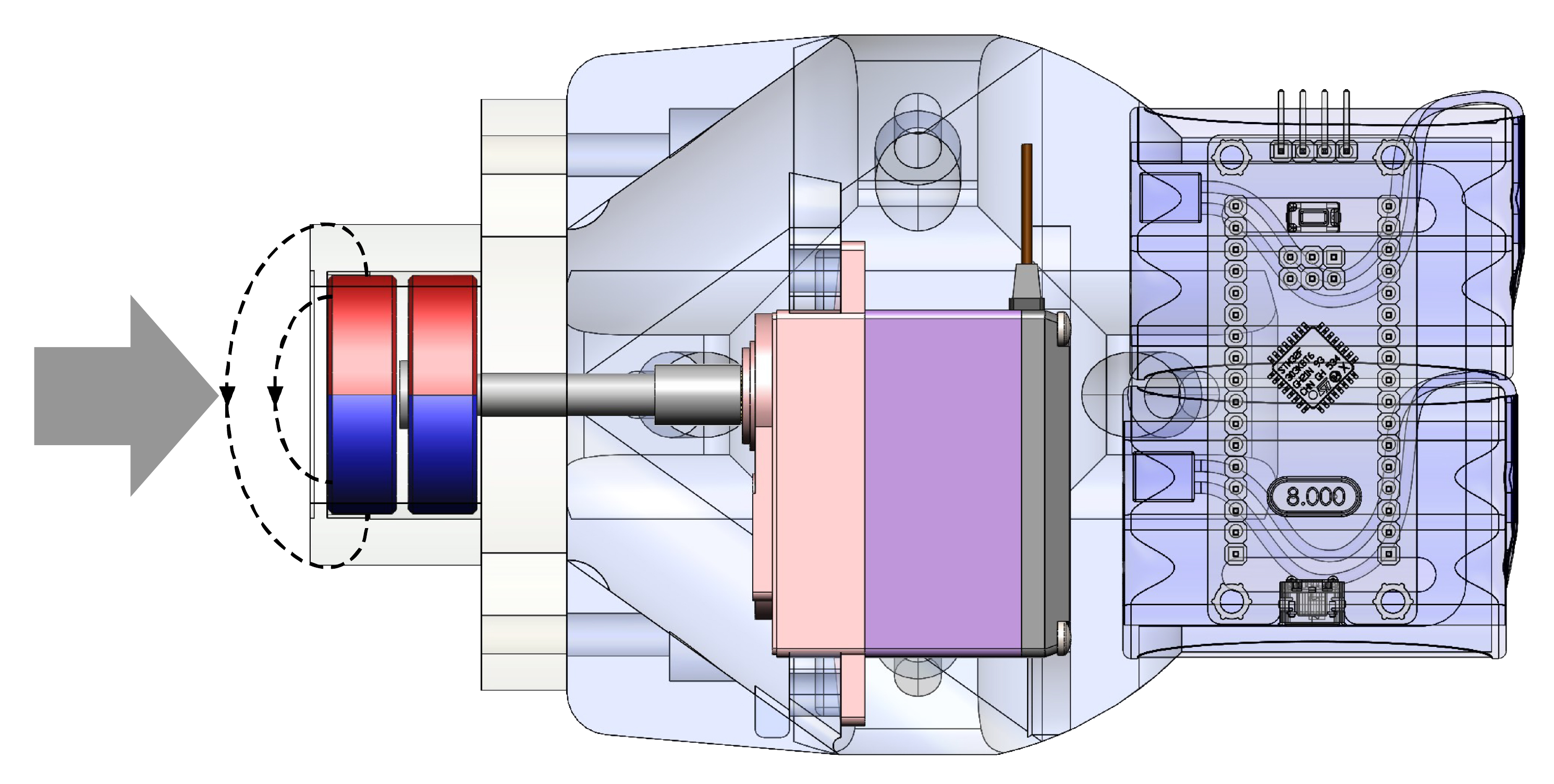}
				\put(-5,40){Docked}
			\end{overpic}
		}
		
		\subfloat[]{
			\begin{overpic}[width=1\textwidth]{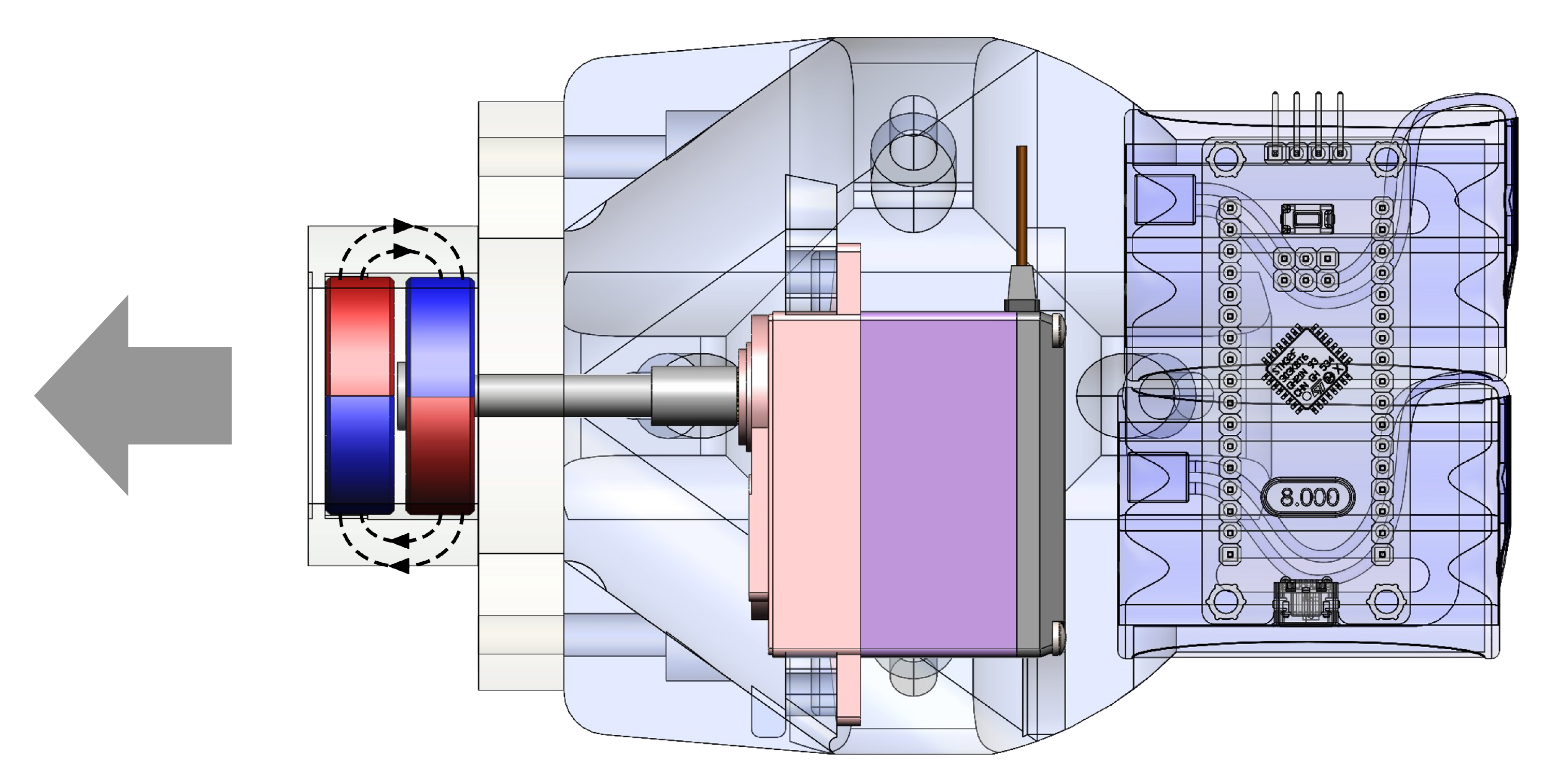}
				\put(-5,40){Undocked}
			\end{overpic}
		}
	\end{minipage}
	\caption{(a) System overview of the docking system. (b) The docked state and (c) undocked state when the magnet is switched on and off.}
	\label{fig:docking_sys}
\end{figure}

To catch and release the building modules, a crucial design task is an appropriate docking system. 
Among the many creative mechanisms proposed are a hook and loop driven by servomotors \cite{o2014self, paulos2015automated} and a ball-socket mechanism integrating a funnel to guide the docking action \cite{mateos2019autonomous}. 
However, there is only a short time window for the accurate alignment of these gender-opposite docking systems on a turbulent water surface. 
To promote the docking success rate, some studies \cite{DBLP:journals/corr/abs-2001-04293, liu2014self} increase the acceptance area by enlarging the connectors or utilizing multiple connectors, which increases the weight, volume, and manufacturing cost. 
Other studies \cite{bhattacharjee2021magnetically, haghighat2016characterization} build connectors based on magnets or electromagnets that can conduct instant docking for transient opportunities. Usually, however, the permanent magnetic connection creates additional constraints for path planning, and electromagnets cannot offer robust connections under low voltages ($\leq 24\ \rm V$).

Therefore, we design an instant docking system using a switchable permanent magnet to capture the building modules by means of ferromagnetic interfaces.
Compared with the electromagnet connection system \cite{zhang2016self}, this docking system provides the same magnetic force but with many advantages, including a lighter mass, lower voltage, and longer duration.

Fig. \ref{fig:docking_sys} (a) exhibits the main components of the docking system. 
Fig. \ref{fig:docking_sys} (b) and (c) show how the switchable magnet in the docking system is actuated by a servo motor. An ESP8266 board is used to communicate with the processing unit and switch the magnet on and off. 
Our measurements of the attraction force and connection strength between two docking systems indicate that
when two switched-on magnets are attracting each other, the longitudinal connection force is up to $570\ \rm N$, while the lateral force is $150\ \rm N$. 
Furthermore, even if only one magnet is on and attracting ferromagnets, the connection forces are still $340\ \rm N$ and $67\ \rm N$, respectively.
Therefore, the magnetic force is strong enough to build a solid connection.
To avoid lateral separation and reduce the impact from the body during docking, an elastic wire line shock damper is installed at the bottom, which can restore its shape when disconnected. 
This flexible connection allows the docking system to twist at about $90\ \rm degree$ and shift in the longitudinal ($4\ \rm cm$ at most) and lateral ($2.5\ \rm cm$ at most) directions, allowing the system to tolerate relatively large misalignment.

\section{NONLINEAR MODEL PREDICTIVE CONTROL}
\label{sect:control}

To achieve precise motion control, we develop a real-time NMPC method that is effective for all forms of the TransBoat. 
NMPC is a nonlinear optimal control strategy that can predict the future response and compute optimal control commands for the robot based on an explicit process model \cite{Gruene2017}.
To obtain the robot model, Fossen's USV model \cite{fossen2011handbook} is adjusted for the transformable USV and its model parameters are derived via a system parameter identification procedure. 

\subsection{Dynamic Modeling}

We represent the TransBoat using a three-dimensional inertial coordinate ($O_w \text{-} X_w Y_w Z_w$) system. 
As only the motion on the water surface is of concern, the position and orientation of the robot are defined as $\boldsymbol{\eta }=\left[ x\,\,y\,\,\psi \right] ^T$, relative to the center of mass (COM).
Also, a body-fixed coordinate ($O_b \text{-} X_b Y_b Z_b$) is set at the body center of the TransBoat with the $X_b$ axis toward the USV front,
which is also considered to be the COM in light of the structural symmetry.
In the body frame, the robot velocity is denoted as $ \mathbf{v} =\left[ u\,\,v\,\,r \right] ^T$.

When describing the dynamic model of the TransBoat,
an assumption is made that the model variation relative to the expansion length is a quasi-static process, 
because during movement its shape changes sufficiently slowly to meet the stability requirement.

According to the Fossen model in \cite{fossen2011handbook}, the dynamic model of the TransBoat can be represented as
\begin{equation} \label{eq:dynamic_eta}
	\boldsymbol{\dot{\eta}}=\mathbf{R}\left( \psi \right) \mathbf{v} ,
\end{equation}
\begin{equation} \label{eq:dynamic_v}
	\mathbf{\widetilde M \dot{v}}+\mathbf{\widetilde C}\left( \mathbf{v} \right) \mathbf{v}+ \mathbf{\widetilde D}\left( \mathbf{v} \right) \mathbf{v}=\mathbf{F} ,
\end{equation}
where the model parameters can be divided into two categories in accordance with their relevance to the expansion length $l$ of the TransBoat. 

One category is only related to the TransBoat's state rather than its shape.
\begin{itemize}
\item The transformation matrix $\mathbf{R}\left( \psi \right)$ converting a state vector from body frame to inertial frame is
\begin{equation}
\mathbf{R}\left( \psi \right) =\left[ \begin{matrix}
\cos \psi&		-\sin \psi&		0\\
\sin \psi&		\cos \psi&		0\\
0&		0&		1\\
\end{matrix} \right] .
\end{equation}
\end{itemize}

The other parameters are functions of the robot state and expansion length, and are indicated by a tilde symbol ``$\widetilde{\quad}$''.
\begin{itemize}
\item As origin $O_b$ coincides with the COM, the mass matrix $\mathbf{\widetilde M}$ is
\begin{equation}
\mathbf{\widetilde M} = \textrm{diag} \left\{ m_{1}\left( l \right), m_{2}\left( l \right), m_{3}\left( l \right) \right\} ,
\end{equation}
which is a diagonal matrix combining the robot's inertial mass and the added mass.

\item The matrix of Coriolis and centripetal terms $\mathbf{\widetilde C}\left( \mathbf{v} \right)$ is
\begin{equation}
\mathbf{\widetilde C}\left(\mathbf{v} \right) =
\left[ \begin{matrix}
0&		0&		-m_{2}\left( l \right) v\\
0&		0&		m_{1}\left( l \right) u\\
m_{2}\left( l \right) v&		-m_{1}\left( l \right) u&		0\\
\end{matrix} \right]. 
\end{equation}

\item Owing to the low movement speed and the symmetrical structure, the drag matrix $\mathbf{\widetilde D}$ is expressed as a linear damping term
\begin{equation}
\mathbf{\widetilde D} =
\textrm{diag} \{X_u\left( l \right),  Y_v\left( l \right), N_r\left( l \right) \}.
\end{equation}
\end{itemize}
\begin{figure} [htbp] 
	\centering
	\includegraphics[width=0.8\linewidth]{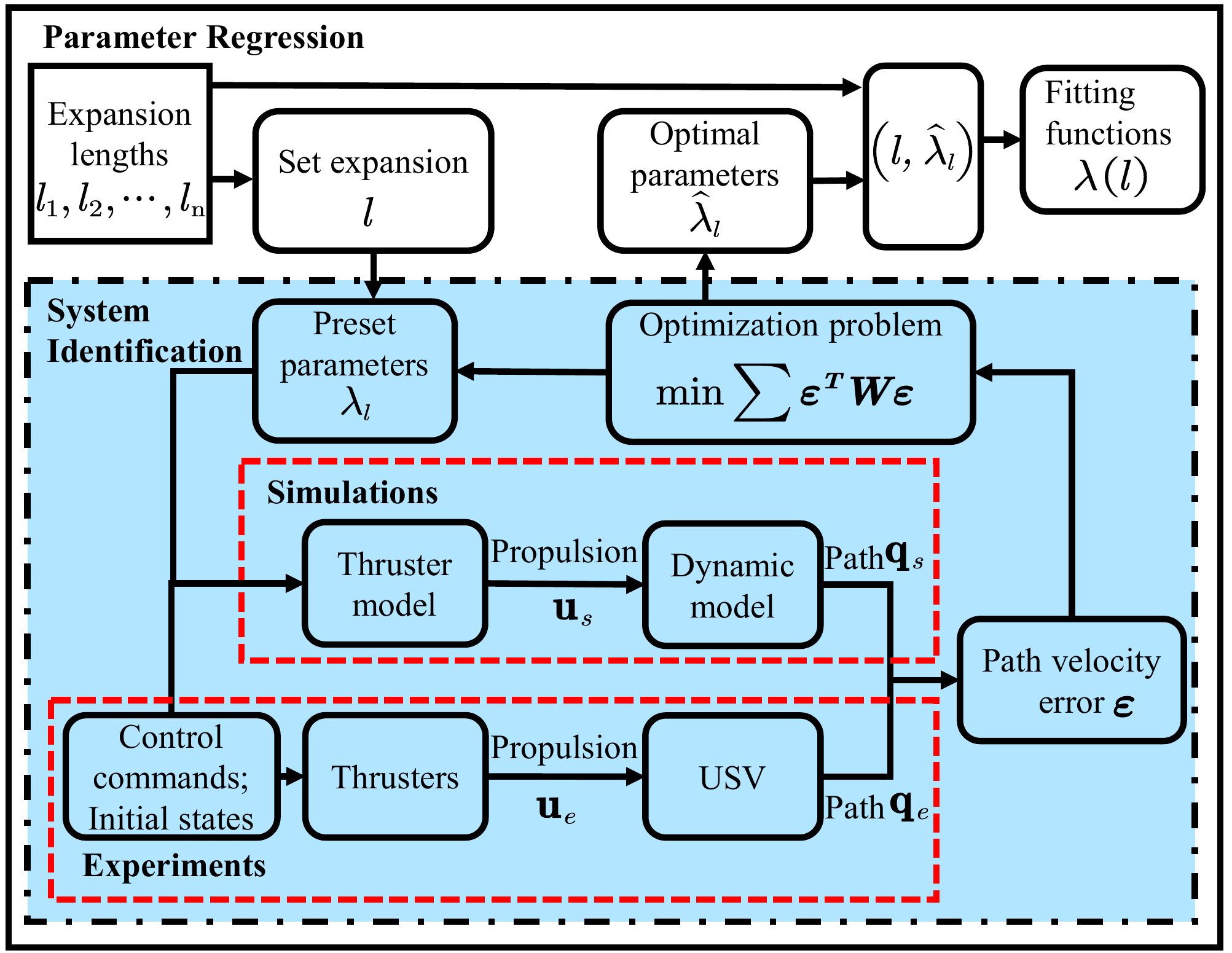}
	\caption{Block diagram of the gray-box parameter identification. }
	\label{fig:SystemID}
\end{figure}

\begin{figure*} [htbp] 
	\centering
	\begin{minipage}[b]{.68\linewidth}
		\centering
		\subfloat[]{
			\begin{overpic}[width=0.32\textwidth]{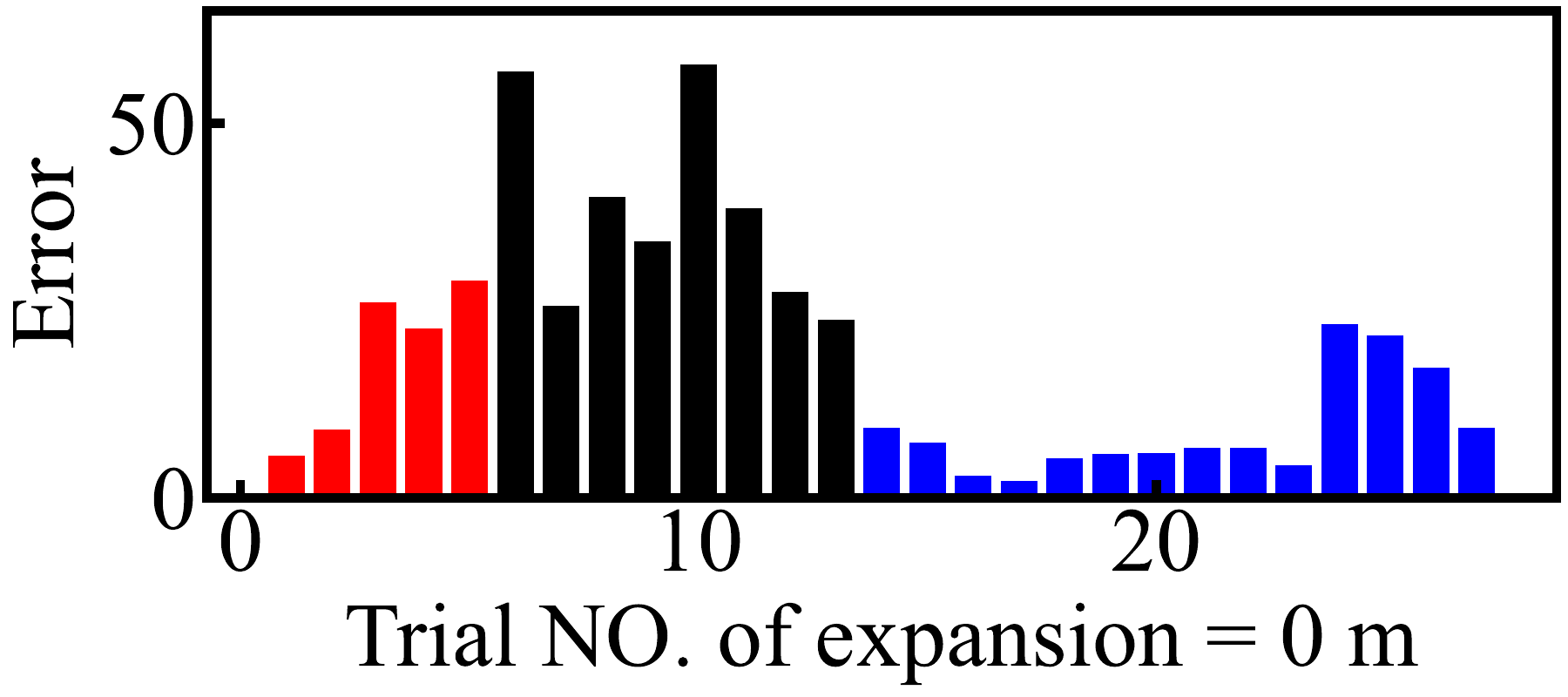}
				\put(5,45){\textcolor[rgb]{1,0,0}{\small{Circling}}}
				\put(31,45){\small{Spinning}}
				\put(60,45){\textcolor[rgb]{0,0,1}{\small{Straight-line}}}
			\end{overpic}
		}
		\subfloat[]{
			\begin{overpic}[width=0.32\textwidth]{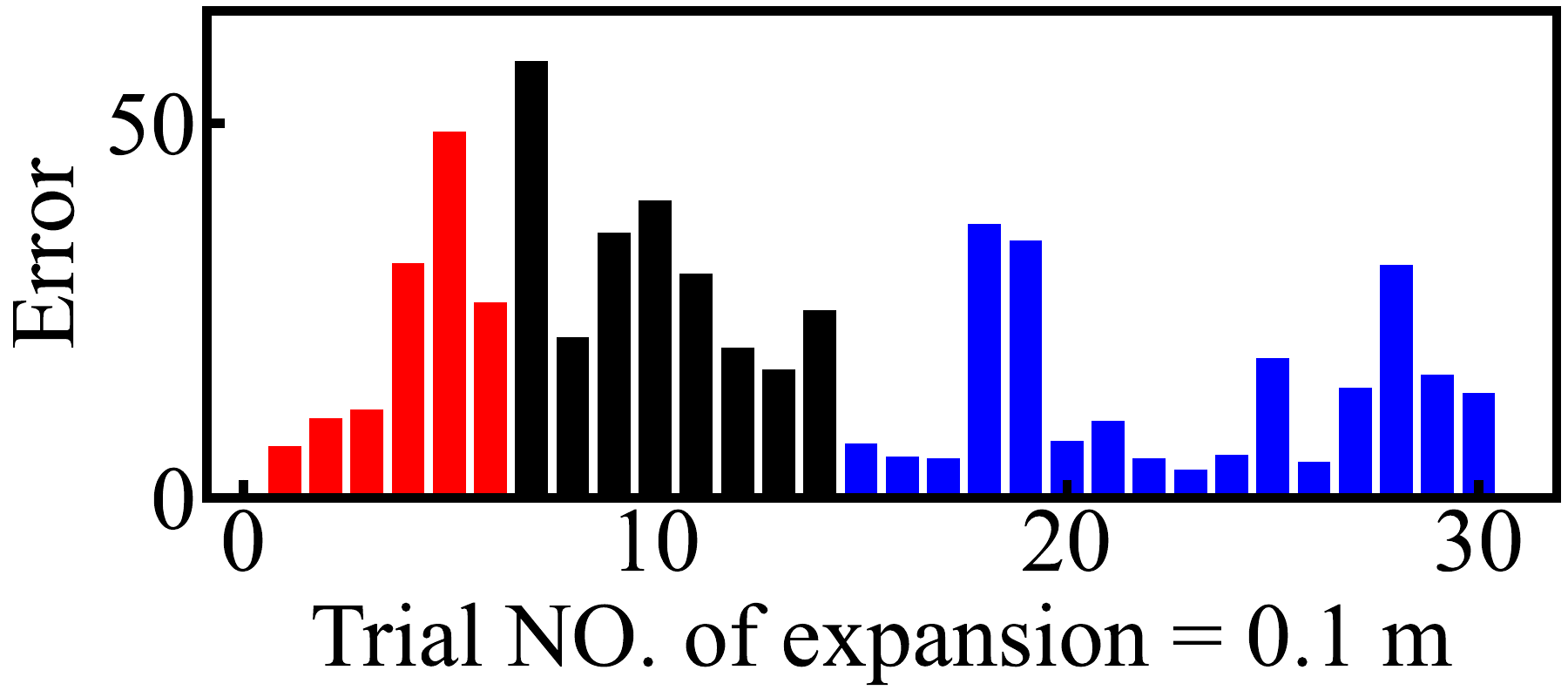}
				\put(5,45){\textcolor[rgb]{1,0,0}{\small{Circling}}}
				\put(31,45){\small{Spinning}}
				\put(60,45){\textcolor[rgb]{0,0,1}{\small{Straight-line}}}
			\end{overpic}
		}
		\subfloat[]{
			\begin{overpic}[width=0.32\textwidth]{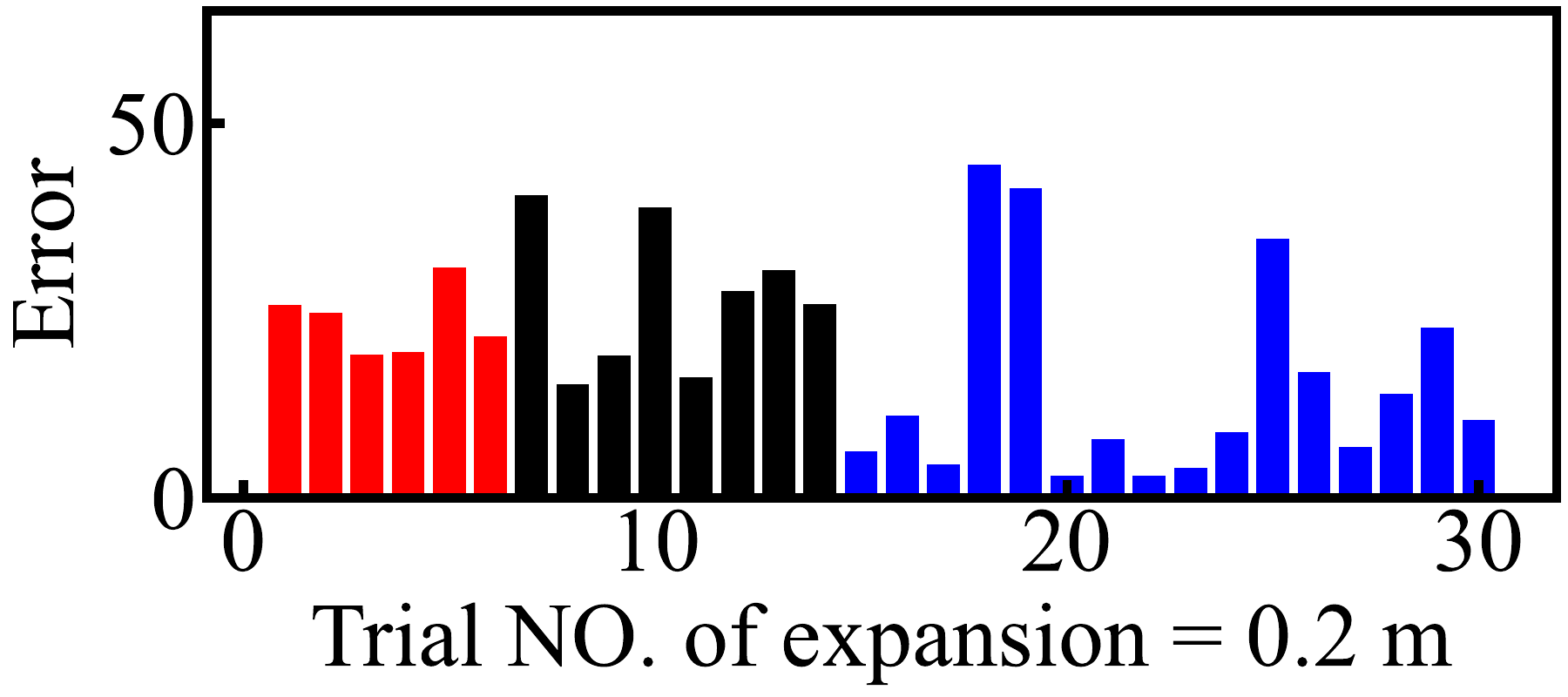}
				\put(5,45){\textcolor[rgb]{1,0,0}{\small{Circling}}}
				\put(31,45){\small{Spinning}}
				\put(60,45){\textcolor[rgb]{0,0,1}{\small{Straight-line}}}
			\end{overpic}
		}
	
		\subfloat[]{
			\includegraphics[width=0.32\textwidth]{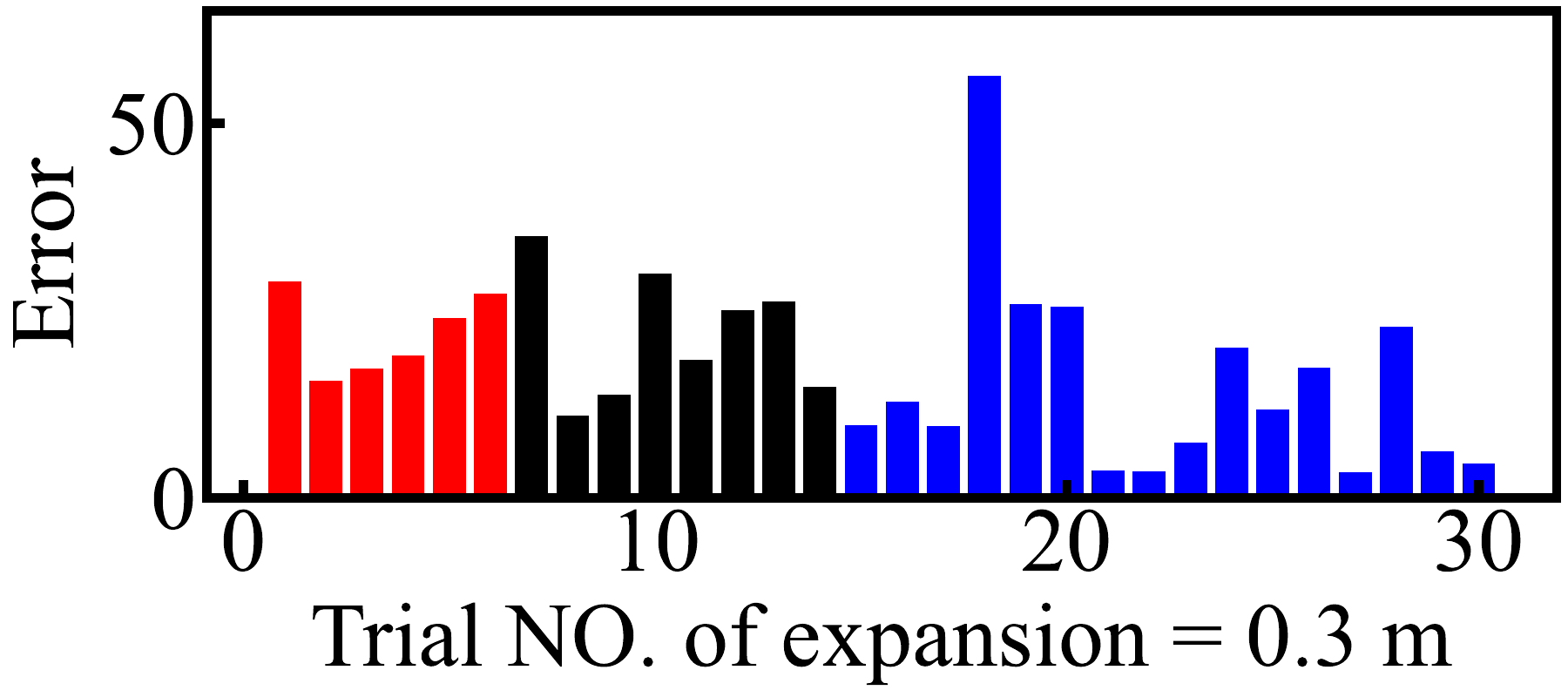}
		}
		\subfloat[]{
			\includegraphics[width=0.32\textwidth]{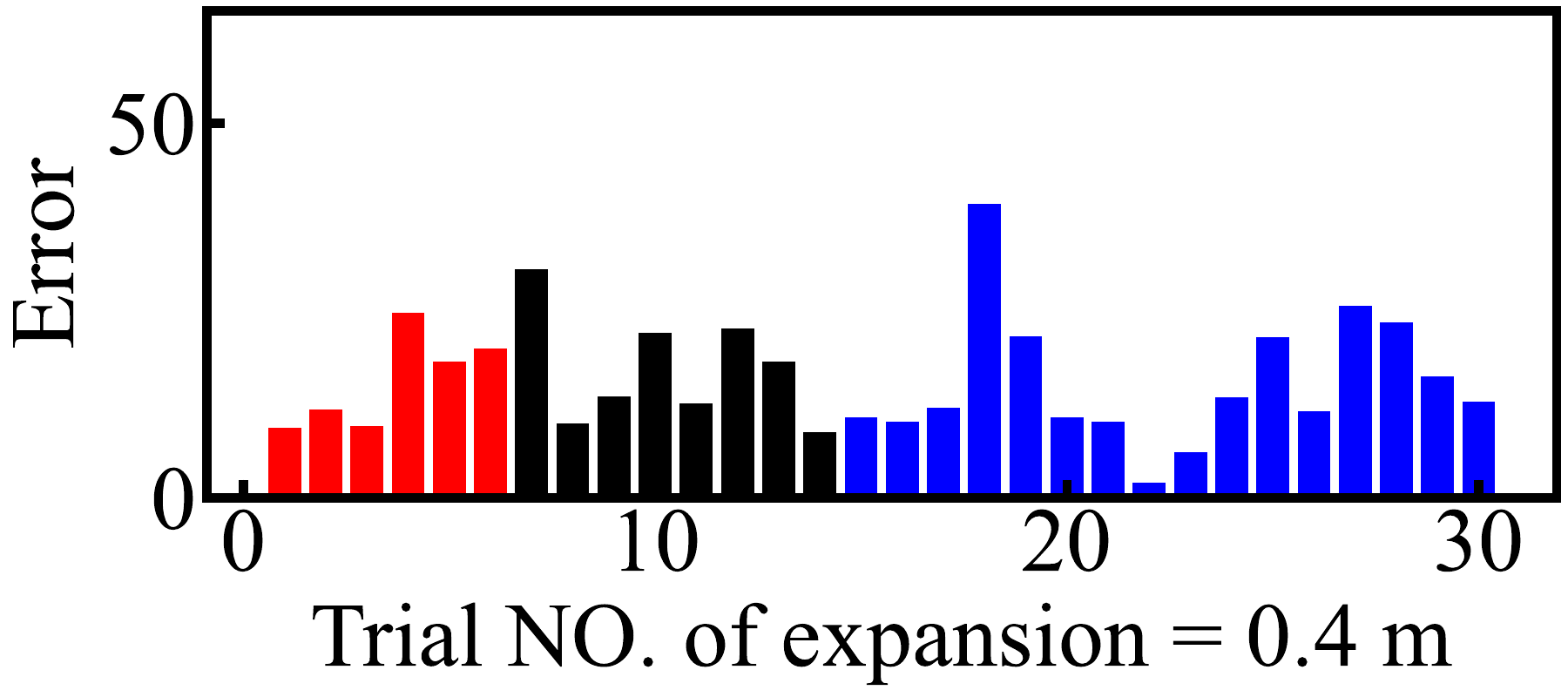}
		}
		\subfloat[]{
			\includegraphics[width=0.32\textwidth]{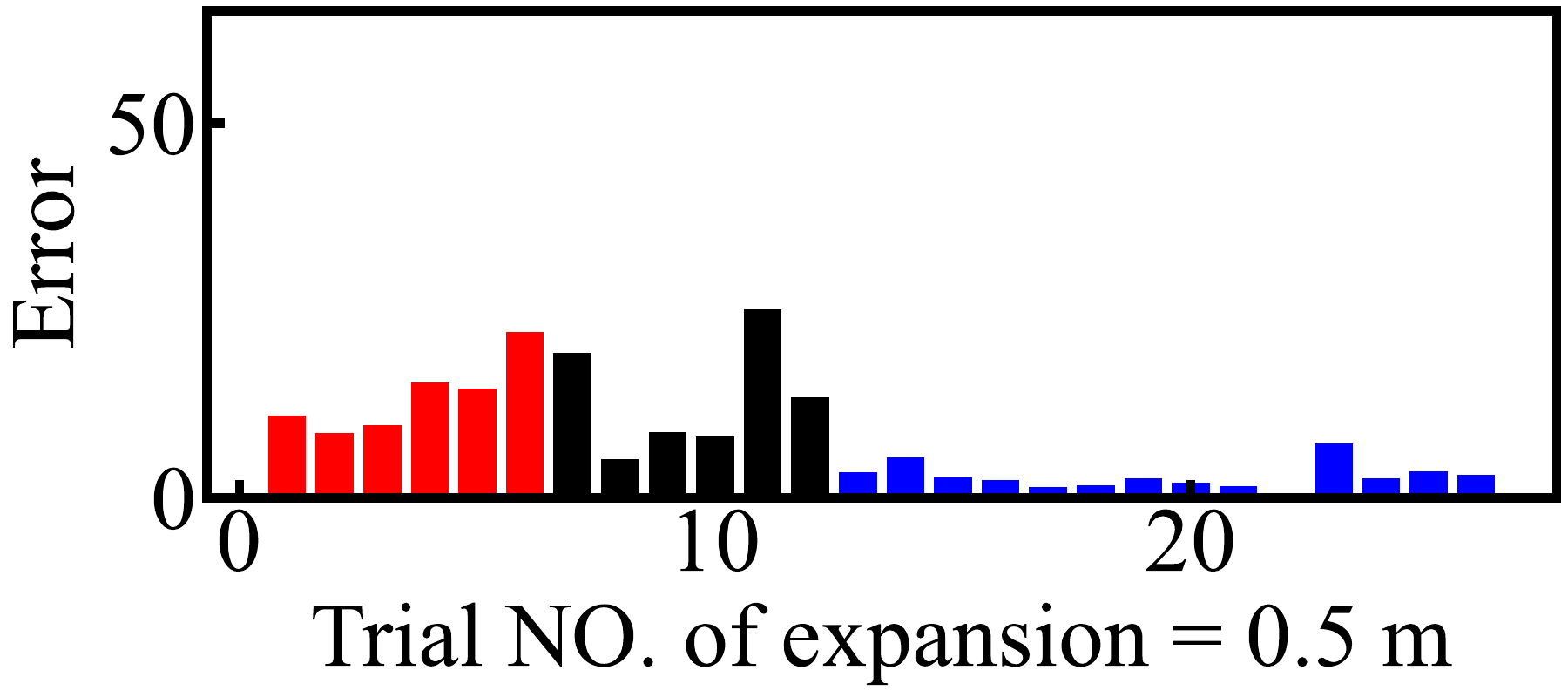}
		}
	\end{minipage}
	\begin{minipage}[b]{.3\linewidth}
		\centering
		\subfloat[]{
			\includegraphics[width=1\linewidth]{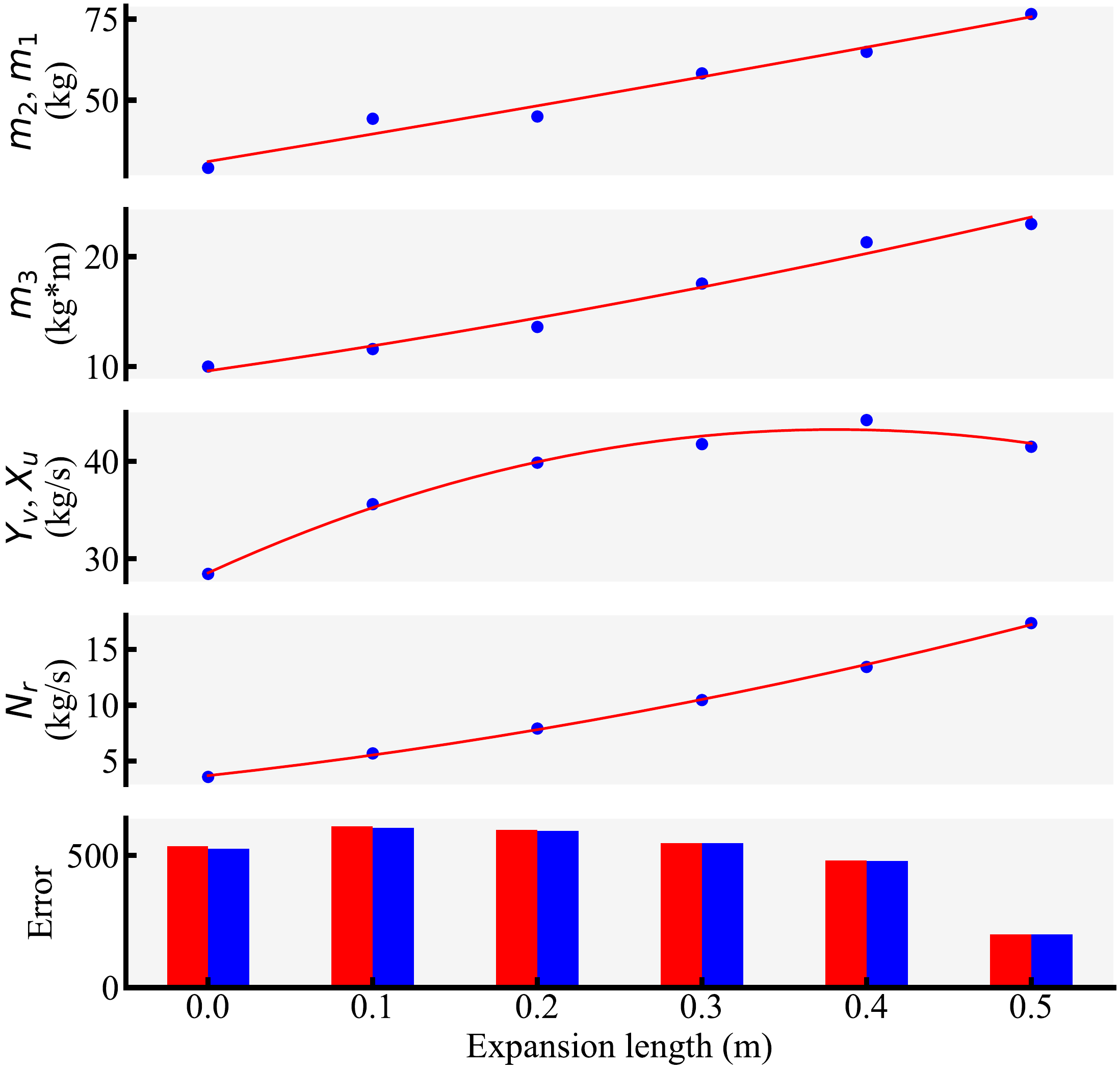}
		}
	\end{minipage}
	\caption{(a) $\sim$ (f) Residual errors after the system identification in tests of expansion $= 0 \sim 0.5\ \text{m}$, where NO. is short for number. In each test, the error $\boldsymbol{\varepsilon}$ of each trial is the sum of linear and angular velocity differences at all times between simulation and experiment. (g) Results of system identification and the parameter functions for the dynamic model. The identified parameters before and after the regression are denoted in blue and red, respectively, as are the errors. }
	\label{fig:parameters}
\end{figure*}

\subsection{System Parameter Identification}

Let $\mathbf{q} = \left[ \boldsymbol{\eta}^\mathrm{T} ,\,  \mathbf{v}^\mathrm{T} \right]^\mathrm{T} = \left[ x,\,y,\,\psi,\,u,\,v,\,r \right] ^\mathrm{T}$ 
be the state vector of the TransBoat.
Then the dynamic model from Eqs. (\ref{eq:dynamic_eta}) and (\ref{eq:dynamic_v}) can be reformulated as
\begin{equation} \label{eq:dynamic_q}
\mathbf{\dot{q}} = \mathbf{A}\left( \mathbf{q} \right) \mathbf{q}+\mathbf{B} \mathbf{u},
\end{equation}
where $\mathbf{u}$ is the control vector composed of four propulsion forces as defined in Eq. (\ref{eq:forces}), the coefficients 
\begin{equation}
\mathbf{A}\left( \mathbf{q} \right) =
\left[ \begin{matrix}
\mathbf{0}_{3 \times 3}&		\mathbf{R}\left( \psi \right)\\
\mathbf{0}_{3 \times 3}&		-\mathbf{\widetilde M}^{-1}\left( \mathbf{\widetilde C}\left( \mathbf{v} \right) +\mathbf{\widetilde D} \right)\\
\end{matrix} \right], {\rm and}
\end{equation} 
\begin{equation} \label{eq:B}
\mathbf{B} =
\left[ \begin{array}{c}
\mathbf{0}_{3 \times 4}\\
\mathbf{\widetilde M}^{-1}\mathbf{E}\\
\end{array} \right].
\end{equation}
Therefore, the robot state is updated with the control input $\mathbf{u}$ and current state $\mathbf{q}$.

The dynamic model in Eq. (\ref{eq:dynamic_q}) is a gray-box model with undetermined hydrodynamic parameters, namely, mass coefficients $m_{1},m_{2},m_{3}$ and linear drag coefficients $X_u,Y_v,N_r$.
Each parameter is a function of the expansion length $l$ of the TransBoat.
In this regard, a generalized system identification algorithm \cite{yu2016data, wang2018design} is used together with a regression process, as shown in Fig. \ref{fig:SystemID}.
Specifically, we first set an expansion length $l$, then input the identical control commands and initial states of the experiments into the simulation. Next, we search for the optimal model parameters by minimizing the overall error between the simulated paths and the experimental paths. Following this, a regression procedure is used to fit the polynomial function of expansion $l$ for each parameter.
This method is also valid for various surface vehicles. 

Let $\lambda = \{ m_{1},m_{2},m_{3},X_u,Y_v,N_r \}$ and $\lambda_l$ denote the specific values at a given expansion length $l$. 
The gray-box identification process can be regarded as the optimization problem described below,
\begin{equation} \label{eq:identification}
\begin{aligned}
\mathop {\text{arg}\min} \limits_{\lambda_l}\sum_t{\boldsymbol{\varepsilon}\left( t \right) ^T\mathbf{W} \boldsymbol{\varepsilon} \left( t \right)}, 
\\
s.t.   \lambda ^l_{l}\le \lambda _l \le \lambda ^u_{l} ,
 \end{aligned}
\end{equation}
where $\boldsymbol{\varepsilon} \left( t \right) = \mathbf{v}_e\left( t \right) - \mathbf{v}_s\left( t \right)$ is the error between the simulated velocity $\mathbf{v}_s$ and the experimental velocity $\mathbf{v}_e$,
$\mathbf{W}$ is the diagonal weight matrix, and
$\lambda ^l_{l}$ and $\lambda ^u_{l}$ denote the lower and upper bounds of $\lambda_l$, respectively.

After the identification process, a series of expansion-parameter pairs $l \text{-} \lambda_l$ are determined from the experimental data. Then the parameter functions are fitted by polynomial regression as follows,
\begin{equation} \label{eq:parameters}
\lambda \left( l \right) =\sum_{j=0}^M{c_j l^j} ,
\end{equation}
where $M$ is the function order and $c_j$ is the coefficient.

\subsection{Nonlinear Model Predictive Control}

We now have the dynamics of the TransBoat, which can be expressed in a state-space representation for any symmetrical form, as in Eq. (\ref{eq:dynamic_q}).
In this section, an NMPC algorithm, proposed in \cite{Gruene2017, wang2018design}, is designed to guarantee that the TransBoat can accurately track the desired trajectories in different forms.

Within every finite prediction horizon $T$ of the MPC law, the optimal control sequence is obtained by minimizing a predicted performance cost function. 
Then, in the sampling time slot $[t,t+T]$, the optimization problem can be formulated as
\begin{equation} \label{eq:}
\begin{aligned}
\underset{\mathbf{u}\left( \tau \right)}{\min}  \quad
&\mathbf{J}\left( \mathbf{q}\left( \tau \right) ,\mathbf{u}\left( \tau \right) \right) 
\\
\text{s}.\text{t}. \quad
& \mathbf{q}\left( \tau \right)  \in \overline{Q},\ 
\mathbf{u}\left( \tau \right) \in \overline{U}, 
\\
& \mathbf{q}\left( t \right) = \mathbf{q}_0, \
\forall \tau \in [t,t+T],
\end{aligned}
\end{equation}
where $\overline{Q} = [\mathbf{q}_{\min}, \mathbf{q}_{\max}]$ and $\overline{U} = [\mathbf{u}_{\min}, \mathbf{u}_{\max}]$ are the feasible sets of states and control inputs of the system, respectively, and 
$\mathbf{q}_0$ stands for the state feedback measured by sensors at time $t$.
$\mathbf{J}\left( \mathbf{q}\left( \tau \right) ,\mathbf{u}\left( \tau \right) \right)$ denotes the cost function which can be described as
\begin{equation}
\mathbf{J}\left( \mathbf{q}\left( t \right) ,\mathbf{u}\left( t \right) \right) 
= \sum_{\tau = t}^{t+T}{\mathbf{J}_\tau \left( \mathbf{q}\left( \tau \right) ,\mathbf{u}\left( \tau \right) \right)}+\mathbf{J}_N\left( \mathbf{q}\left( t+T \right) \right),
\end{equation}
where $\mathbf{J}_\tau$ is the stage cost function based on the prediction of Eq. (\ref{eq:dynamic_q}), and 
$\mathbf{J}_N$ is the terminal cost.
In our NMPC algorithm, they are implemented as the following quadratic forms
\begin{equation}
\begin{aligned}
\mathbf{J}_\tau \left( \mathbf{q} ,\mathbf{u} \right)
& = \boldsymbol{\varepsilon}_q \left( \tau \right) ^T\mathbf{Q} \boldsymbol{\varepsilon}_q \left( \tau \right) +
\boldsymbol{\varepsilon}_u \left( \tau \right) ^T\mathbf{H} \boldsymbol{\varepsilon}_u \left( \tau \right),
\\
\mathbf{J}_N\left( \mathbf{q} \right) 
& = \boldsymbol{\varepsilon}_q \left( t+T \right) ^T\mathbf{Q}_N \boldsymbol{\varepsilon}_q \left( t+T \right),
\end{aligned}
\end{equation}
where $ \boldsymbol{\varepsilon}_q=\mathbf{q}-\mathbf{q}_0 $ and $\mathbf{q}_0$ denotes the target state. $ \boldsymbol{\varepsilon}_u=\mathbf{u} \left( \tau+1 \right)-\mathbf{u} \left( \tau \right) $ stands for the variation in control inputs. 
The optimization process can minimize them to those that can smooth the boat's motion.
$\mathbf{Q}$ and $\mathbf{H}$ are the positive definite matrices that penalize these two deviations
and $\mathbf{Q}_N$ is the terminal penalty matrix that can enhance the NMPC algorithm.
These weighting matrices are defined as follows:
\begin{equation}
\begin{aligned}
\mathbf{Q} &= \mathrm{diag}\left( 8000,8000,4000,0.01,0.01,0.01 \right) , \\
\mathbf{H} &= \mathrm{diag}\left( 0.01,0.01,0.01,0.01 \right),
\end{aligned}
\end{equation}
where $\mathbf{Q}_N$ is set to the same values as $\mathbf{Q}$.

This problem can be rapidly solved by the CasADi solver \cite{andersson2019casadi}, a nonlinear optimization tool, within 0.1 seconds in each control cycle.
The solution, an optimal control sequence $ \mathbf{u}^*$, is computed online.
To improve the robustness of the NMPC algorithm, only the first solution of the control sequence is executed, and the entire algorithm is conducted repeatedly for every control interval.
The algorithm ends when the difference between the current state and the target is lower than a preset threshold.

\begin{table}[tbp]
	\centering
	\caption{TECHNICAL SPECIFICATIONS OF TRANSBOAT}
	\label{tab:spec}
	\begin{tabular}{lcc}
		\toprule[1.5pt]
		\textbf{Parameter} & \textbf{Contracted form} & \textbf{Expanded form} \\
		\midrule
		Size (m) & 1.2  & 2.2  \\
		Mass(kg) & 42 & 42 \\
		Max. Speed (m/s) &  0.6   & 0.4 \\
		Max. Steering Speed ($^{\circ}$/s) &  139.80  & 60.16 \\ % 2.44, 1.05
		Capacity (kg) & 70 & 70 \\
		Battery Life (hours) & 4  & 4  \\
		\bottomrule[1.5pt]
	\end{tabular}
\end{table}

\begin{figure*} [thpb]
	\centering
	\subfloat[]{
		\begin{minipage}[b]{0.3\textwidth}
			\centering
			\includegraphics[width=0.9\textwidth]{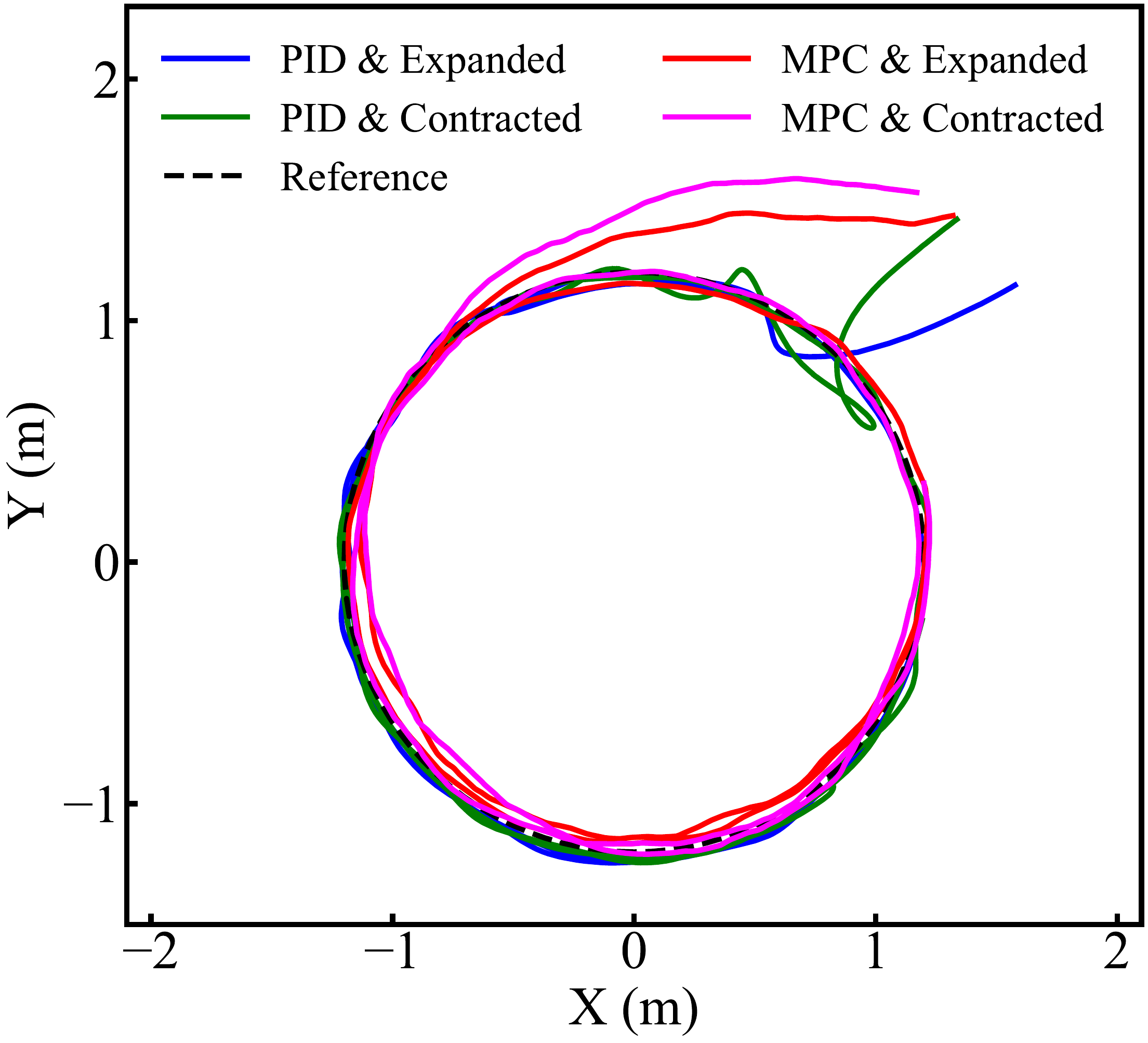}
			\newline
			\includegraphics[width=1\textwidth]{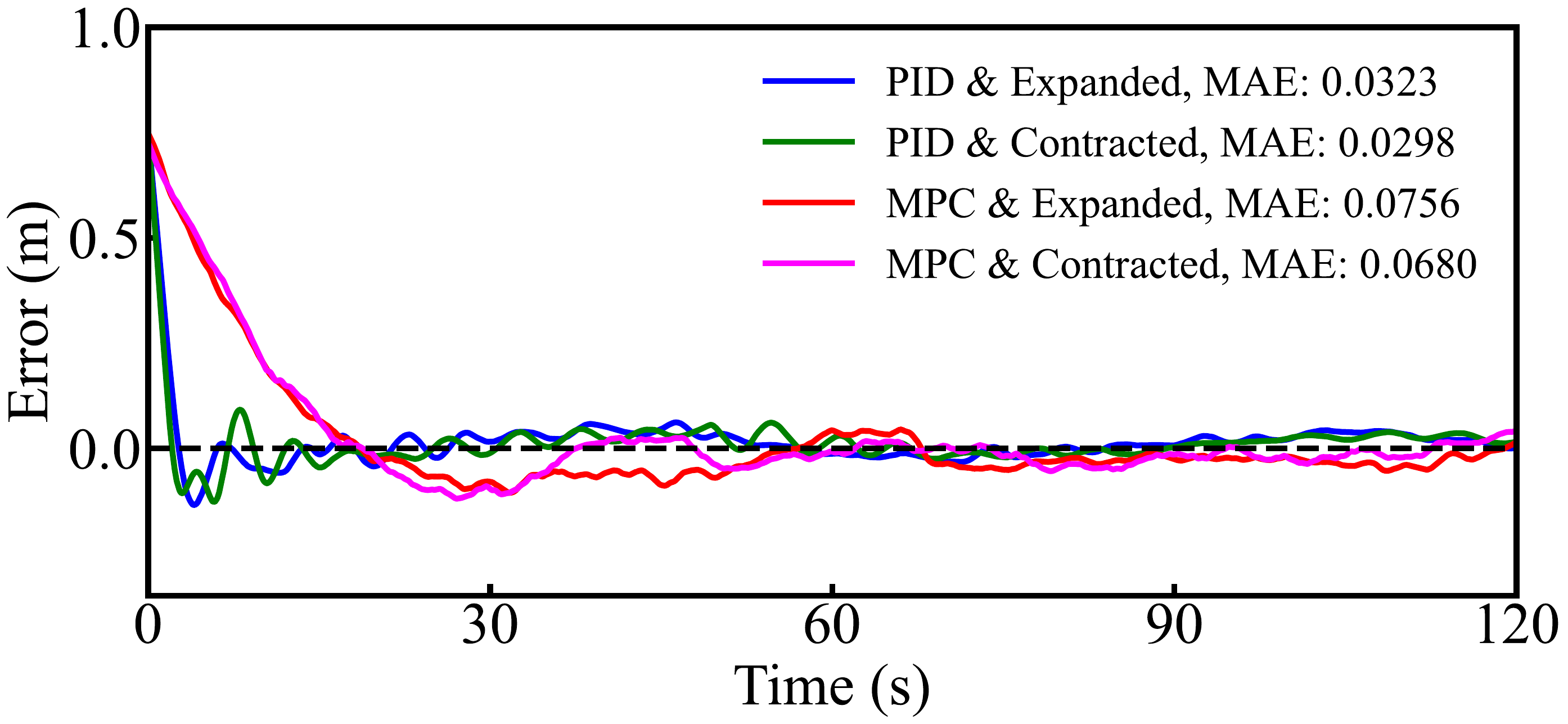}
			\newline
			\includegraphics[width=1\textwidth]{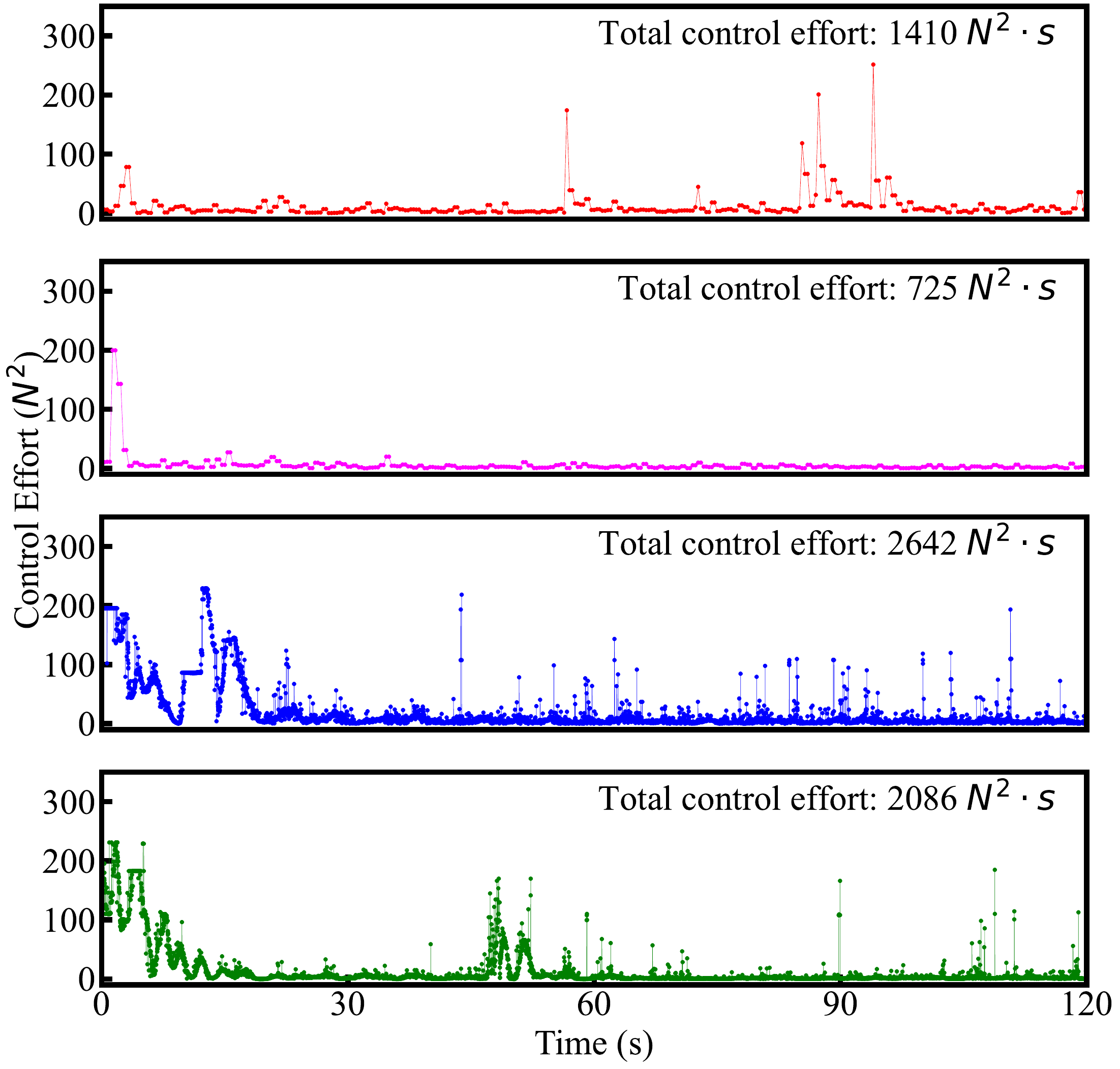}
		\end{minipage}
	}
	\subfloat[]{
		\begin{minipage}[b]{0.3\textwidth}
			\centering
			\includegraphics[width=0.9\textwidth]{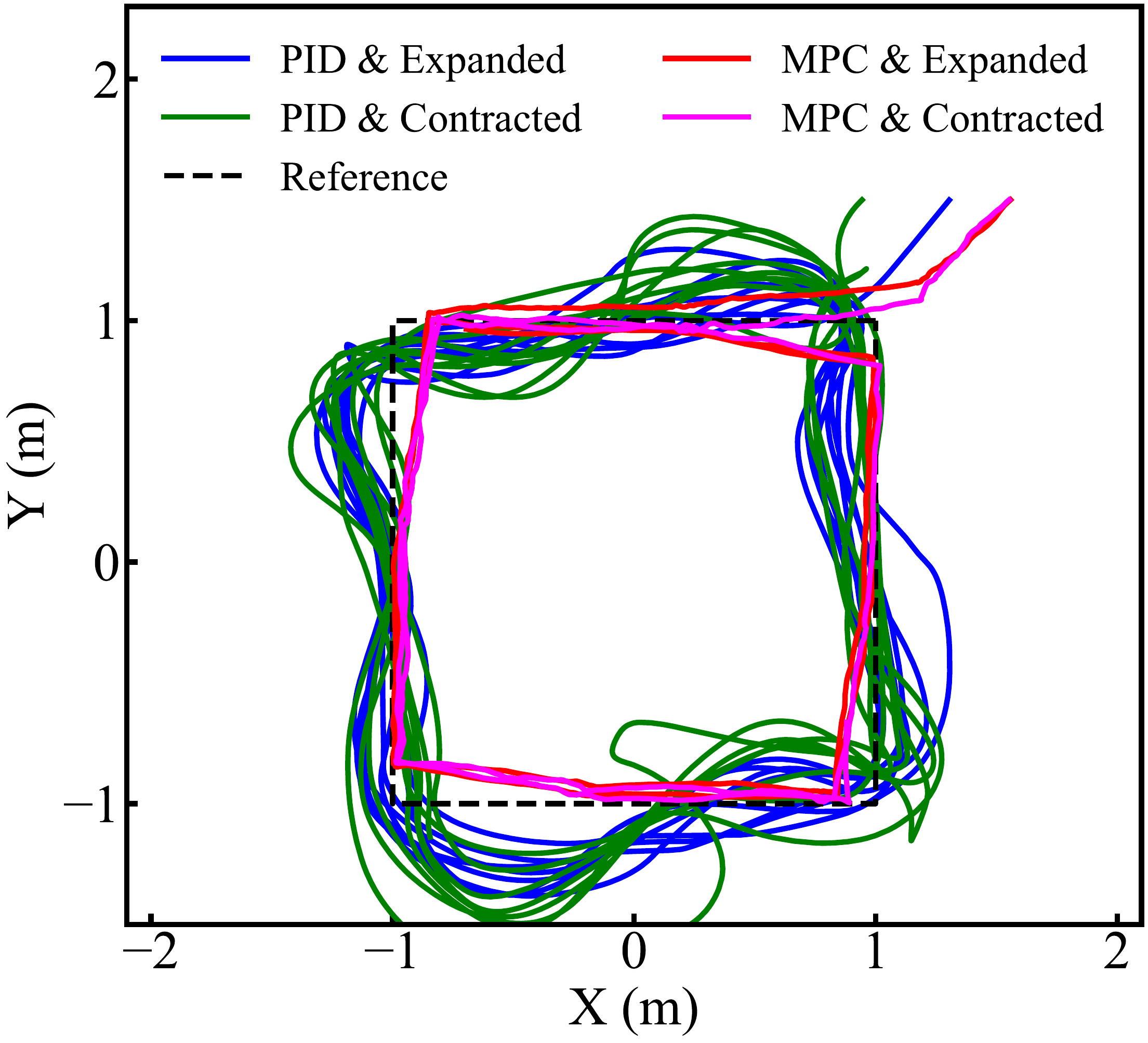}
			\newline
			\includegraphics[width=1\textwidth]{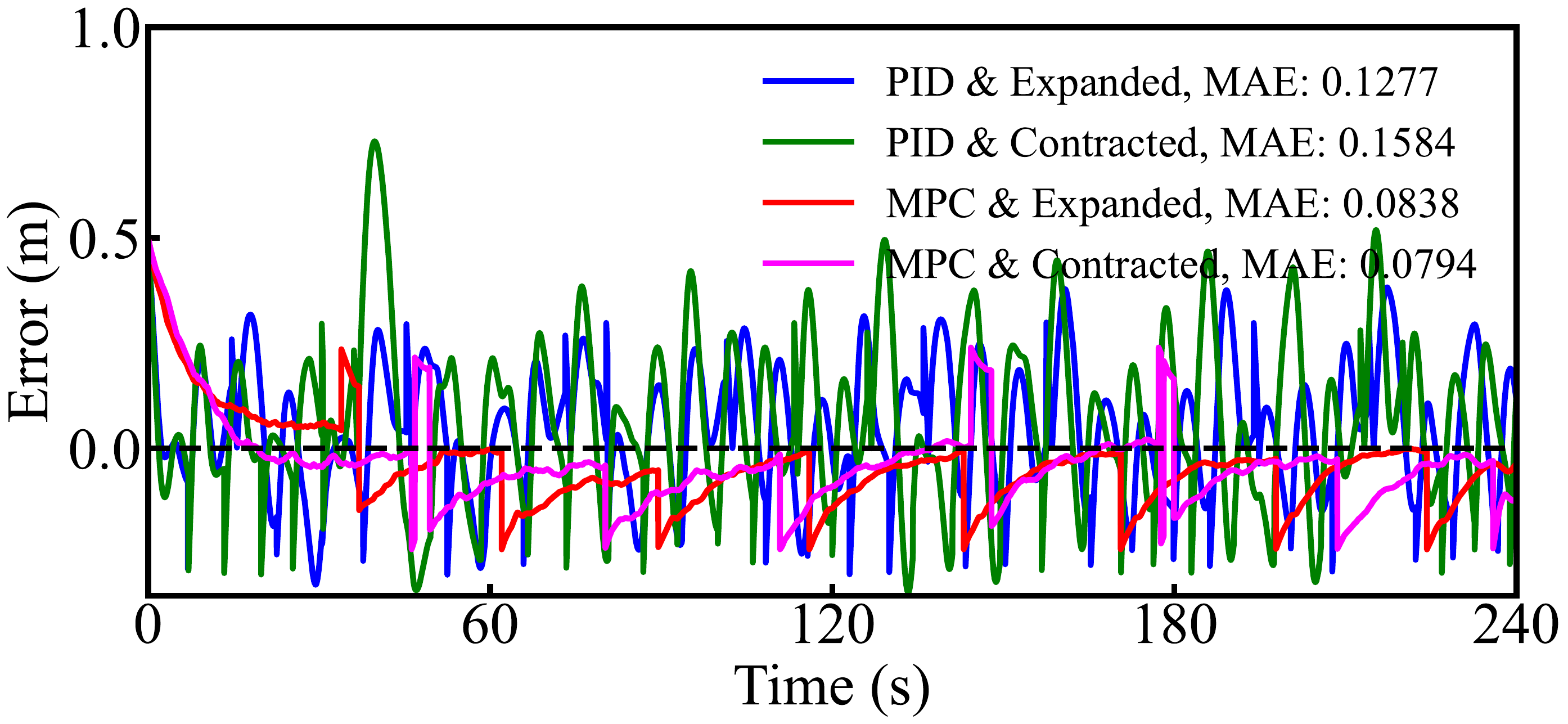}
			\newline
			\includegraphics[width=1\textwidth]{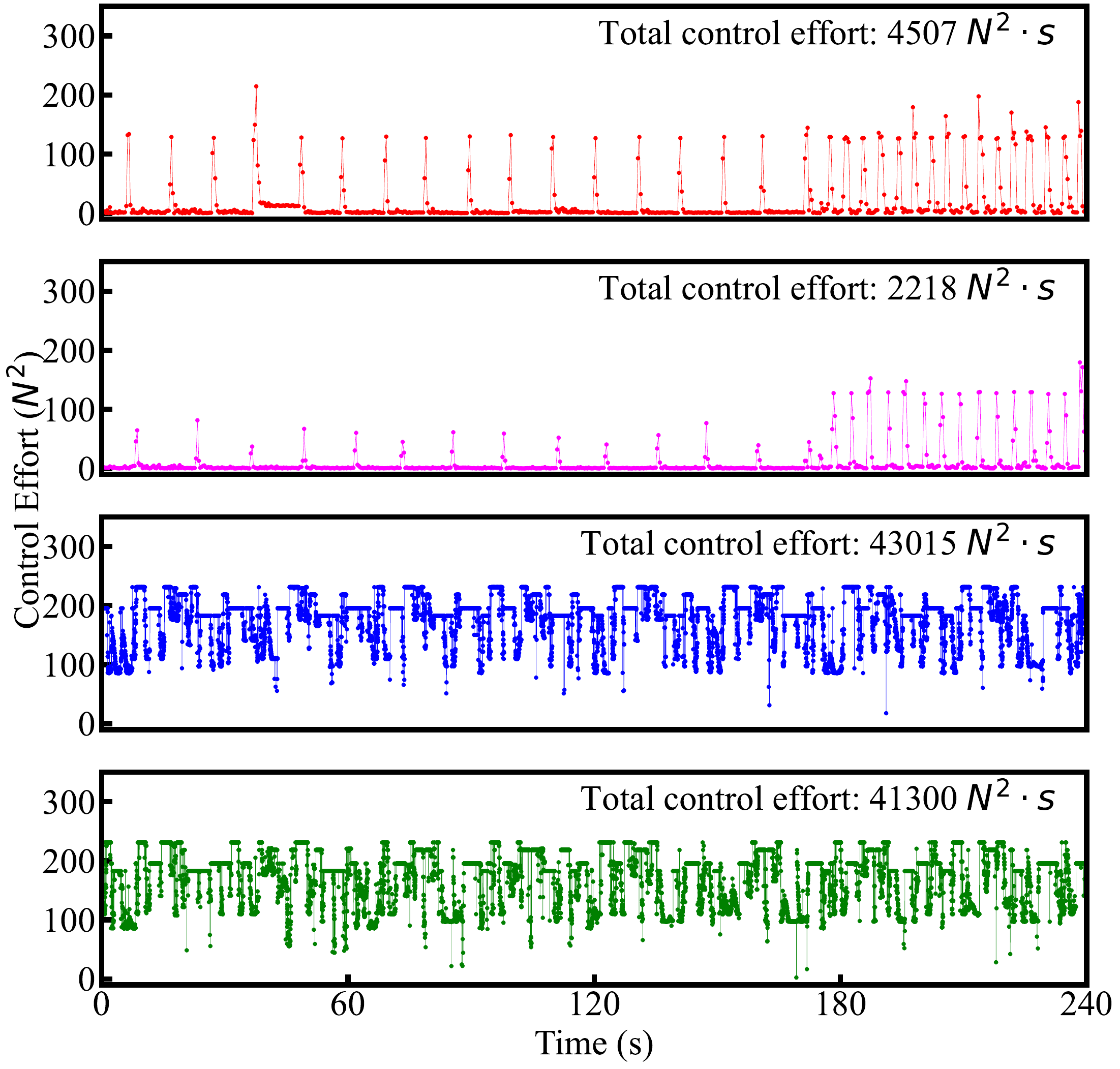}
		\end{minipage}
	}
	\subfloat[]{
		\begin{minipage}[b]{0.3\textwidth}
			\centering
			\includegraphics[width=0.9\textwidth]{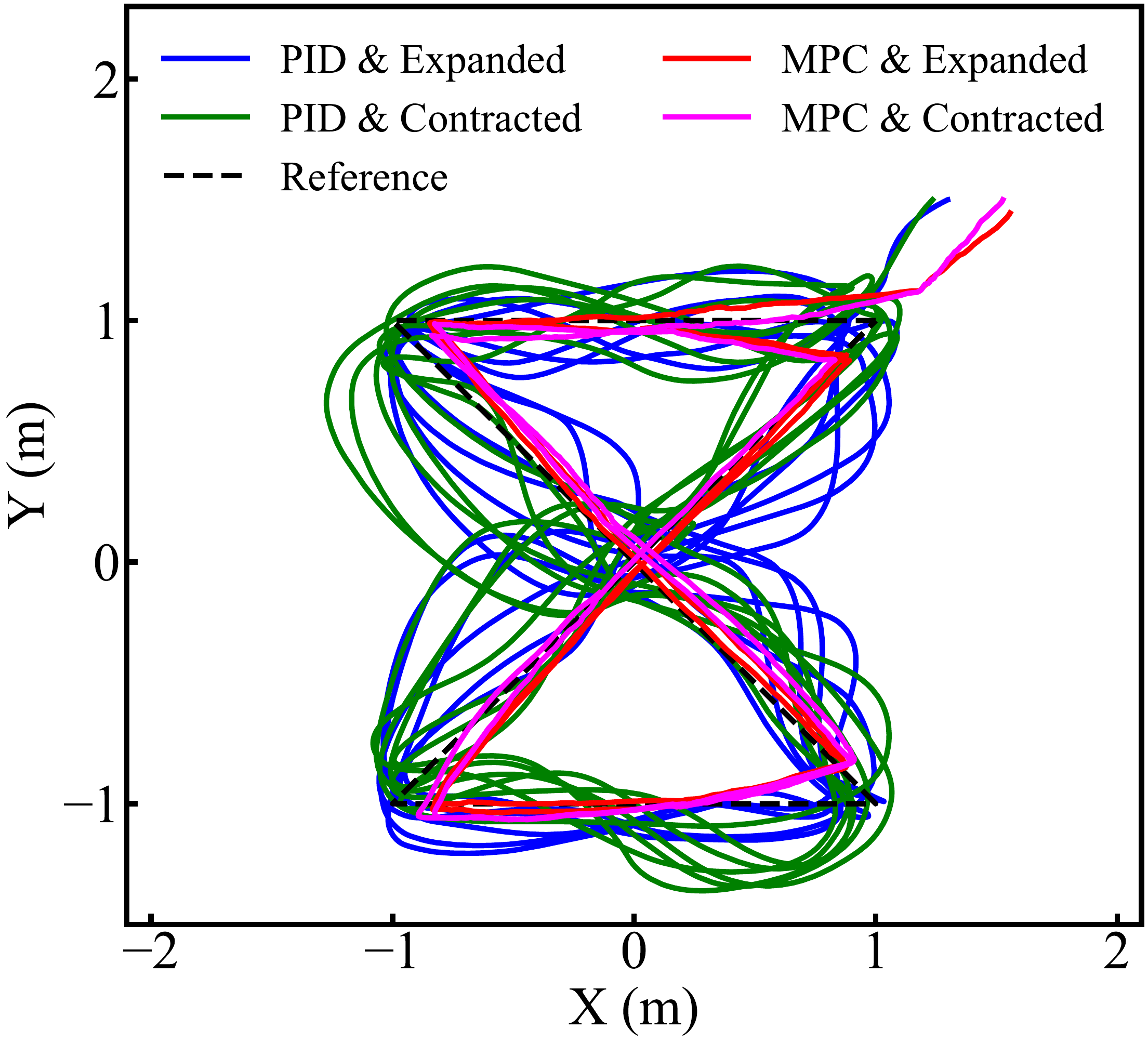}
			\newline
			\includegraphics[width=1\textwidth]{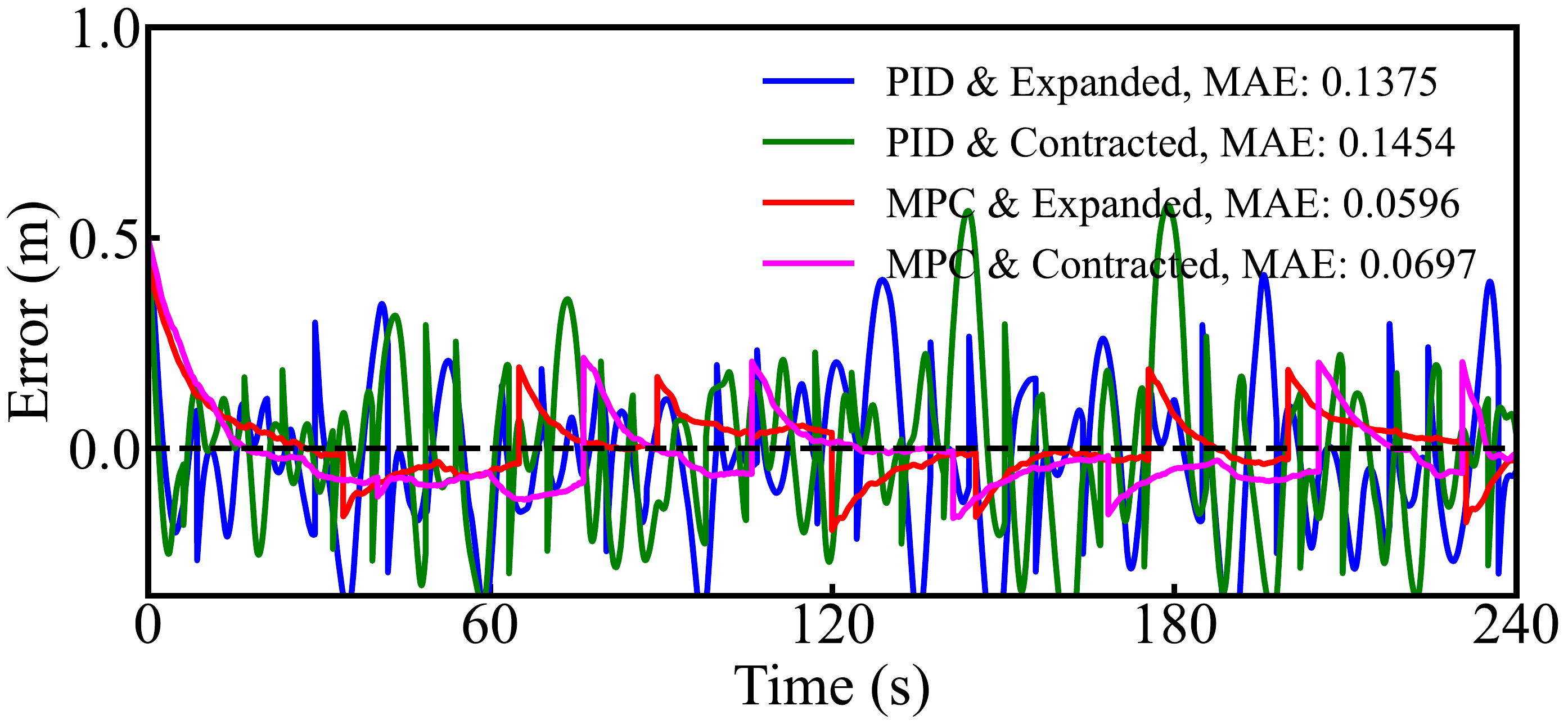}
			\newline
			\includegraphics[width=1\textwidth]{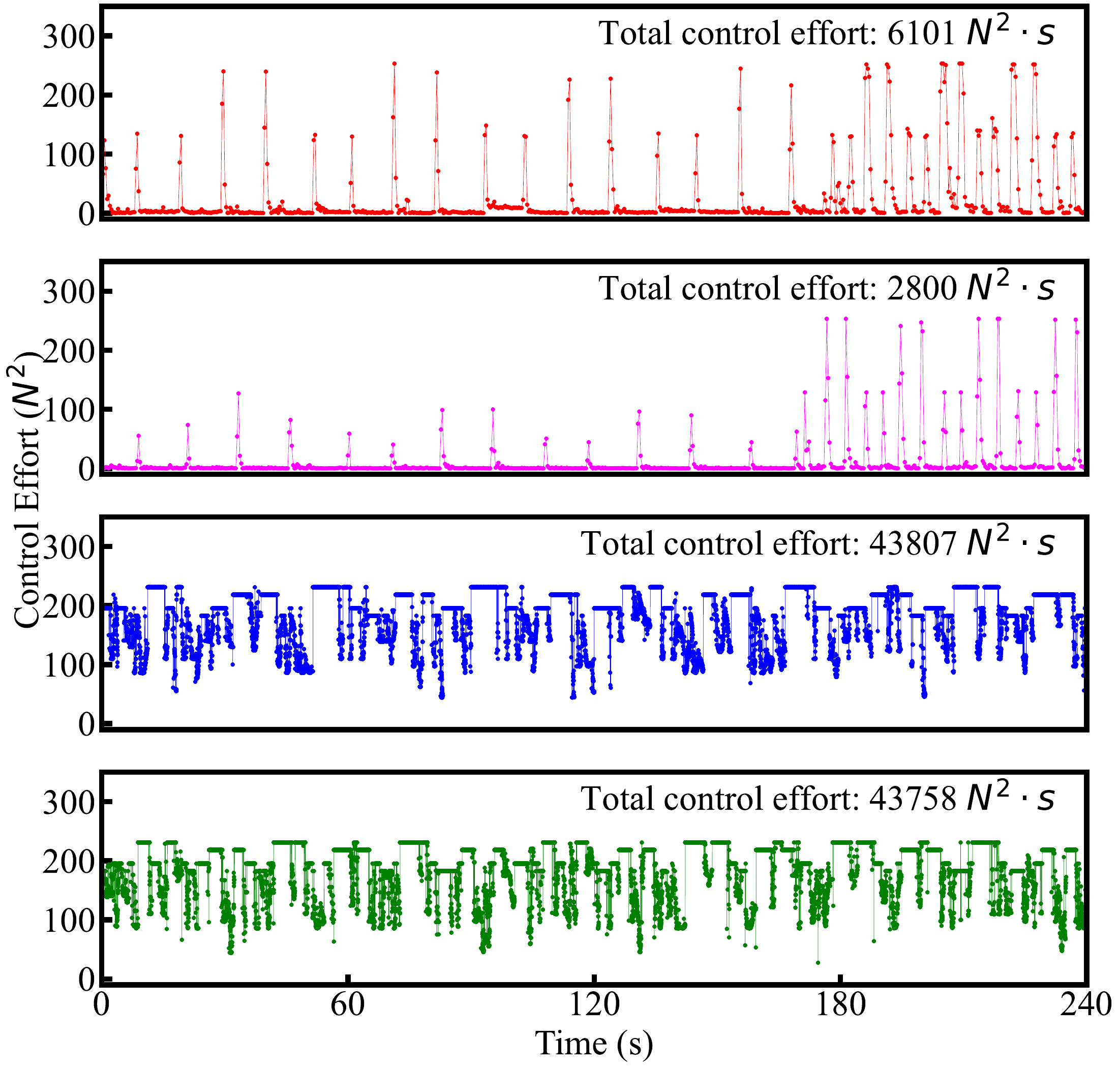}
		\end{minipage}
	}
	\caption{Experimental paths, errors, and control effort in tracking (a) circular, (b) square, and (c) hourglass trajectories. The mean absolute errors (MAE) of all trials are calculated. }
	\label{fig:tracking}
\end{figure*}

\section{EXPERIMENTS}
\label{sect:experiment}
In this section, the TransBoat system is first developed. Its key specifications are listed in Table \ref{tab:spec}.
The two extreme forms, contracted and expanded, are chosen to reveal its performance. 
The comparison shows that both the maximum translational and rotational velocities of the contracted form are higher than those of the expanded form, proving that the contracted form has better mobility.

Four experiments (parameter identification, trajectory tracking, docking, and bridge building) are successively conducted in an indoor pool with water dimension of $ 6\ {\rm m} \times 6\ {\rm m} \times 0.4\ {\rm m}$.
First, in the system identification experiment, the model parameters of the TransBoat dynamics are acquired.
Then, the trajectory tracking experiment proves the effectiveness of the USV model by comparing it with the PID control scheme, which also validates the transportability.
Next, a docking experiment is performed in both calm and turbulent water to test the docking performance of different forms, which is crucial for the pickup and assembly of modules.
Finally, a bridge-building demonstration in turbulent water verifies the system. 

\begin{figure} [htbp] 
	\centering
	\includegraphics[width=1\linewidth]{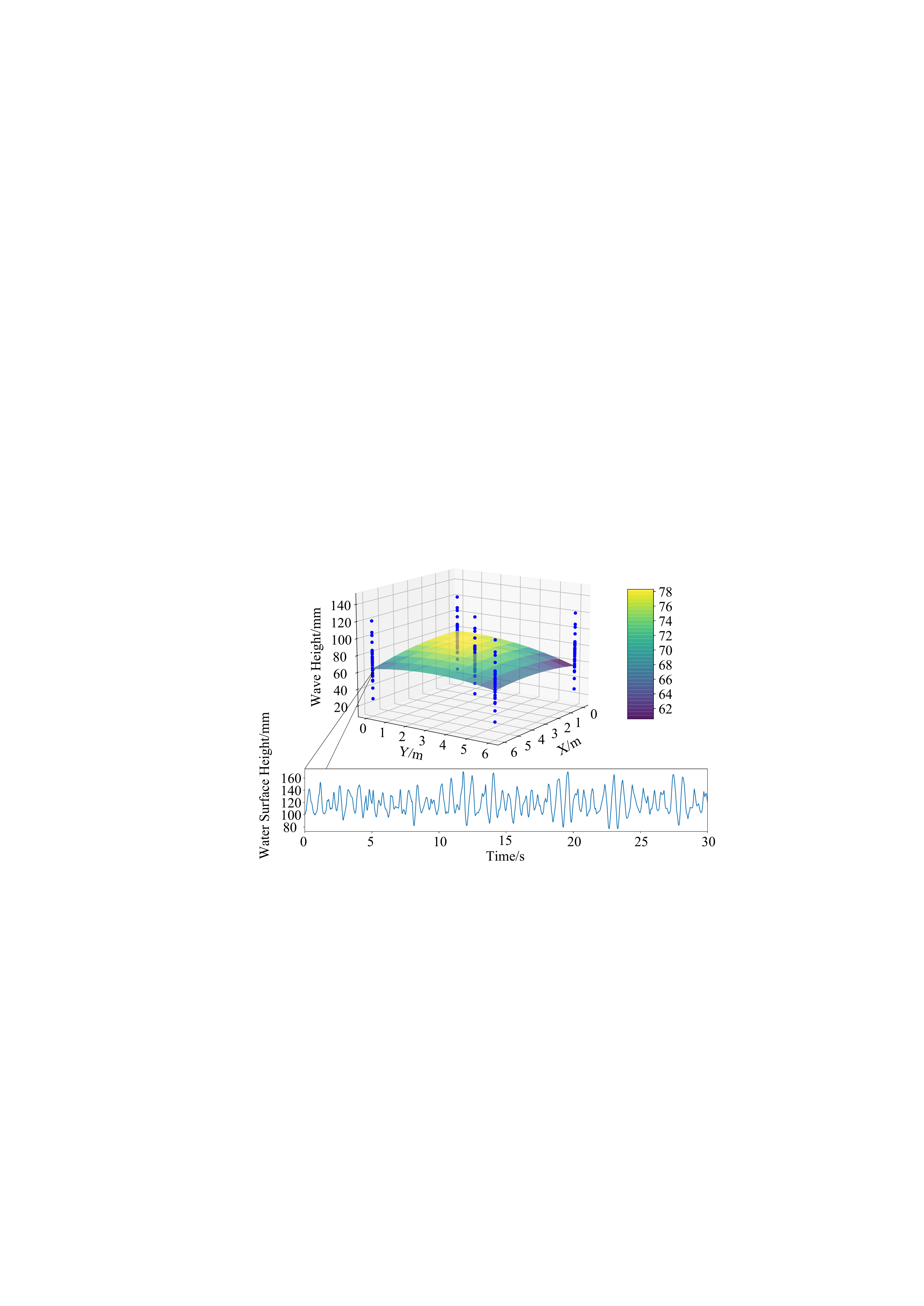}
	\caption{Wave field created by a wave generating machine set up in the pool. }
	\label{fig:wave}
\end{figure}

\begin{figure*} [htbp] 
	\centering
	\includegraphics[width=0.8\linewidth]{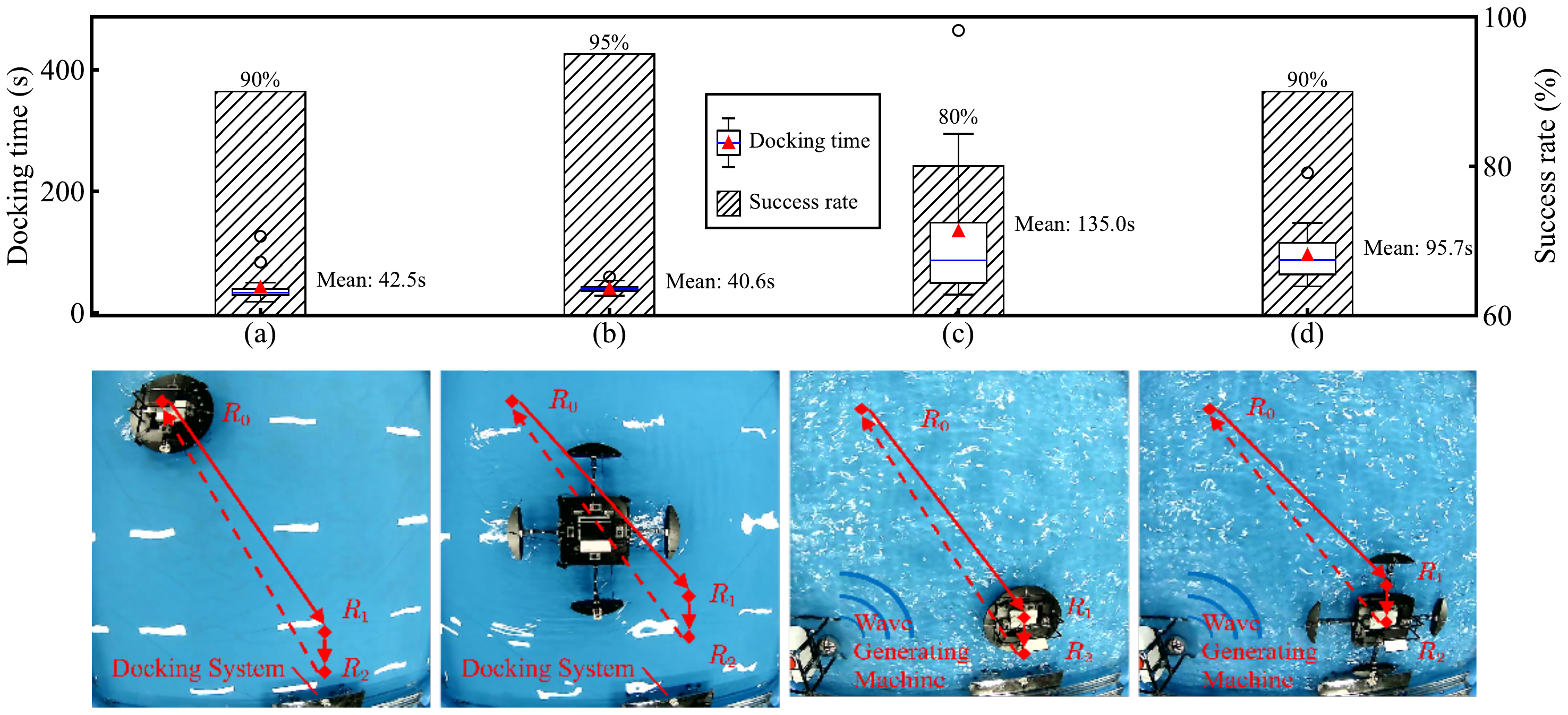}
	\caption{Top: Box plots of the success rates and consumed docking times in experiments (a) $\sim$ (d). Bottom: Top views of the docking experiments of (a) the contracted form and (b) the expanded form in calm water, and (c) the contracted form and (d) the expanded form in turbulent water.}
	\label{fig:docking_success_rate}
\end{figure*}

\subsection{Parameter Identification}

The identification aims to obtain the parameters in Eq. (\ref{eq:parameters}) for the dynamics and subsequent control of the TransBoat,
which includes two steps: parameter identification and regression, as shown in Fig. \ref{fig:SystemID}.
In the first step, three types of tests are performed to gather motion data covering most motion types of the TransBoat, namely, straight-line running, circling, and spinning movements. 
The simulations of the dynamic model use the same initial states and control commands as the experimental data to iteratively calculate the simulated paths by presetting the model parameters.
The velocity errors $\varepsilon$ between the simulated and experimental paths vary with different parameter values, and can be minimized by solving the optimization problem (\ref{eq:identification}), as exhibited in Fig. \ref{fig:parameters} (a) $\sim$ (f).
In this paper, due to the TransBoat's structural symmetry, the parameters of the X- and Y- axes are supposed to be correspondingly equal, that is, $m_1 = m_2$ and $X_u = Y_v$.
Then, a trust-region-reflective algorithm is employed to numerically solve (\ref{eq:identification}).
Considering the transformability of the TransBoat, we repeat this test for 6 selected forms (extension from $0\ \rm m$ to $ 0.5\ \rm m$),
each identifying a set of model parameters, as depicted in Fig. \ref{fig:parameters} (g).
In the second step, the parameter functions are fitted with the identified values for the different forms based on the least squares method.
Without loss of accuracy, quadratic polynomial functions are chosen for the regression, the coefficients of which are shown in Table \ref{tab:para_func}.

\begin{table}[tbp]
	\centering
	\caption{COEFFICIENTS OF PARAMETER FUNCTIONS}
	\label{tab:para_func}
	\begin{tabular}{lcccc}
		\toprule[1.5pt]
		      & $m_1, m_2$ & $m_3$ & $X_u, Y_v$  & $N_r$ \\
		\midrule
		$c_2$ & 0.11317070   & 0.13027175  & -1.01327442 & 0.21534853 \\
		
		$c_1$ & 8.13430784  & 1.88878483  & 9.75583033  & 1.19699306 \\
		
		$c_0$ & 22.82839307 & 7.57931479  & 19.79519544 & 2.27478185 \\
		\bottomrule[1.5pt]
	\end{tabular}
\end{table}

Fig. \ref{fig:parameters} exhibits the identification results, from which two points can be summarized.
(1) After the quadratic regression, only a few increases in velocity error occur when we assign the parameter value on the curve back into the dynamic simulation. Therefore, the fitted curves are credible.
(2) With the expansion, almost all of the identified parameters gradually increase, confirming our design intention that the boat becomes more stable with greater hydrodynamic resistance.
With this result, the parameters for each form can be obtained.

\begin{figure} [htpb]
	\centering
	\subfloat[]{
		\centering
		\includegraphics[width=0.4\textwidth]{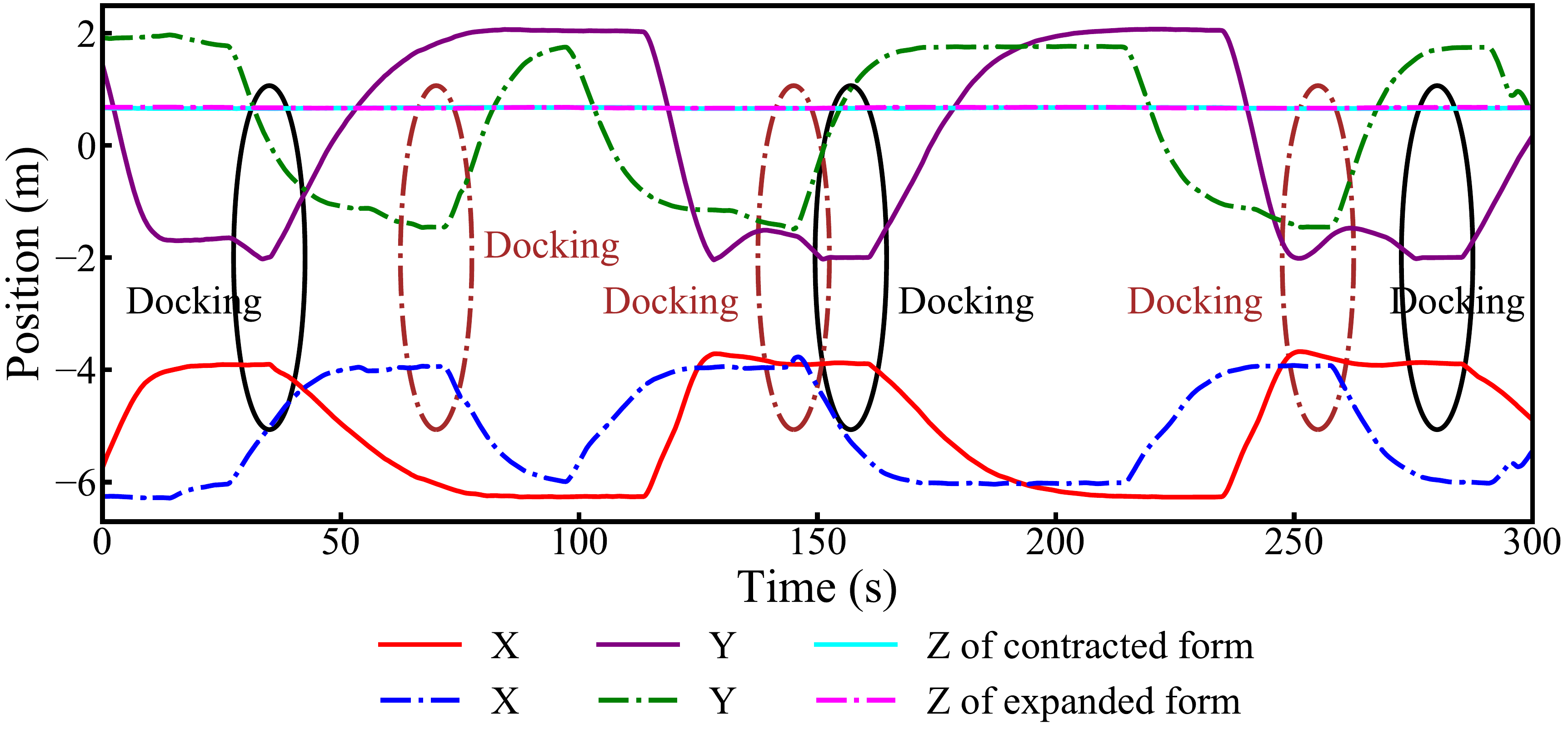}
	}
	
	\subfloat[]{
		\centering
		\includegraphics[width=0.4\textwidth]{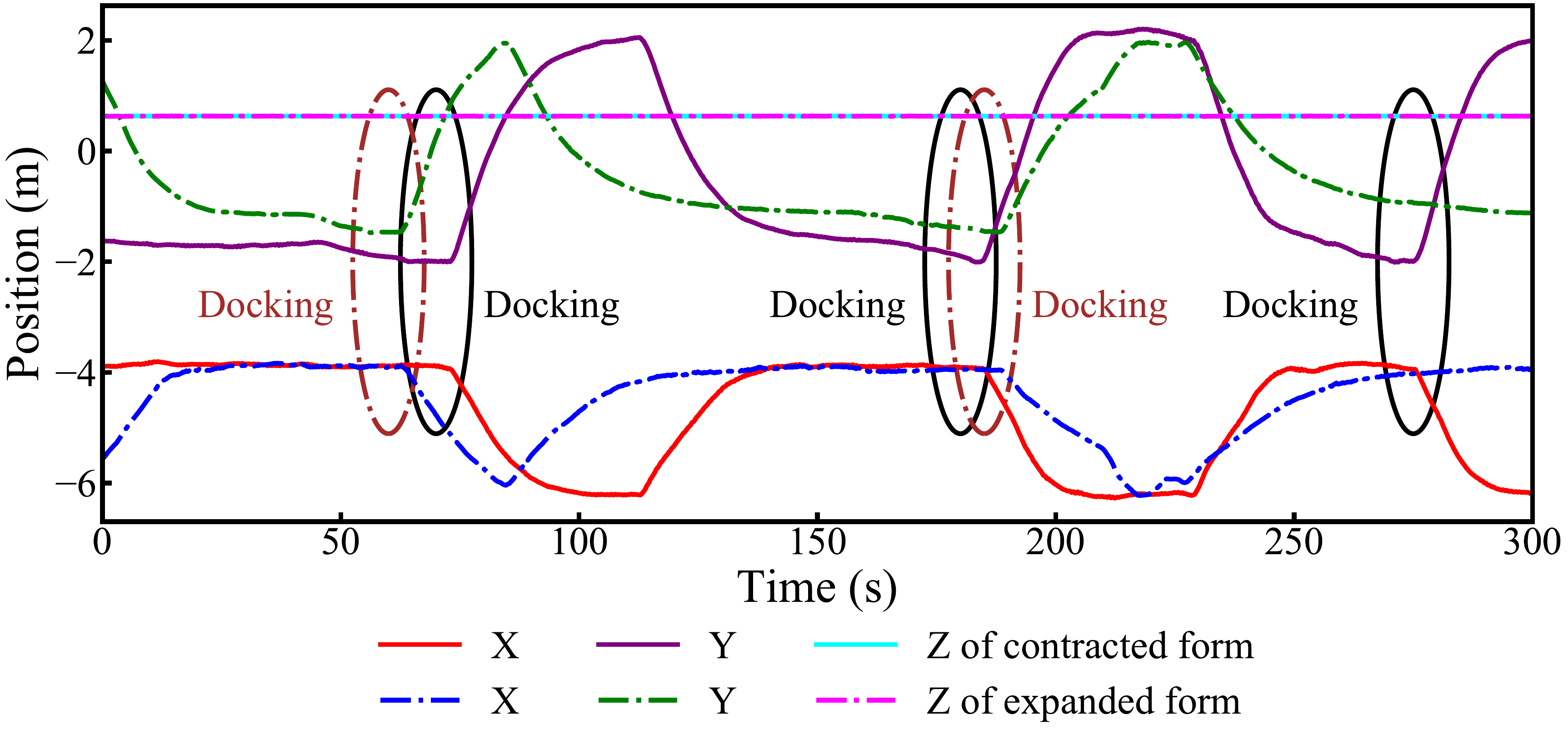}
	}
	\caption{ Experimental results of the positions of the TransBoat in contracted/expanded form performing docking in (a) calm and (b) turbulent water.}
	\label{fig:docking_pos}
\end{figure}

\begin{figure*} [htpb]
	\centering
	\subfloat[]{
		\centering
		\includegraphics[width=0.32\textwidth]{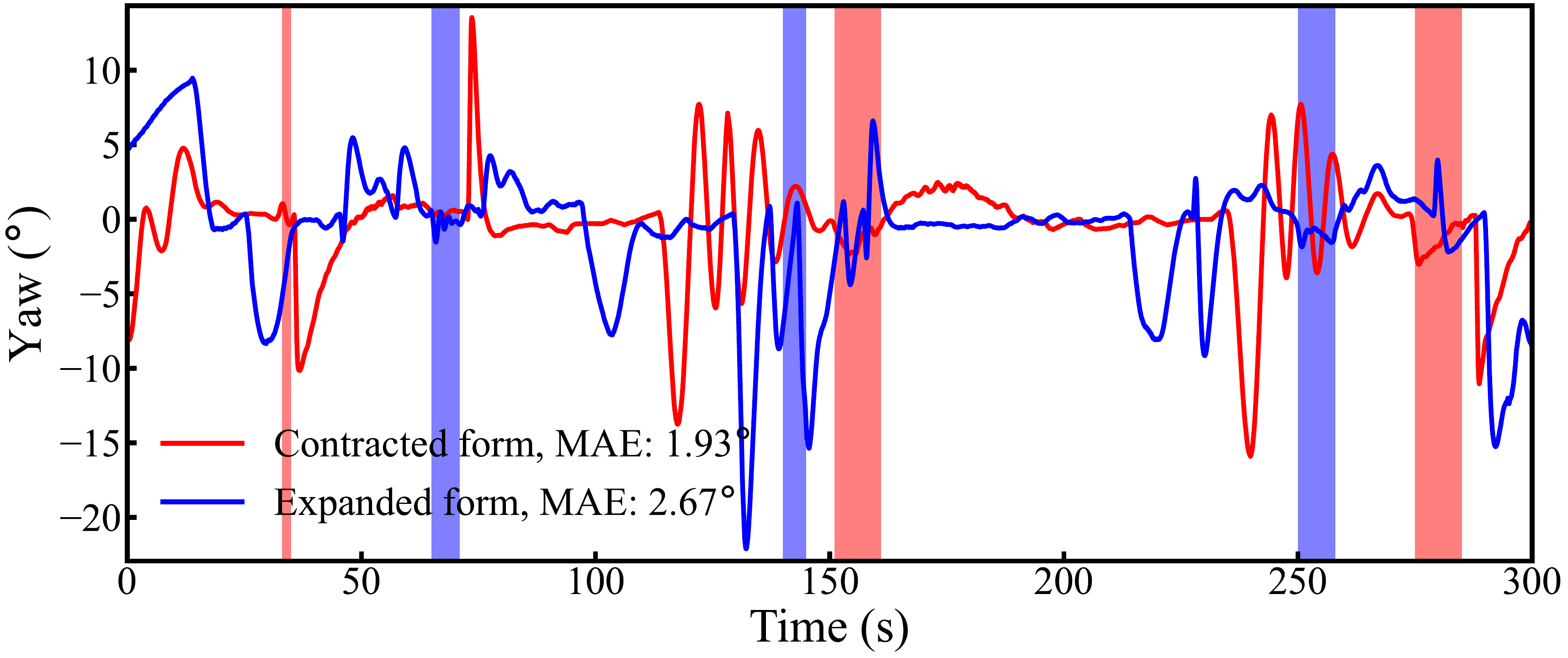}
	}
	\subfloat[]{
		\centering
		\includegraphics[width=0.32\textwidth]{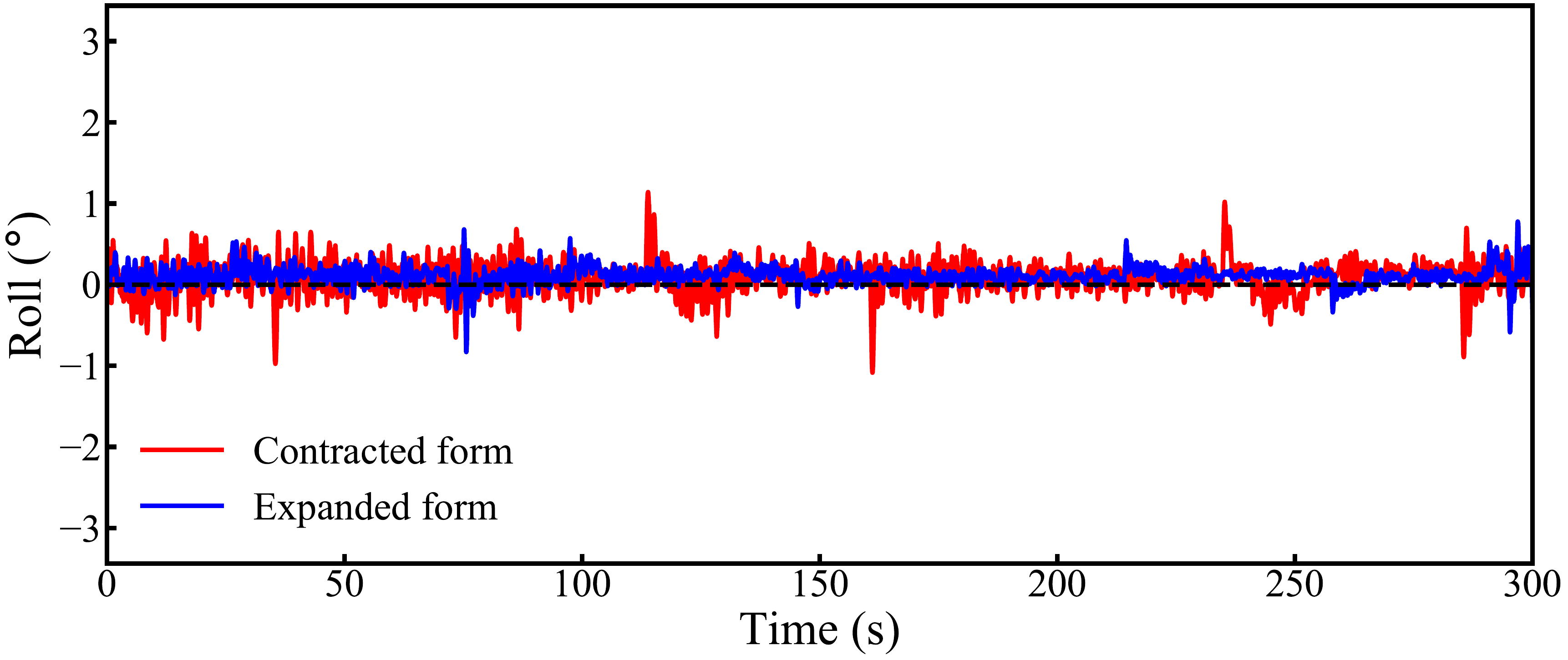}
	}
	\subfloat[]{
		\centering
		\includegraphics[width=0.32\textwidth]{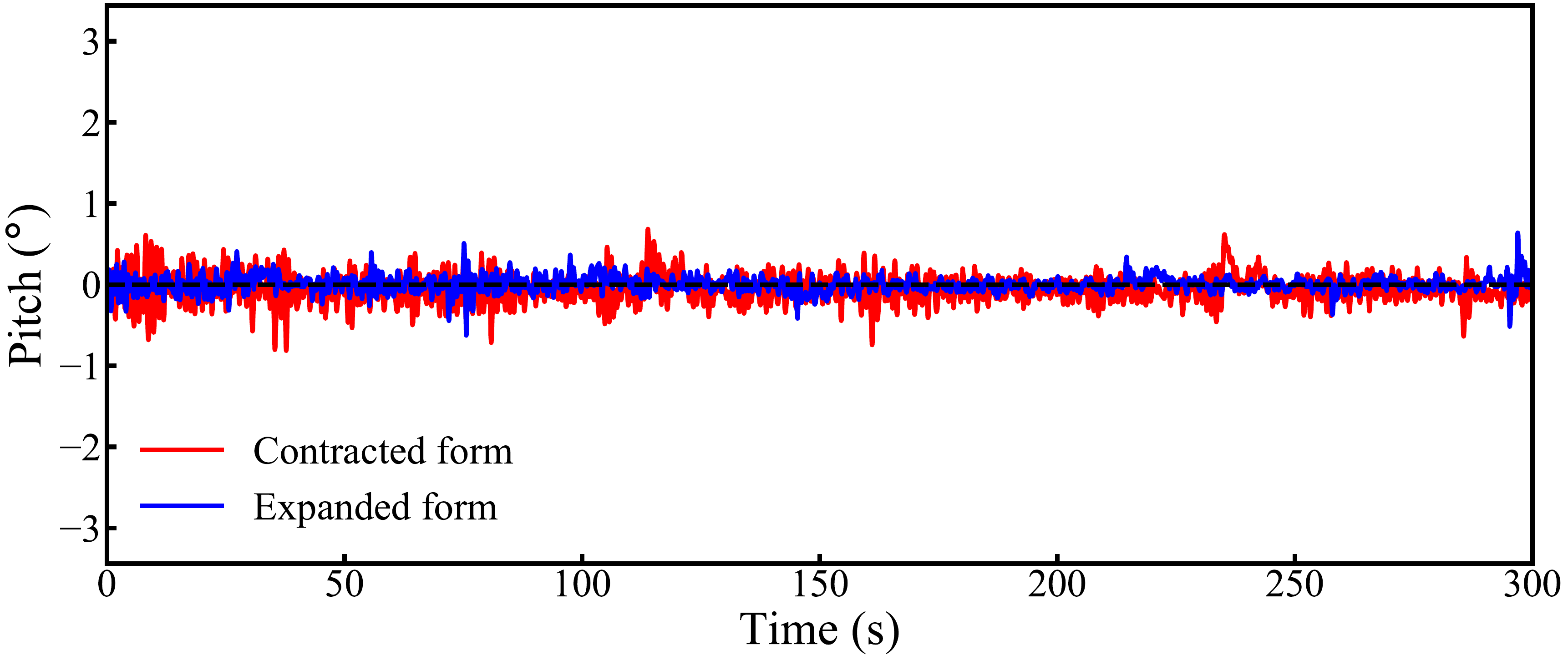}
	}
	\newline
	\subfloat[]{
		\centering
		\includegraphics[width=0.32\textwidth]{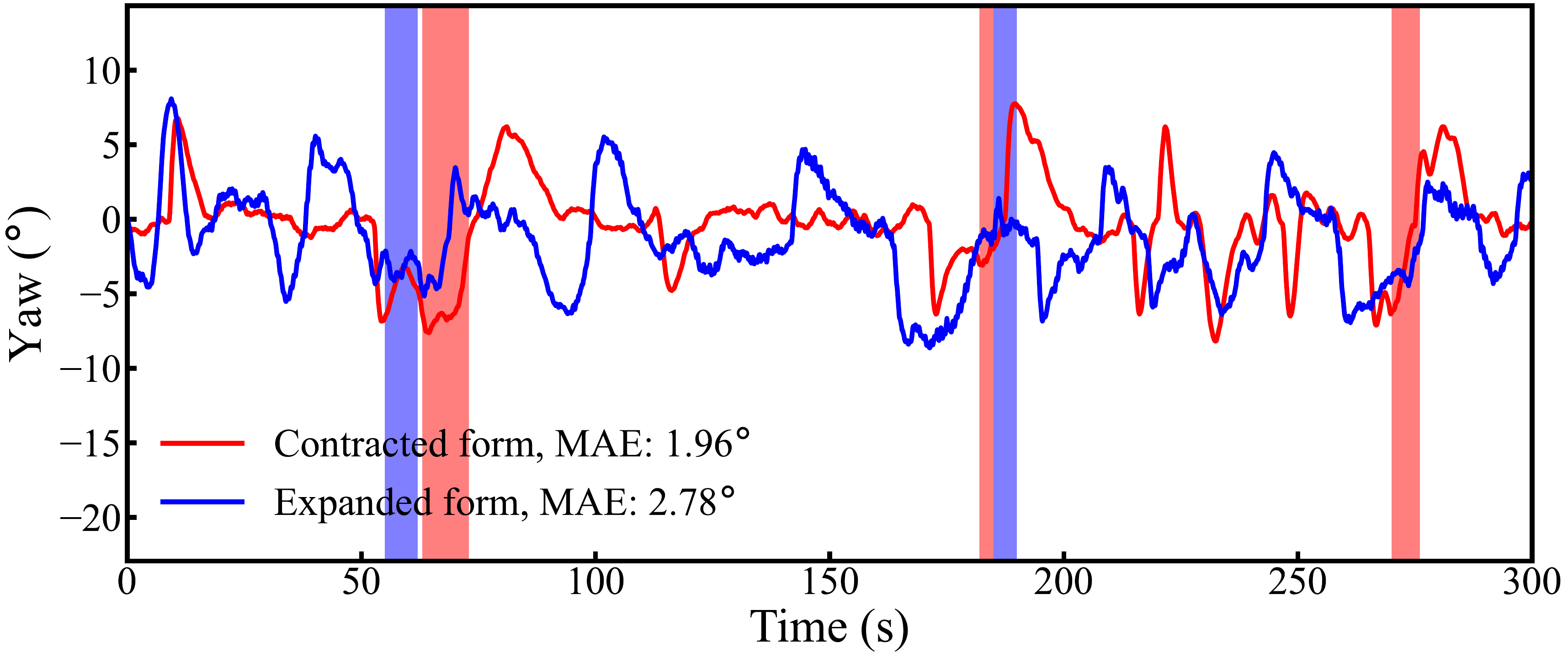}
	}
	\subfloat[]{
		\centering
		\includegraphics[width=0.32\textwidth]{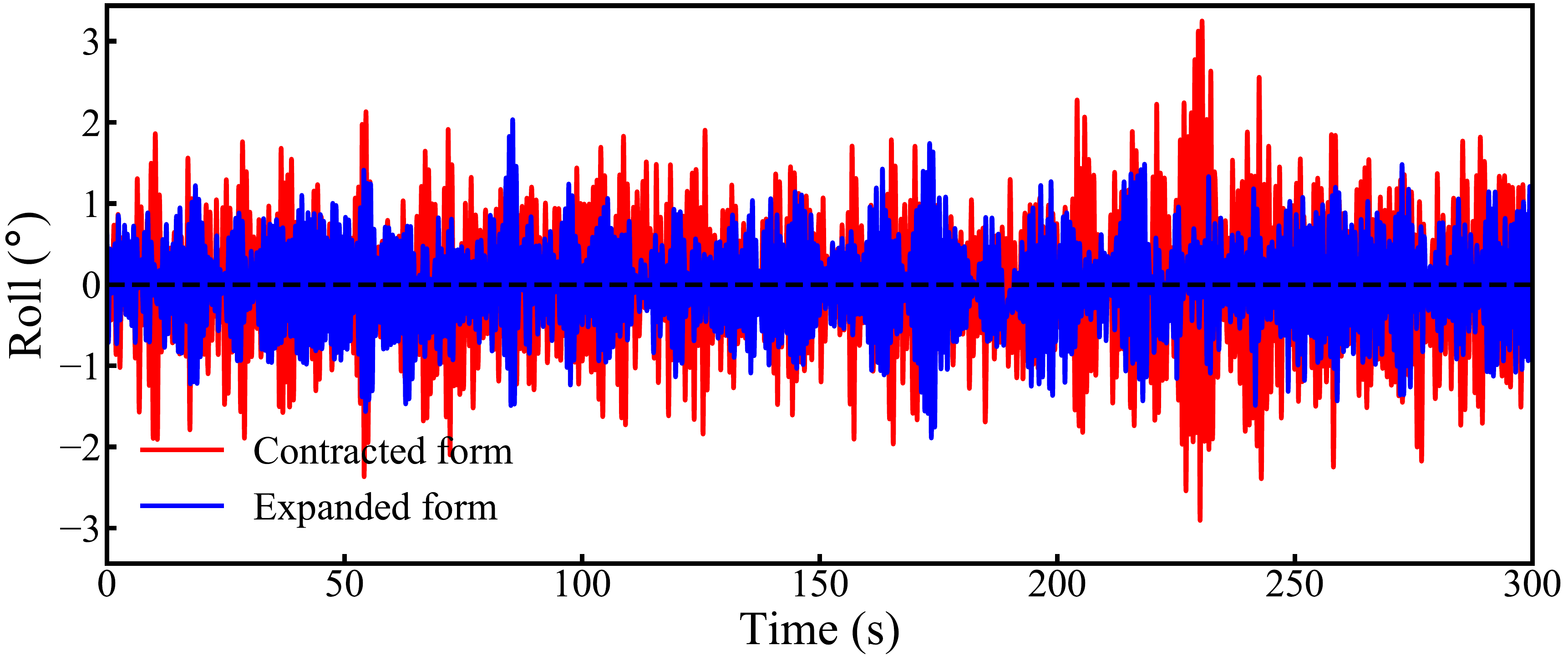}
	}
	\subfloat[]{
		\centering
		\includegraphics[width=0.32\textwidth]{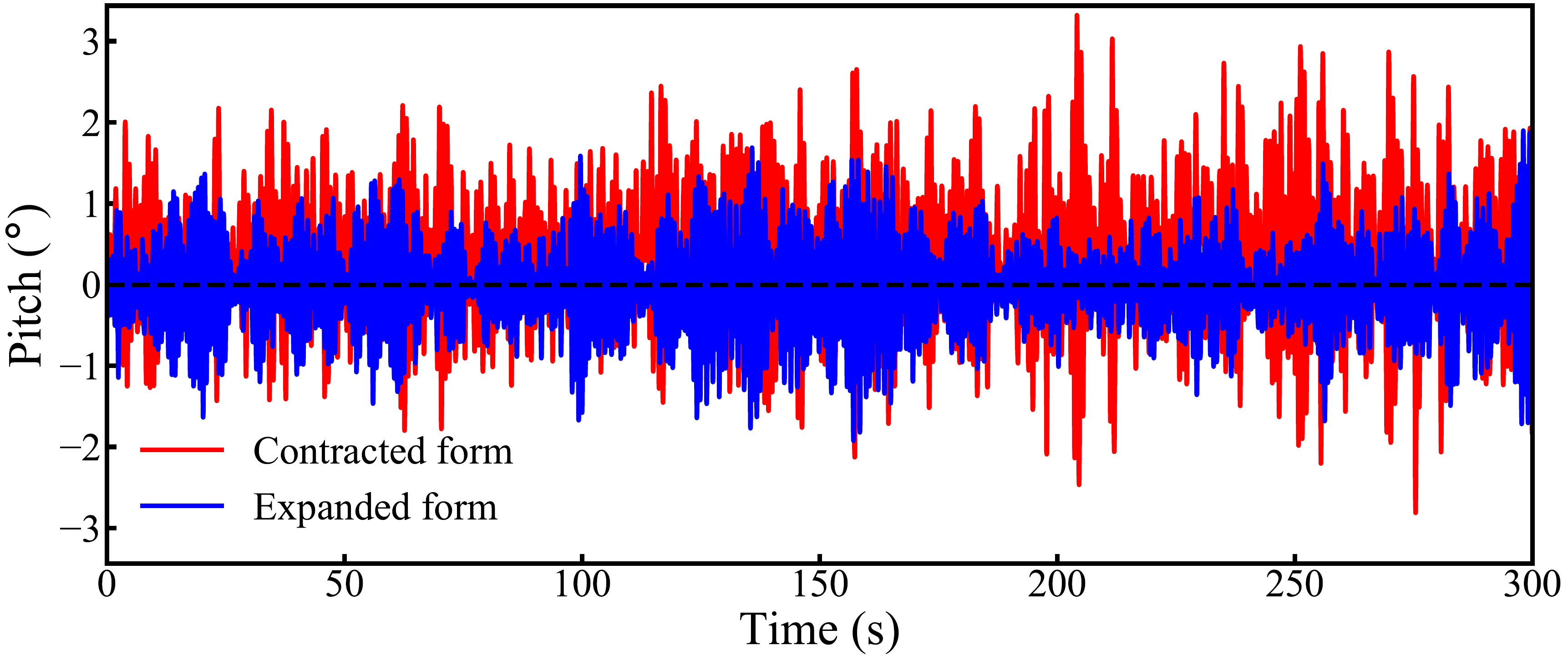}
	}
	\caption{Experimental results of the orientations of the TransBoat in contracted/expanded form performing docking in (a) (b) (c) calm and (d) (e) (f) turbulent water. The red and blue shadows in (a) and (d) represent the docking periods of compacted and extended forms, respectively.}
	\label{fig:docking_angle}
\end{figure*}

\subsection{Trajectory Tracking}

\begin{table}[tbp]
	\centering
	\caption{ PID CONTROLLER GAINS}
	\label{tab:pid_gains}
	\setlength{\tabcolsep}{1.6mm}{
	\begin{tabular}{ccccccc}
		\toprule[1.5pt]
		 & \multicolumn{3}{c|}{Contracted Form}    & \multicolumn{3}{c}{Expanded Form}       \\ \hline
		 & longitudinal & lateral & \multicolumn{1}{c|}{rotational} & longitudinal & lateral & rotational \\ \hline
		$K_P$           & 349     & 349        & 67            & 433             & 349        & 107    \\
		$K_I$           & 1.4     & 1.5        & 1.6           & 1.4             & 1.6        & 1.5    \\
		$K_D$           & 0.9     & 1          & 1.1           & 0.9             & 1.1        & 1      \\  
		\bottomrule[1.5pt]
	\end{tabular}}
\end{table}

Based on the dynamic model identified, the derived NMPC method is implemented for the TransBoat movements.
To authenticate the model and controller, three trajectory tracking tests are conducted, in which the trajectories embody a circle, square, and hourglass comprising two touching triangles.
Each test involves both the contracted and expanded forms of the TransBoat and each is controlled by both the fine-tuned PID and NMPC methods.
We employed the Ziegler–Nichols method \cite{654876}, which is a standard approach to tuning the PID controller. Its major procedure is first zeroing the integral and differential gains, then raising the proportional gain until the system is unstable, and finally setting the PID gains as a function of the proportional gain and the frequency of oscillation at the point of instability.
Three independent PID controllers are tuned according to the Ziegler–Nichols method and used to control the longitudinal, lateral, and rotational motions, respectively, of which the PID gains ($K_P, K_I, K_D$) are exhibited in Table \ref{tab:pid_gains}.

Fig. \ref{fig:tracking} portrays the tracking errors and the control effort $\mathbf{u}^{T}\mathbf{u}$, from which we can draw two conclusions.
First, when comparing the errors of the two control schemes, we can see that in the trials of square and hourglass trajectories, the NMPC is notably superior to the PID, although it is inferior in the circular tracking trials. 
In other words, the NMPC, no matter in what form, performs better than the PID when tracking the segmented trajectories in which we are interested.
Both forms confirm the effectiveness of the identified dynamics in the previous section.
Second, by comparing the total control effort $\int\mathbf{u}^{T}\mathbf{u}dt$ across all trials, we can find that the NMPC consumed less control effort, also regarded as the energy, than the PID, which indicates that the former is more suitable for long-distance movements.
In addition, the TransBoat in the expanded form tracks much more accurately than that in the contracted form, but with higher energy consumption. 
Therefore, the contracted form is more efficient for fast and long-distance movement, while the expanded form is better for precise motion. 

\newcommand{\figsize}{0.1}
\begin{figure*} [htpb]
	\subfloat[]{
		\centering
		\begin{overpic}[width=\figsize\textwidth]{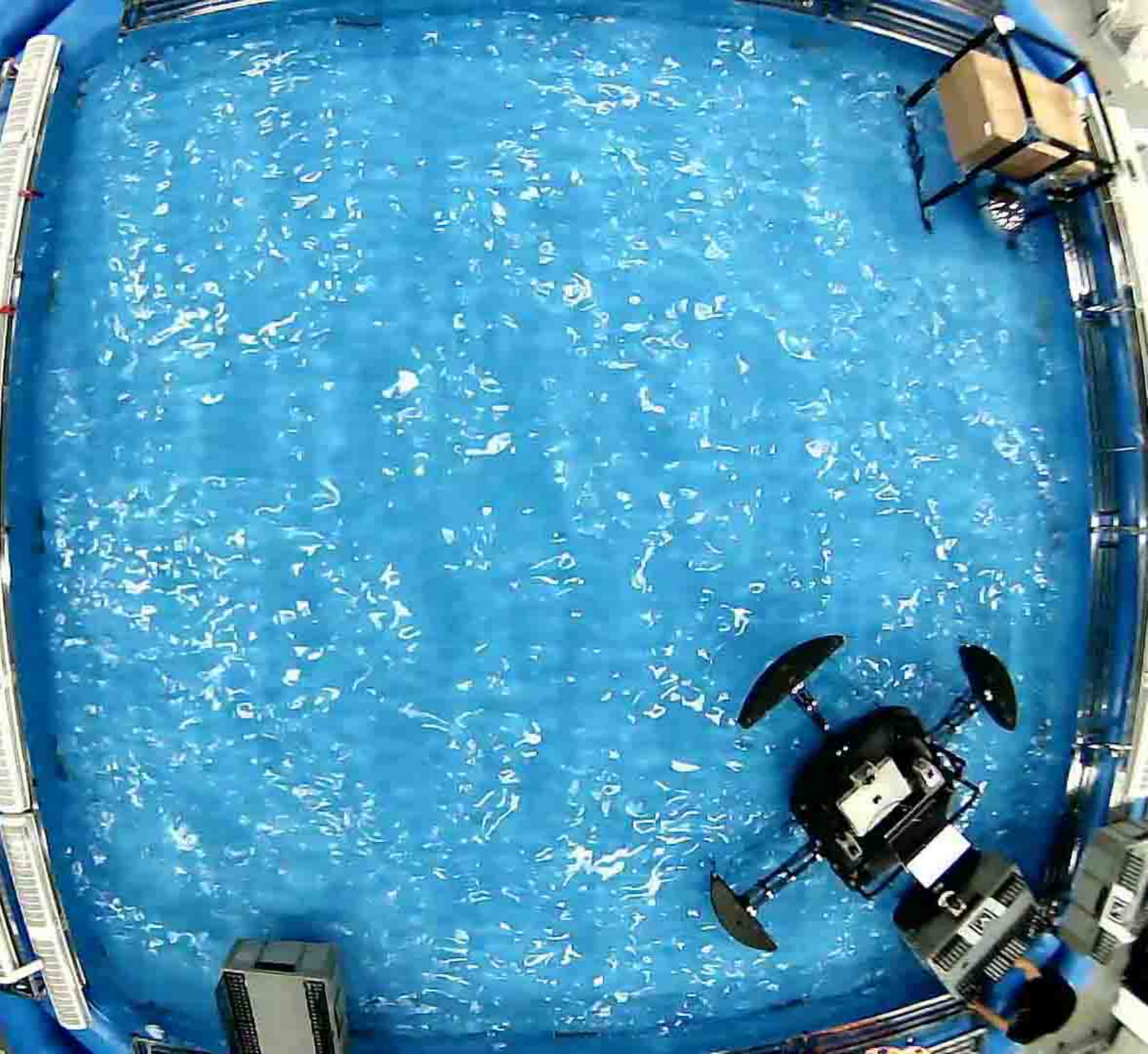}
			\put(25,95){Pickup}
			\put(-10,35){1}
		\end{overpic}
	}
	\hspace{-5pt}
	\subfloat[]{
		\centering
		\begin{overpic}[width=\figsize\textwidth]{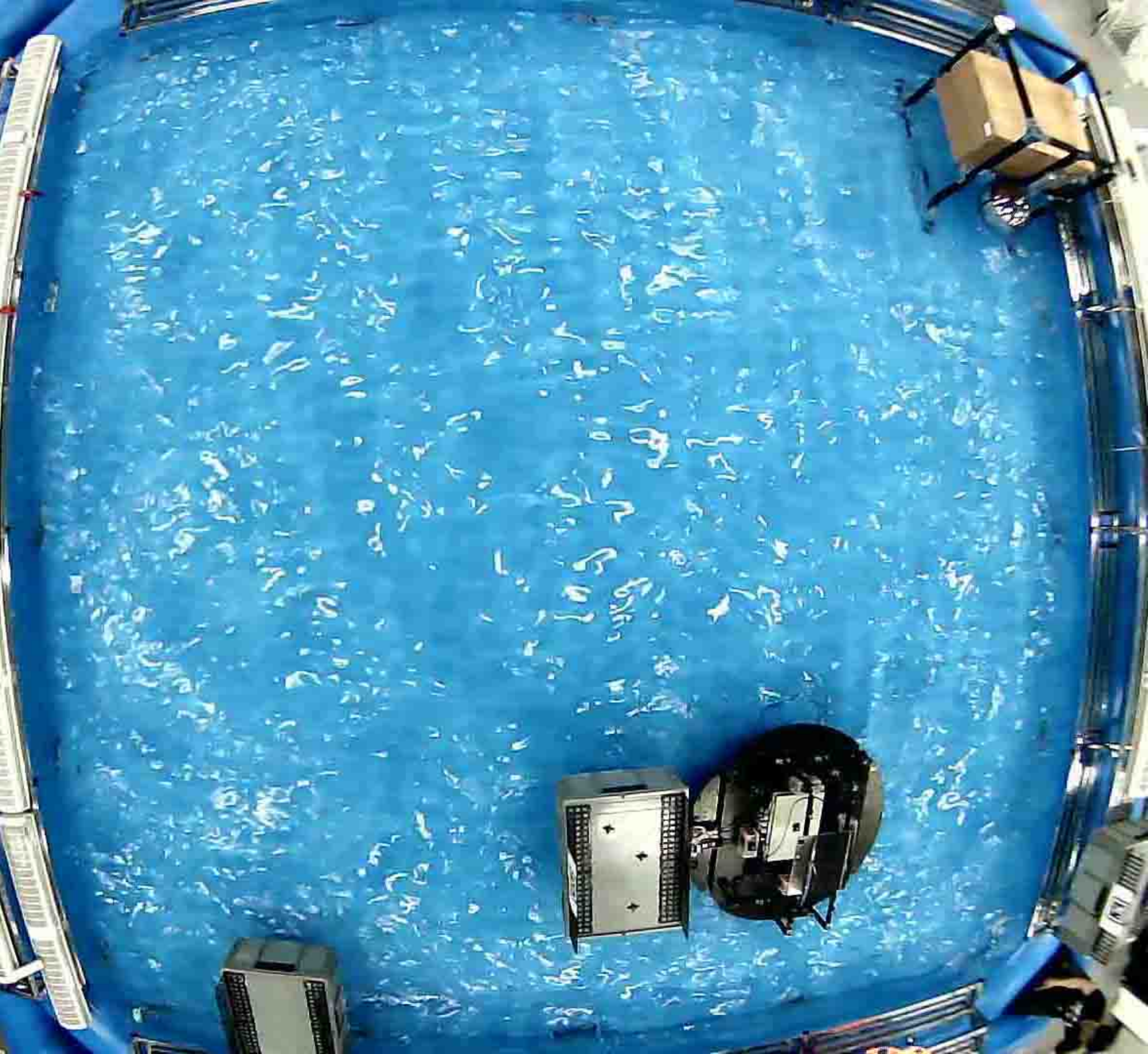}
			\put(20,95){Delivery}
		\end{overpic}
	}
	\hspace{-5pt}
	\subfloat[]{
		\centering
		\begin{overpic}[width=\figsize\textwidth]{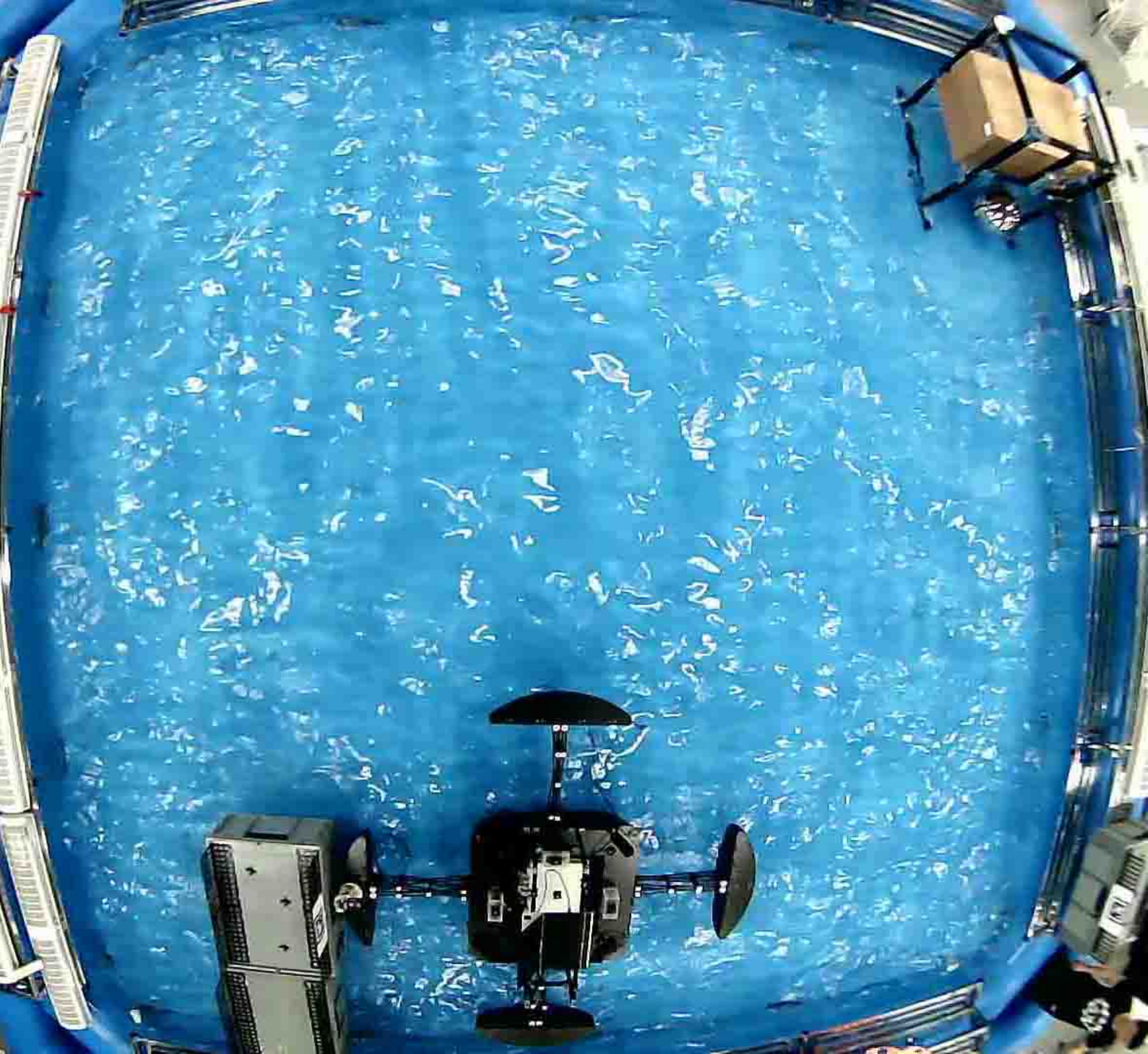}
			\put(15,95){Assembly}
		\end{overpic}
	}
	%\hspace{0.5mm}
	\subfloat[]{
		\centering
		\begin{overpic}[width=\figsize\textwidth]{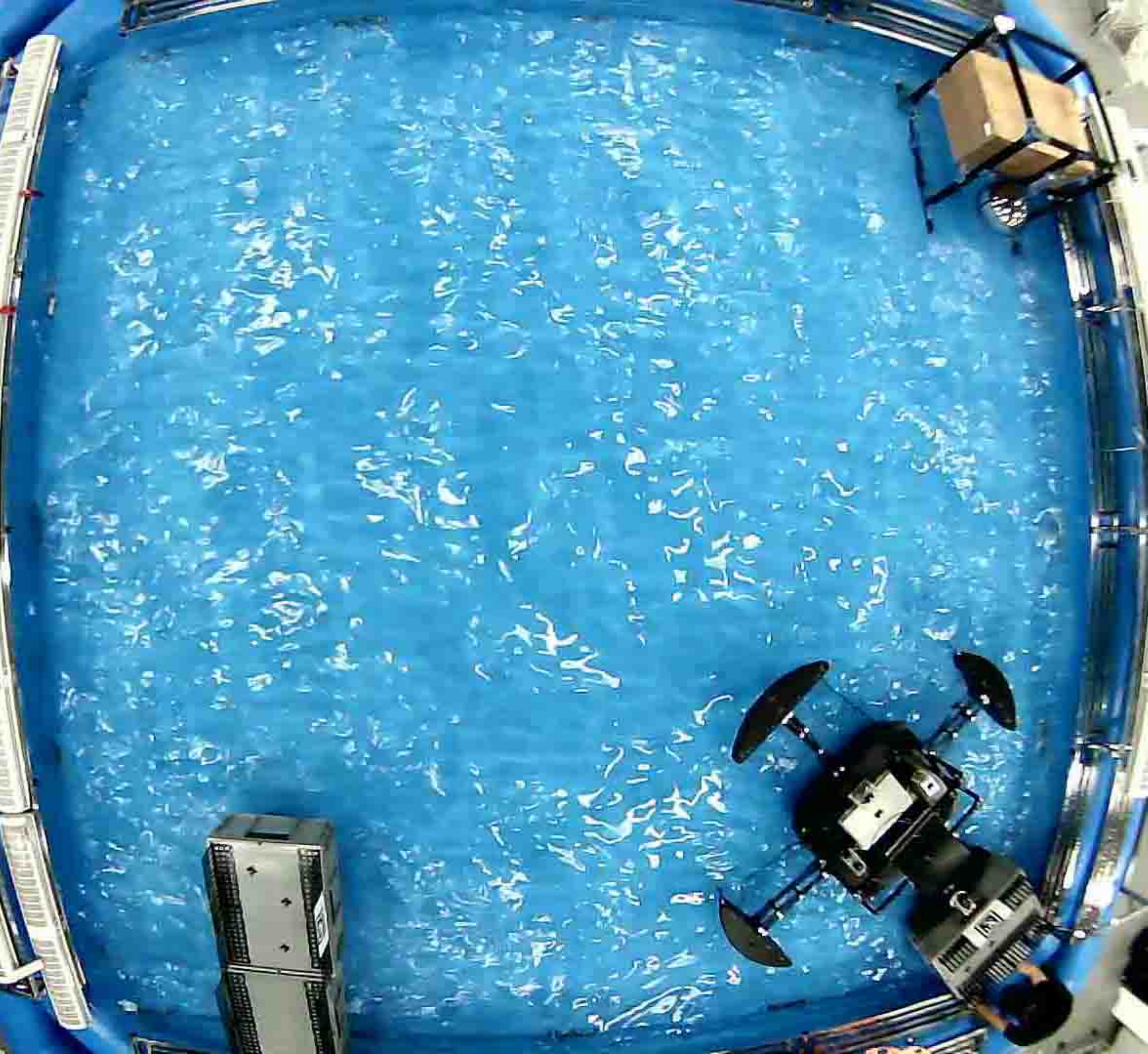}
			\put(25,95){Pickup}
			\put(-10,35){2}
		\end{overpic}
	}
	\hspace{-5pt}
	\subfloat[]{
		\centering
		\begin{overpic}[width=\figsize\textwidth]{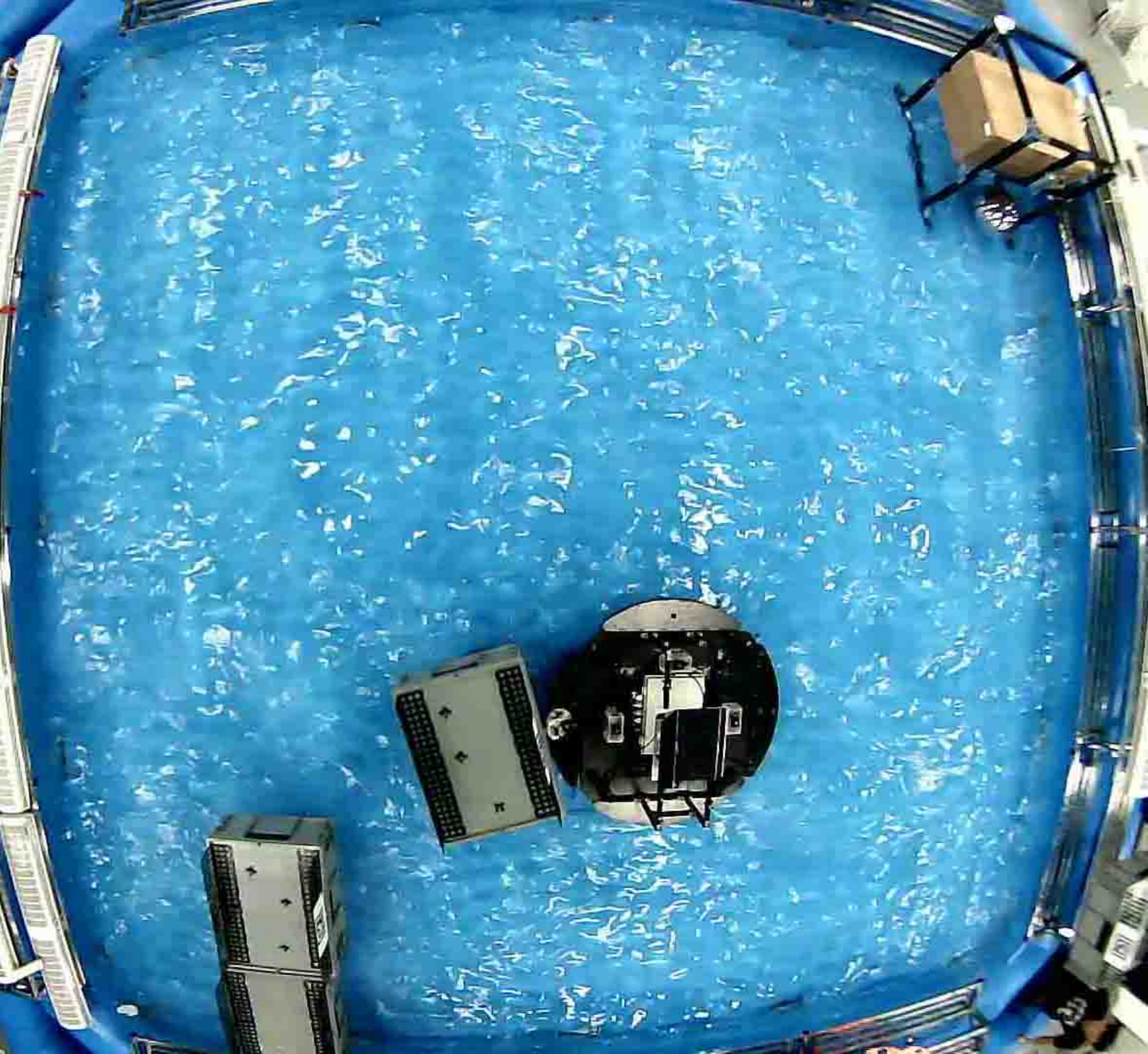}
			\put(20,95){Delivery}
		\end{overpic}
	}
	\hspace{-5pt}
	\subfloat[]{
		\centering
		\begin{overpic}[width=\figsize\textwidth]{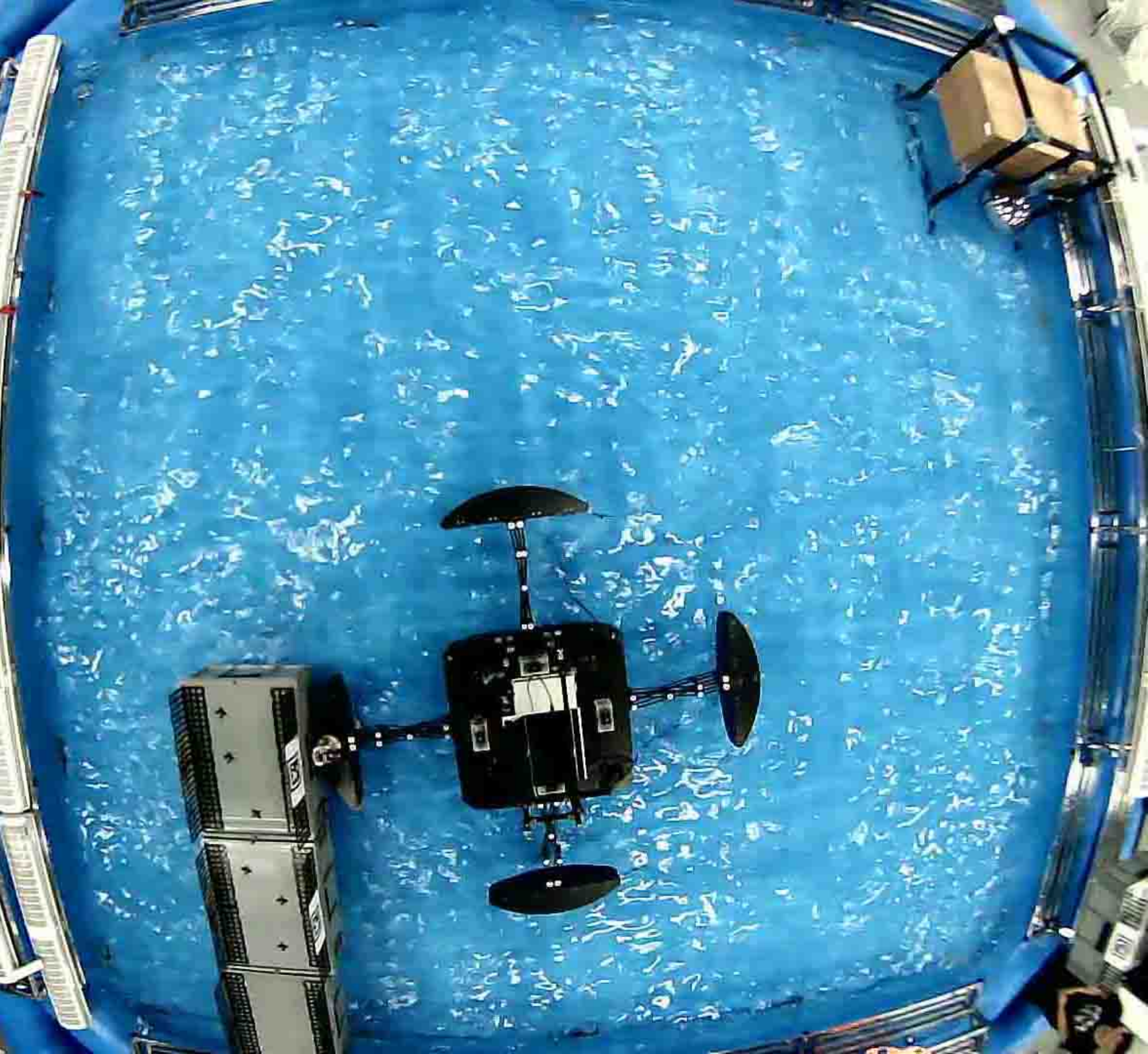}
			\put(15,95){Assembly}
		\end{overpic}
	}
	%\hspace{0.5mm}
	\subfloat[]{
		\centering
		\begin{overpic}[width=\figsize\textwidth]{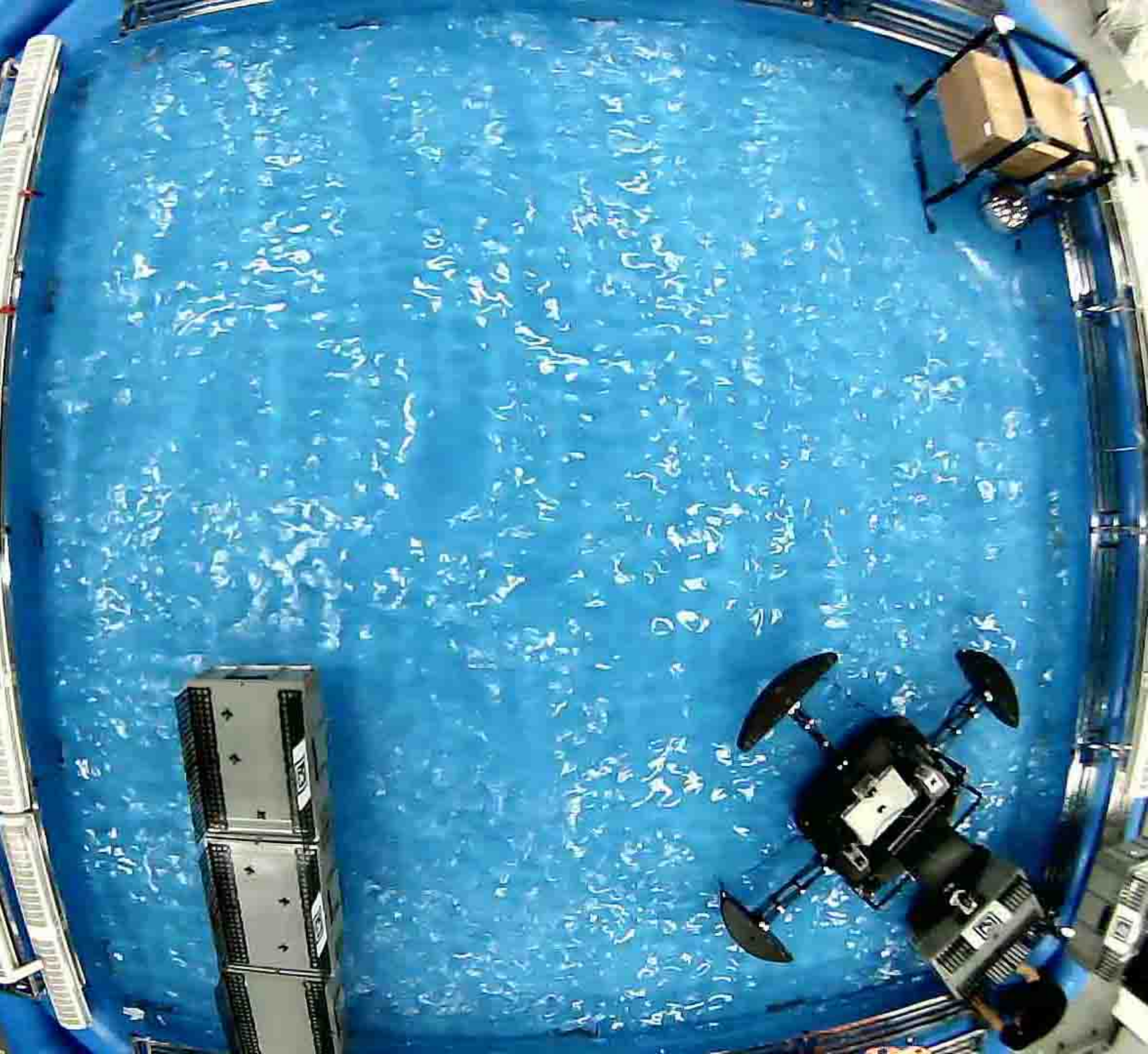}
			\put(25,95){Pickup}
			\put(-10,35){3}
		\end{overpic}
	}
	\hspace{-5pt}
	\subfloat[]{
		\centering
		\begin{overpic}[width=\figsize\textwidth]{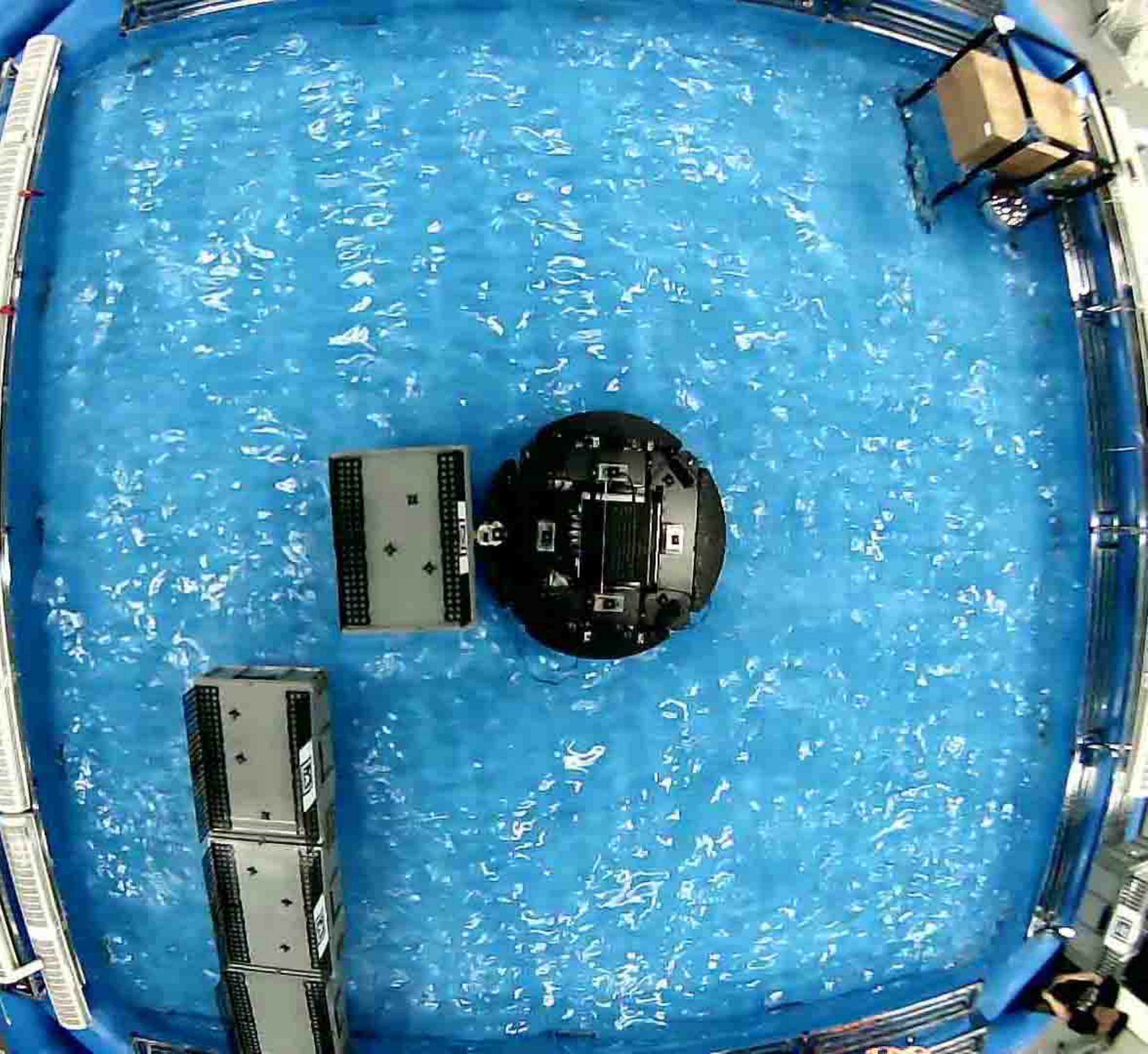}
			\put(20,95){Delivery}
		\end{overpic}
	}
	\hspace{-5pt}
	\subfloat[]{
		\centering
		\begin{overpic}[width=\figsize\textwidth]{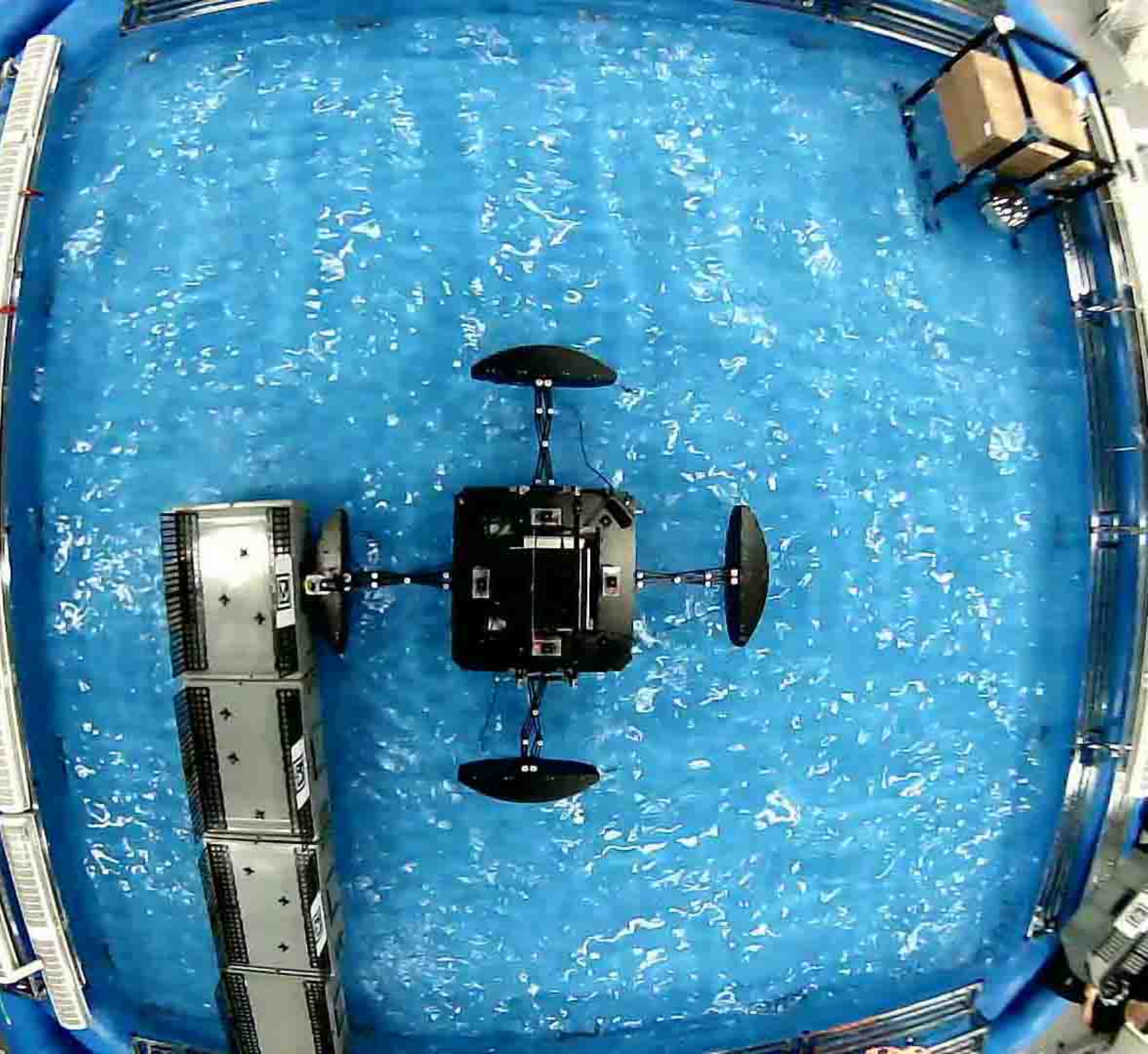}
			\put(15,95){Assembly}
		\end{overpic}
	}
	\hfill
	
	\subfloat[]{
		\centering
		\begin{overpic}[width=\figsize\textwidth]{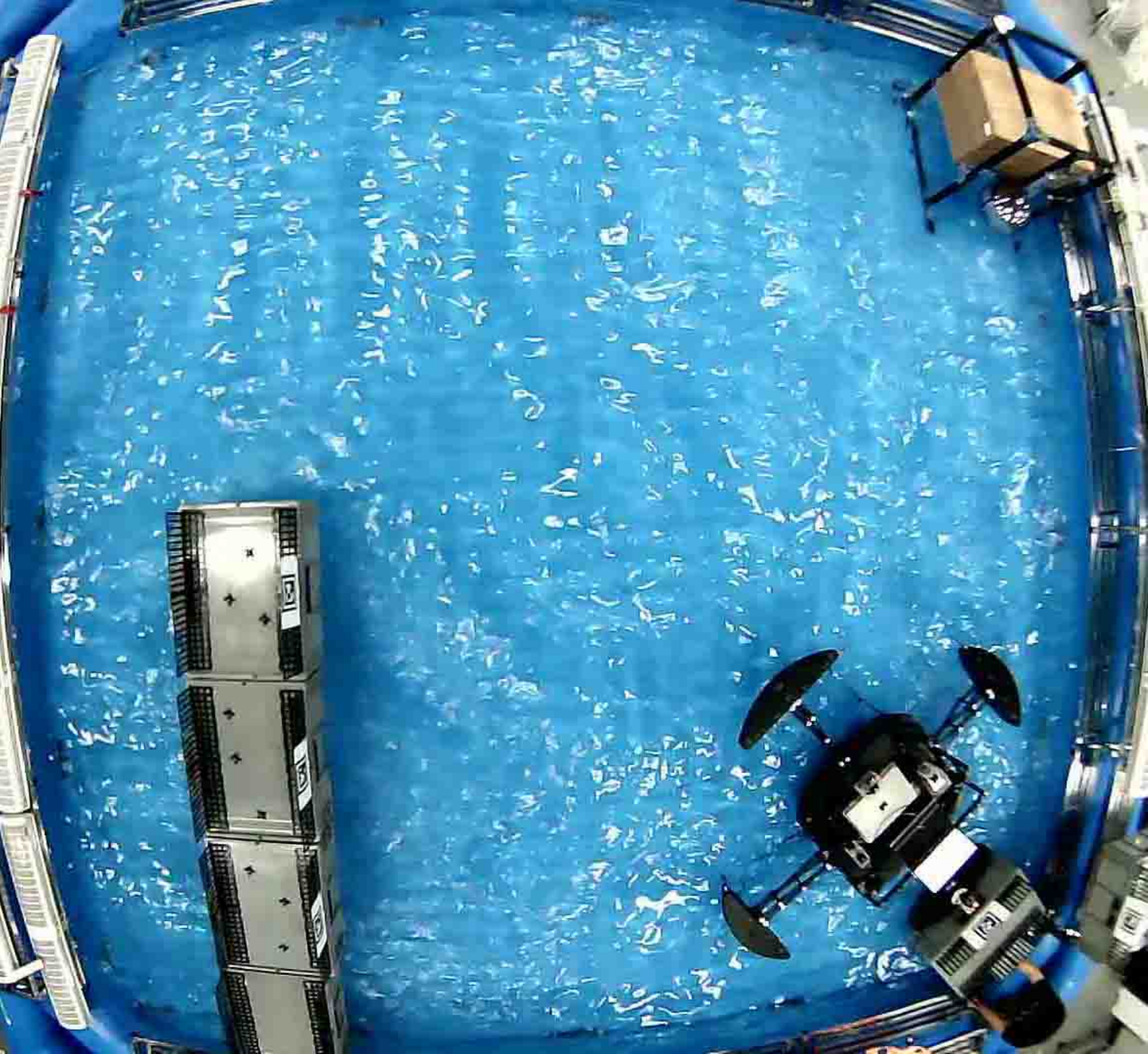}
			\put(-10,35){4}
		\end{overpic}		
	}
	\hspace{-5pt}
	\subfloat[]{
		\centering
		\includegraphics[width=\figsize\textwidth]{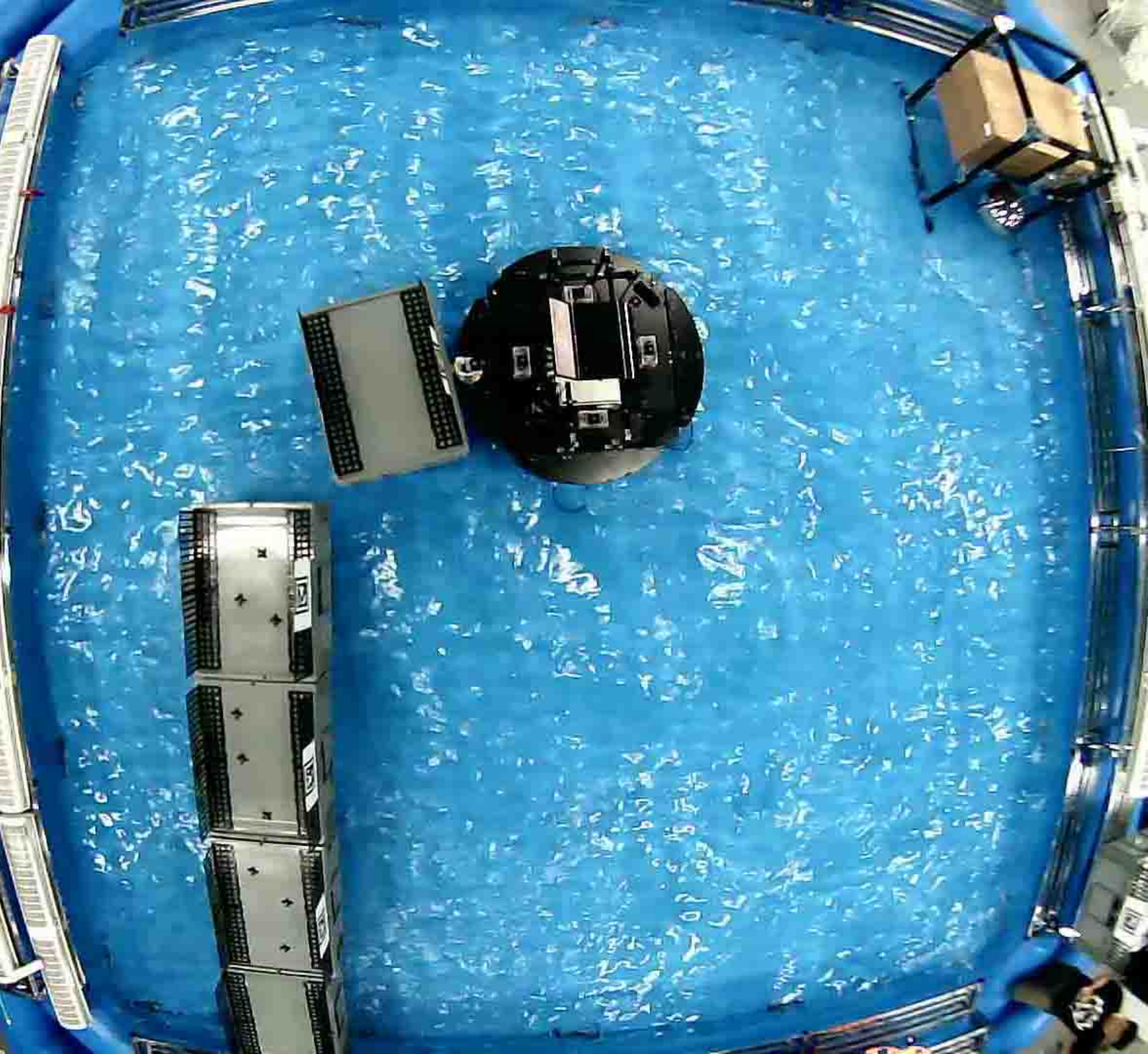}
	}
	\hspace{-5pt}
	\subfloat[]{
		\centering
		\includegraphics[width=\figsize\textwidth]{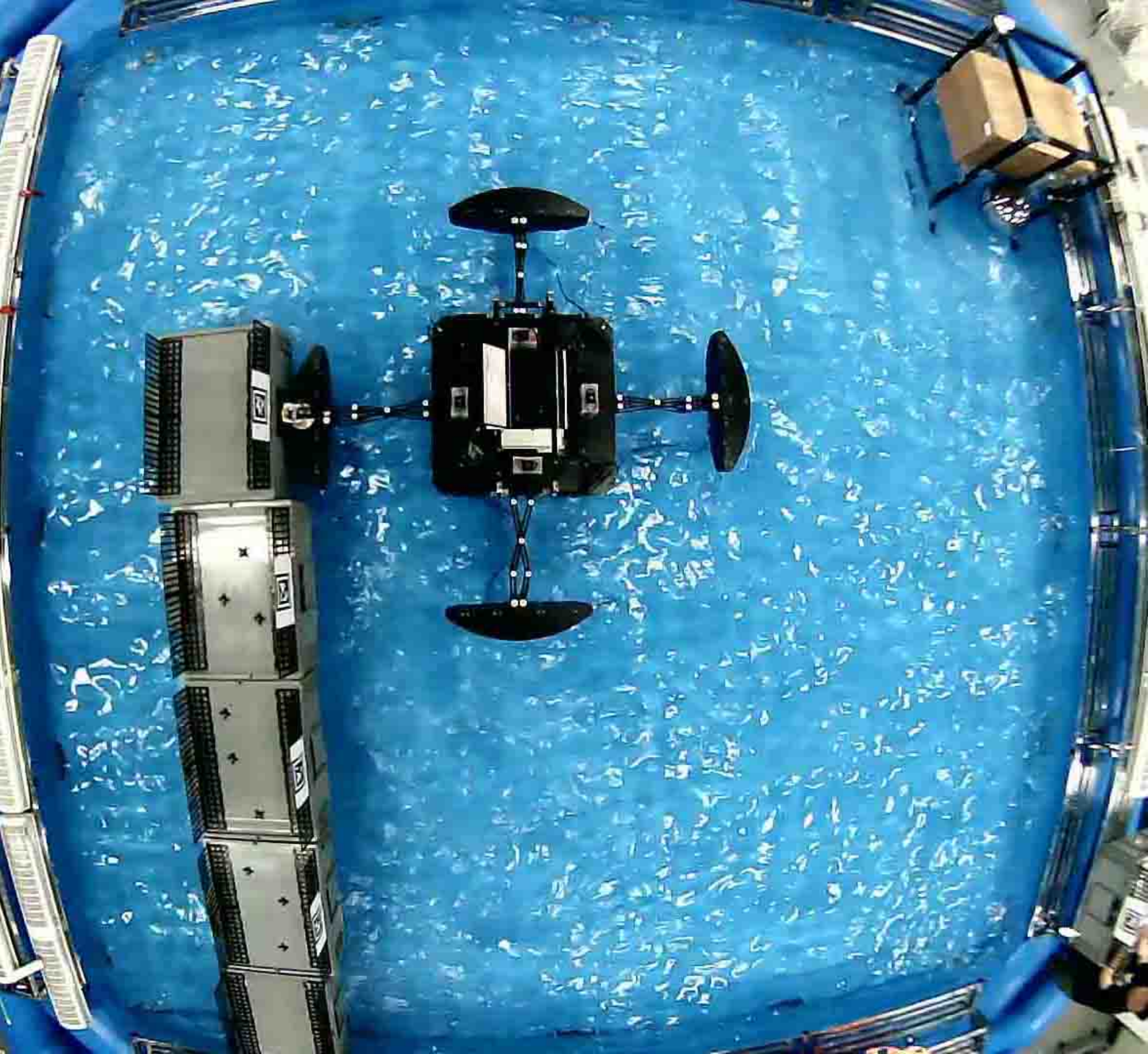}
	}
	%\hspace{0.5mm}
	\subfloat[]{
		\centering
		\begin{overpic}[width=\figsize\textwidth]{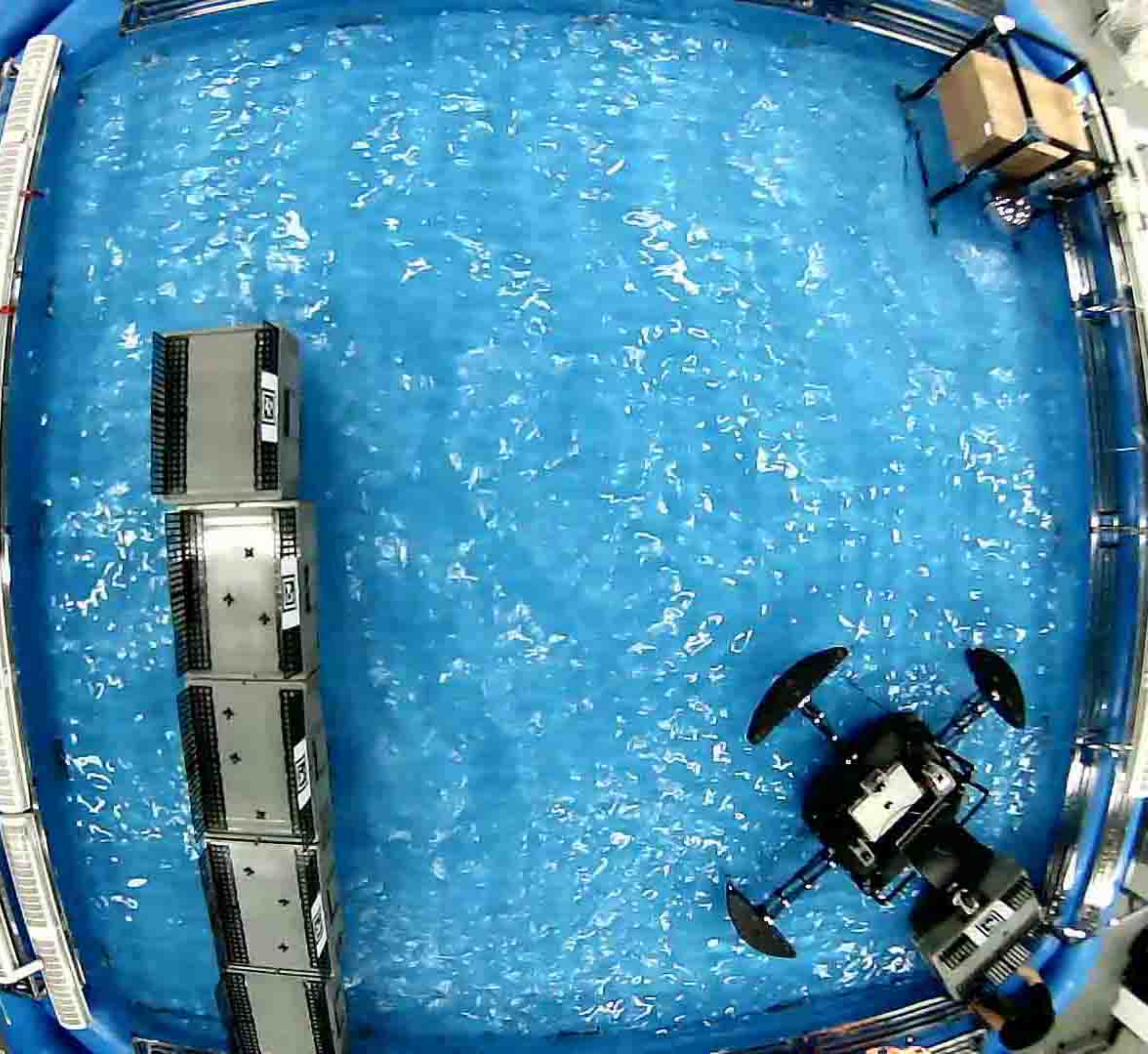}
			\put(-10,35){5}
		\end{overpic}
	}
	\hspace{-5pt}
	\subfloat[]{
		\centering
		\includegraphics[width=\figsize\textwidth]{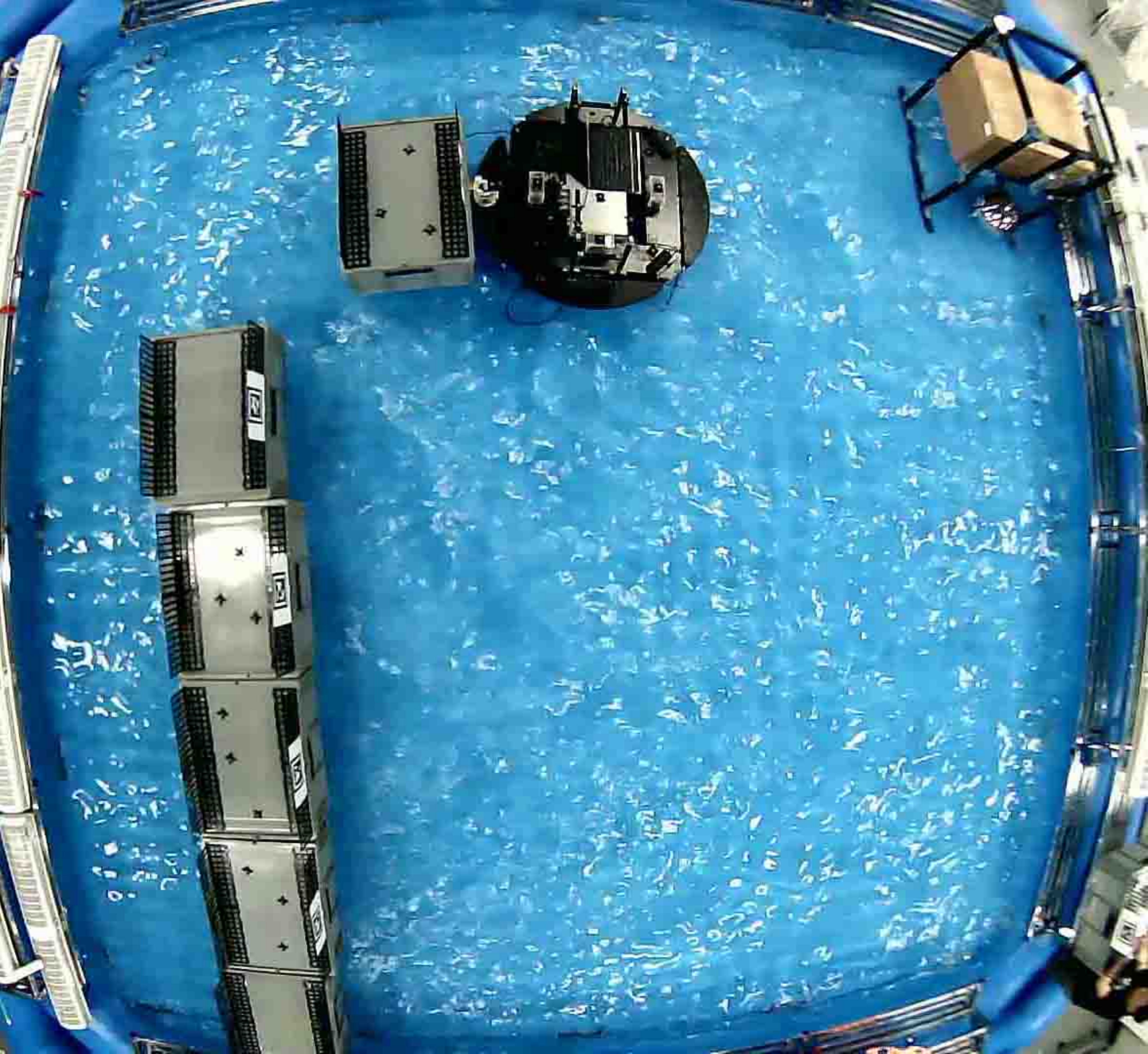}
	}
	\hspace{-5pt}
	\subfloat[]{
		\centering
		\includegraphics[width=\figsize\textwidth]{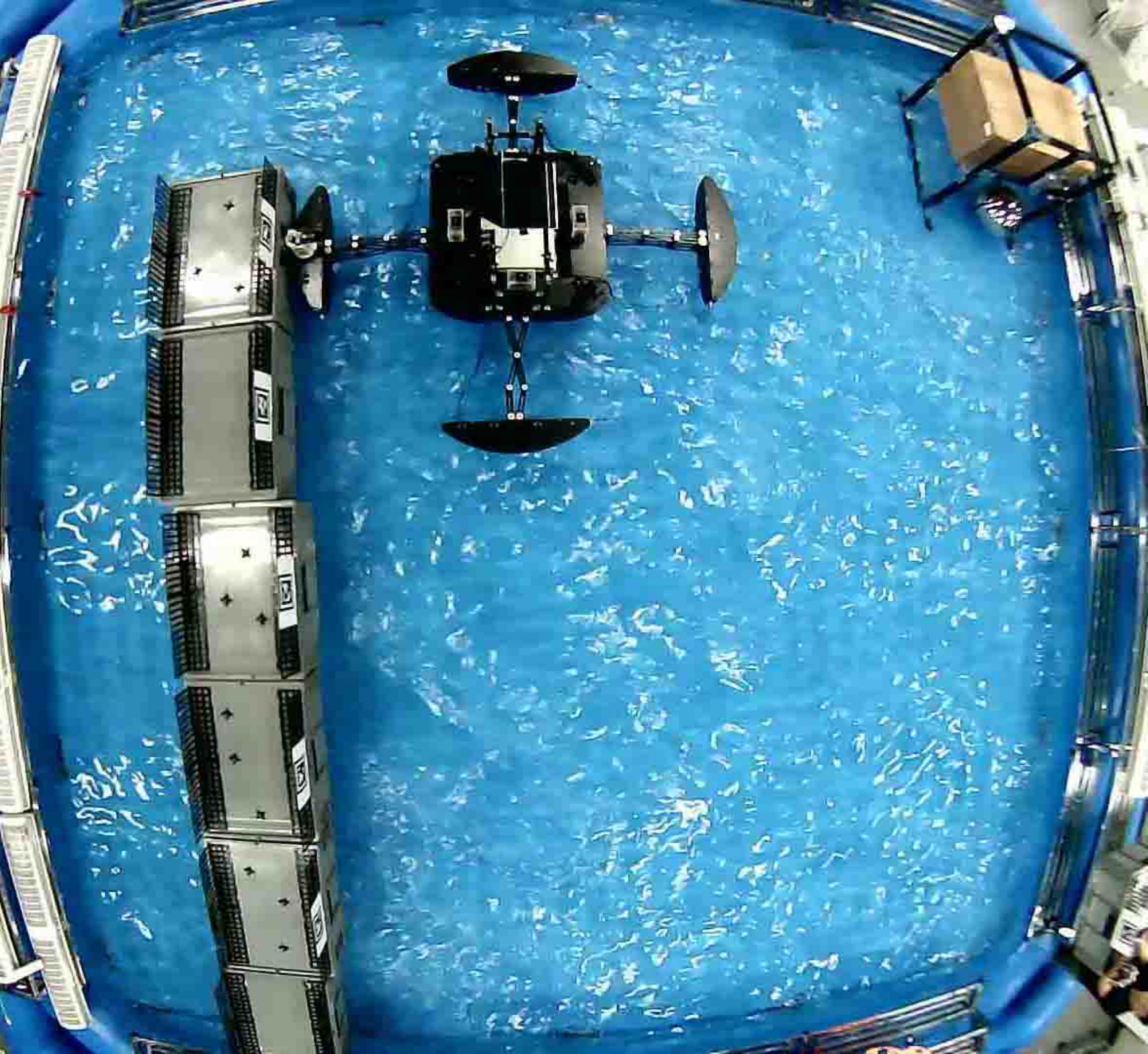}
	}
	%\hspace{0.5mm}
	\subfloat[]{
		\centering
		\begin{overpic}[width=\figsize\textwidth]{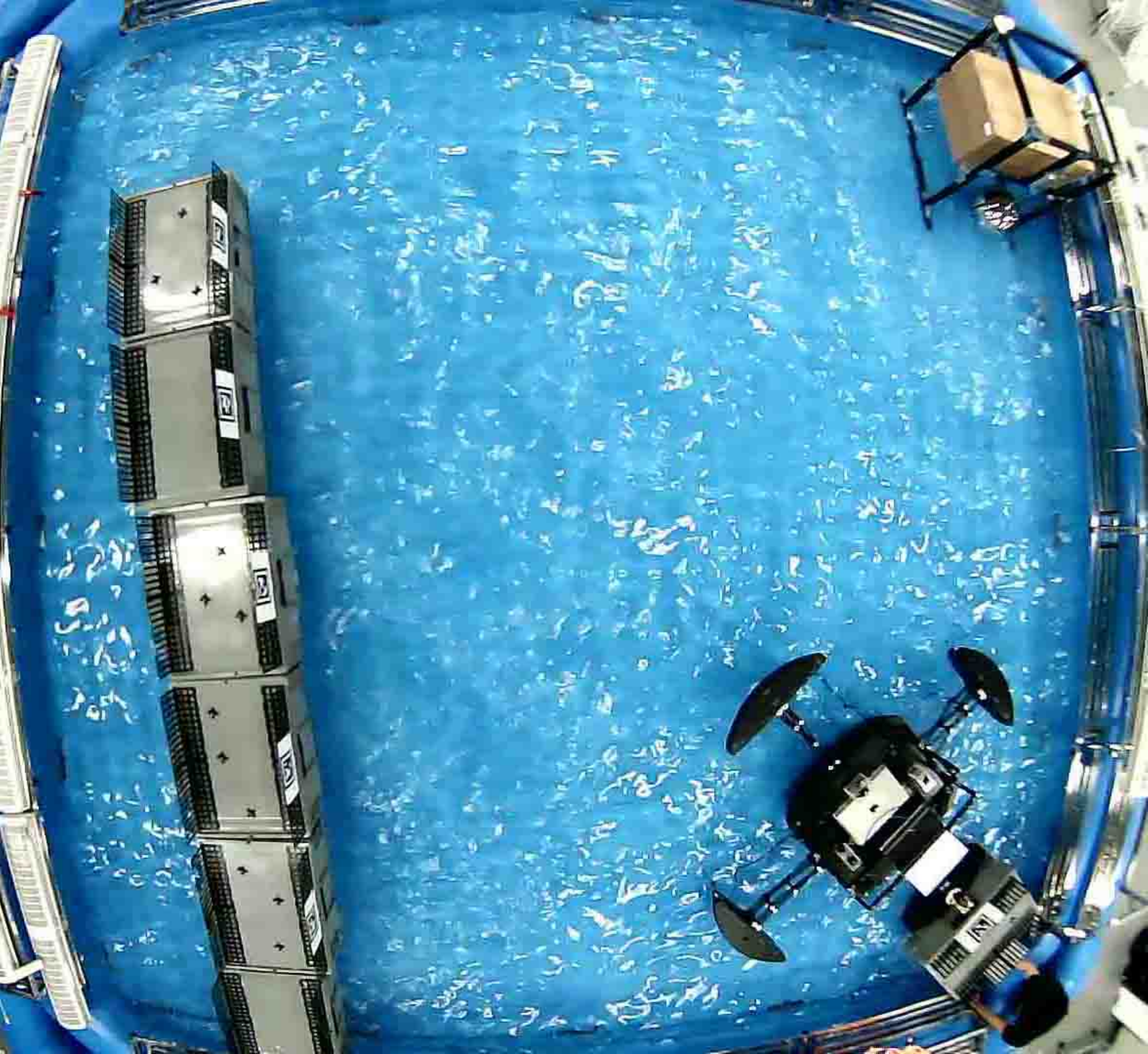}
			\put(-10,35){6}
		\end{overpic}
	}
	\hspace{-5pt}
	\subfloat[]{
		\centering
		\includegraphics[width=\figsize\textwidth]{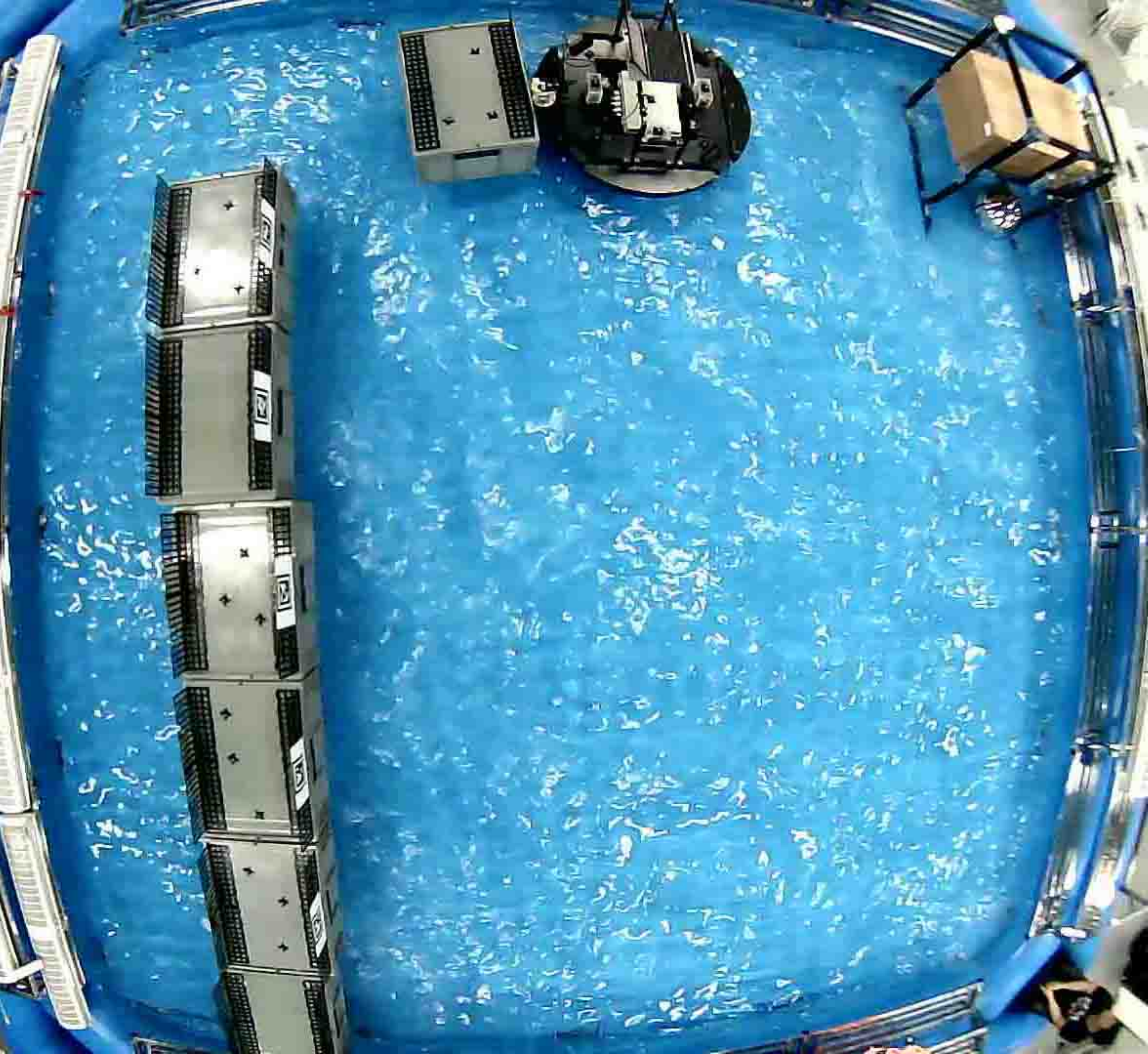}
	}
	\hspace{-5pt}
	\subfloat[]{
		\centering
		\includegraphics[width=\figsize\textwidth]{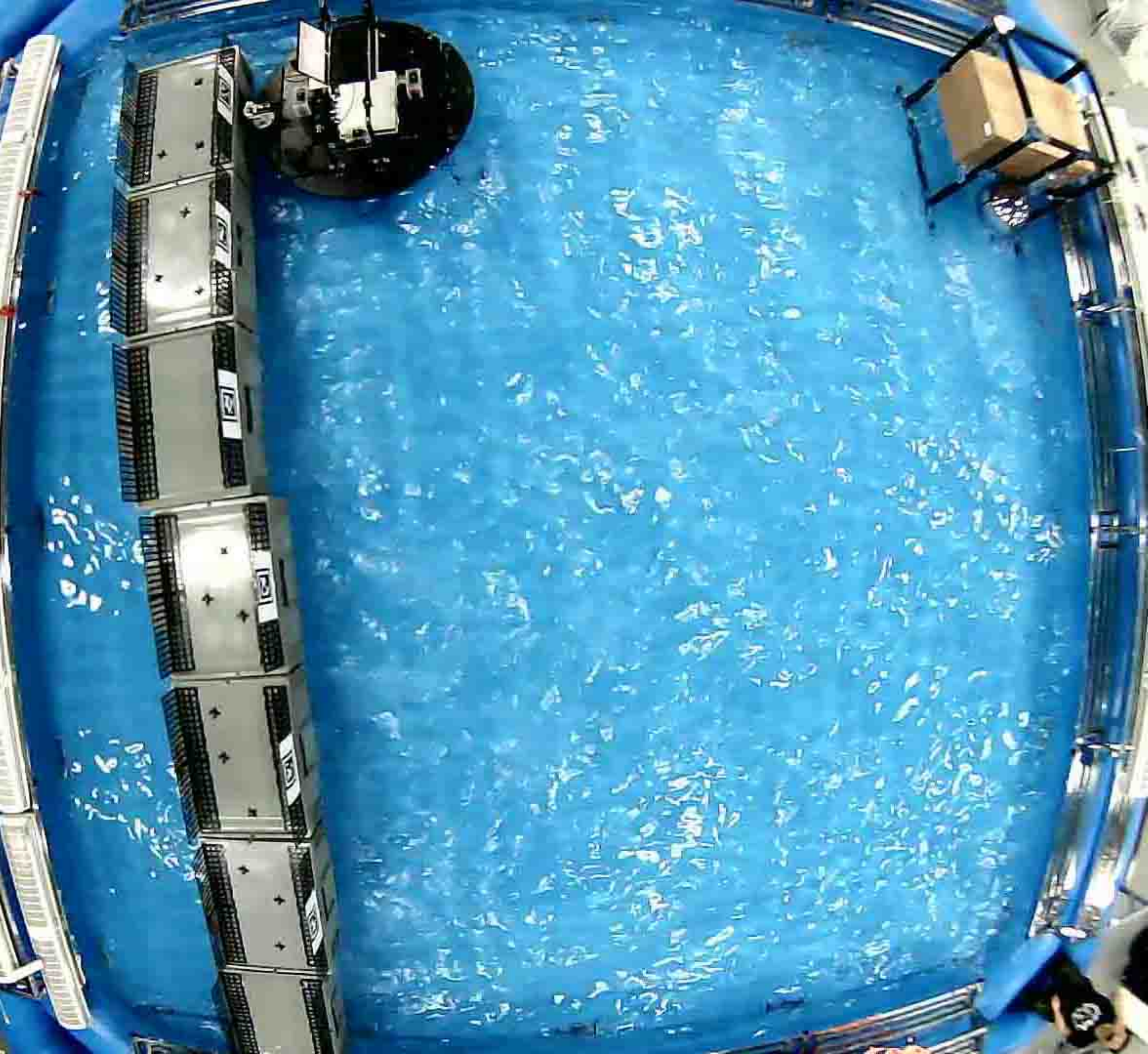}
	}
	\newline
	\caption{Bridge construction process. Adding each module requires three steps: pickup, delivery, and assembly.}
	\label{fig:bridging}
\end{figure*}

\subsection{Docking}

To compare the stability and docking performance of the TransBoat in different forms, docking experiments are performed in which the USV repeatedly docks to an identical docking system located beside the pool. 
A wave generating machine is positioned in the pool to simulate natural waves so that a wavefield can be created with a frequency of approximately $1.5\ \rm Hz$ and an average height of $6.7\ \rm cm$, as shown in Fig. \ref{fig:wave}.
In each test, controlled by the NMPC controller, the TransBoat in contracted or expanded form moves along a preplanned path to perform docking in both calm and turbulent water. 
As shown in Fig. \ref{fig:docking_success_rate}, the preplanned path starts at point $R_0$ and ends at a position $R_2$ where the robot executes the docking actions.
Before reaching the target, to ensure successful docking, the TransBoat first arrives at the preparation position $R_1$, which is $0.4\ \rm m$ away from the target.  
Each test is repeated 20 times.

Fig. \ref{fig:docking_success_rate} shows the experimental results, including the path, success rate, and average consumed time counted from the start time at $R_0$ to the time of successful docking.
From the success rate, we can see that the expanded form surpasses the contracted form in both calm ($95\%\ \text{vs}\ 90\%$) and turbulent water ($90\%\ \text{vs}\ 80 \%$), and that wave disturbance hampers the docking actions.
A comparison of the docking times reveals that in calm water both forms dock in a similarly short time, while in turbulent water the expanded form requires less time ($95.7\ \text{s}$ on average) than the contracted form ($135\ \text{s}$ on average).
Therefore, it is preferable to use the expanded form to execute docking actions in turbulent environments due to the higher success rate and shorter docking time.

Figs. \ref{fig:docking_pos} and \ref{fig:docking_angle} depict the positions and orientations, respectively, during the docking processes.
Each plot exhibits three successful docking sequences among $20$ repetitions of each test.
Particularly in Fig. \ref{fig:docking_angle}, from the roll angles of (b) and (e), 
it is noticeable that the oscillation is mitigated by $24.2\%$ and $32.6\%$ in calm and turbulent water, respectively.
In (c) and (f), the pitch oscillation reduction is $51.5\%$ and $36.1\%$. 
This confirms the stabilizing effect of the expanded form. 
Notice that the yaw angles of trials in calm water in Fig. \ref{fig:docking_angle} (a) seem not stable as expected, although their errors are inferior to those in (d). Several reasons have led to this phenomenon.
First, the average USV velocities in calm water are higher than those in turbulent water, as presented in Fig. \ref{fig:docking_success_rate}, which may worsen the yaw tracking angles.
Second, the poor tracking performance of yaw motion mainly happened during the movement when the orientation control is not our major concern.
Third, at the attraction and separation of the docking action, the yaw angle was disturbed by the attraction force of the magnetic docking mechanism.

\subsection{Bridge Construction}

\begin{algorithm}
	\caption{Bridge construction algorithm}
	\label{alg:Bridging}
	\LinesNumbered 
	\KwIn{pickup location, target locations}
	\KwOut{building a bridge}
	Initialize the TransBoat to contracted form \\
	\While{bridge not finished}{
		\While{no successful pickup}{
			move to the pickup preparation location\\
			expand \& turn on the docking system \\
			move to the pickup position for docking \\
		}
		return to the pickup preparation position \\
		spin to face the bridge \\
		deliver to the assembly preparation position \\
		\If{not last block}{
			expand \\
		}
		move to the assembly position for construction \\
		turn off the docking system \\
		return to the pre-assembly position \\
	}
\end{algorithm}

Based on the NMPC method and experimental summaries, the TransBoat is used to transport six building blocks to construct a temporary floating bridge in this experiment. 
The building scheme and the overall process are shown in Algorithm \ref{alg:Bridging} and Fig.\ref{fig:bridging}, respectively.
The building blocks are all modified by the installation of ferromagnets and permanent magnets on the sides so that they can be captured by the TransBoat and firmly connected.
The permanent magnet setup ensures that once the blocks are near each other, they instantly become attached.

The construction process for each building block contains three steps, namely, pickup, delivery, and assembly. 
In the pickup step, the TransBoat moves to the pickup location, which is invariant in every round, and docks with the purpose-made block.
Then, the TransBoat rapidly delivers the blocks to an assembly preparation location.
Finally, the block is slowly placed at the construction site and steadily assembled to the bridge.
In the pickup and assembly steps, the TransBoat should stabilize itself for precise motion (in expanded form), while in the delivery step, it should move fast (in contracted form).
Through these steps, the TransBoat is capable of fetching the building blocks one by one and constructing the bridge.

\section{Conclusion}
\label{sect:conclusion}
Overwater construction by USVs faces challenges from many environmental disturbances, such as unpredictable waves.   
This paper proposes a transformable omnidirectional USV named the TransBoat and a real-time NMPC algorithm adapted to all TransBoat forms for construction in turbulent water.
The TransBoat is capable of maneuvering in complicated environments with four thrusters and stabilizing in waves by expanding its outrigger hulls.
An instant docking system with a switchable magnet enables the TransBoat to dock with building modules and build a temporary bridge or a floating platform. 
The tracking and docking experiments, respectively, verify the high motion accuracy and docking success rate in the expanded form.
Finally, an automatic bridge construction experiment demonstrates the application potential of the TransBoat.

Our future research will focus on the following aspects. 
First, as the TransBoat can expand its four outrigger hulls independently to transform into asymmetrical forms, our next research subjects will be the relevant modeling and control of this feature. 
Second, future work will address the challenges of planning and collision avoidance required during the construction of sophisticated structures.

%\appendices

%\section*{Acknowledgment}

%\ifCLASSOPTIONcaptionsoff
%  \newpage
%\fi



\bibliographystyle{IEEEtran}
\bibliography{USV.bib}

%\bibliographystyle{IEEEtran}
%\begin{thebibliography}{1}
%\end{thebibliography}


\begin{IEEEbiography}[{\includegraphics[width=1in,height=1.25in,clip,keepaspectratio]{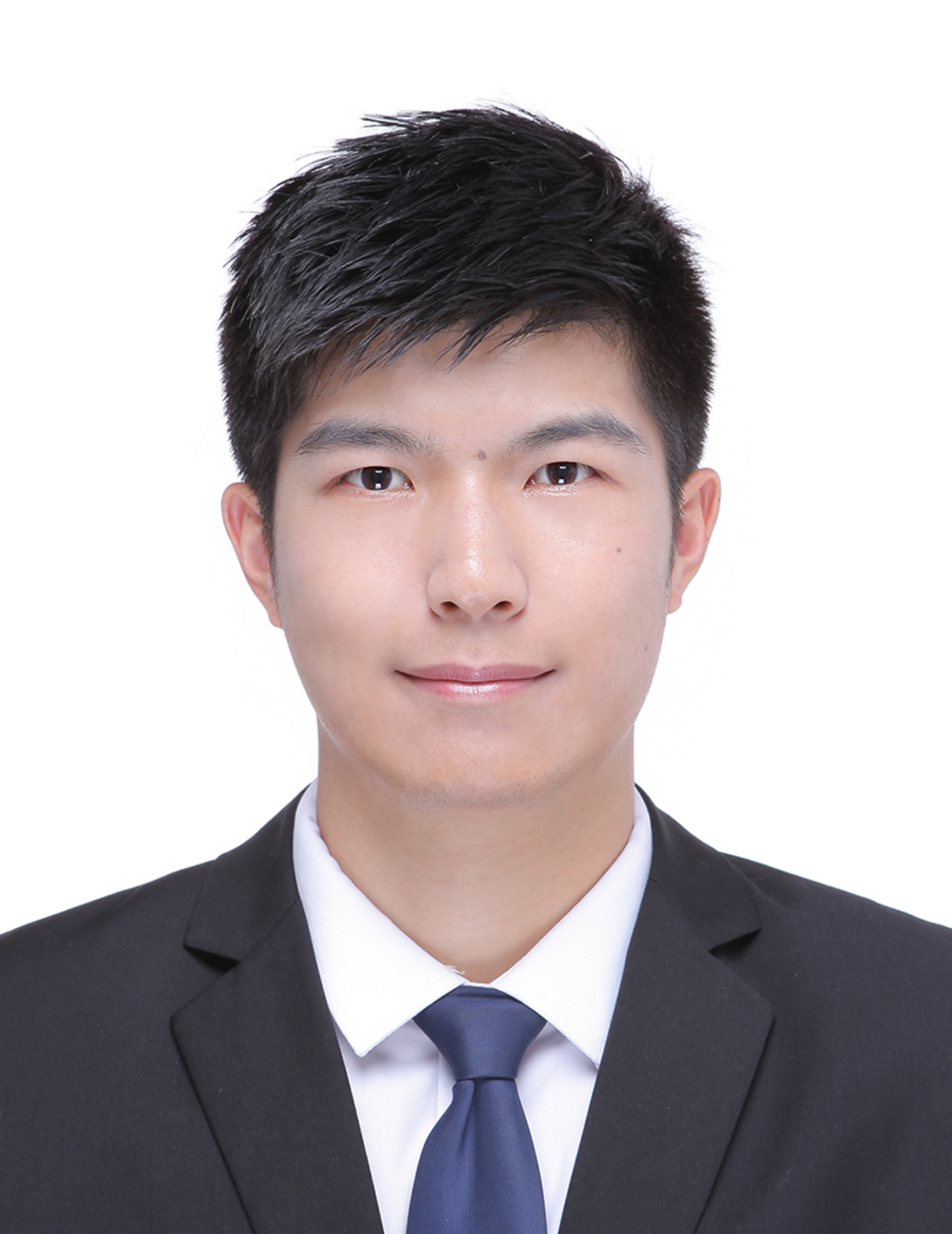}}]{Lianxin Zhang}
	received the B.E. degree in mechanical engineering and automation from Nanjing University of Science and Technology, Nanjing, China, in 2015, and the M.S. degree in mechanics from Tongji University, Shanghai, China, in 2018. He is currently pursuing the Ph.D. degree at The Chinese University of Hong Kong, Shenzhen, Guangdong, China, where he specializes in the design and control of novel unmanned surface vehicles.
\end{IEEEbiography}
\vspace{-50 mm}
\begin{IEEEbiography}[{\includegraphics[width=1in,height=1.25in,clip,keepaspectratio]{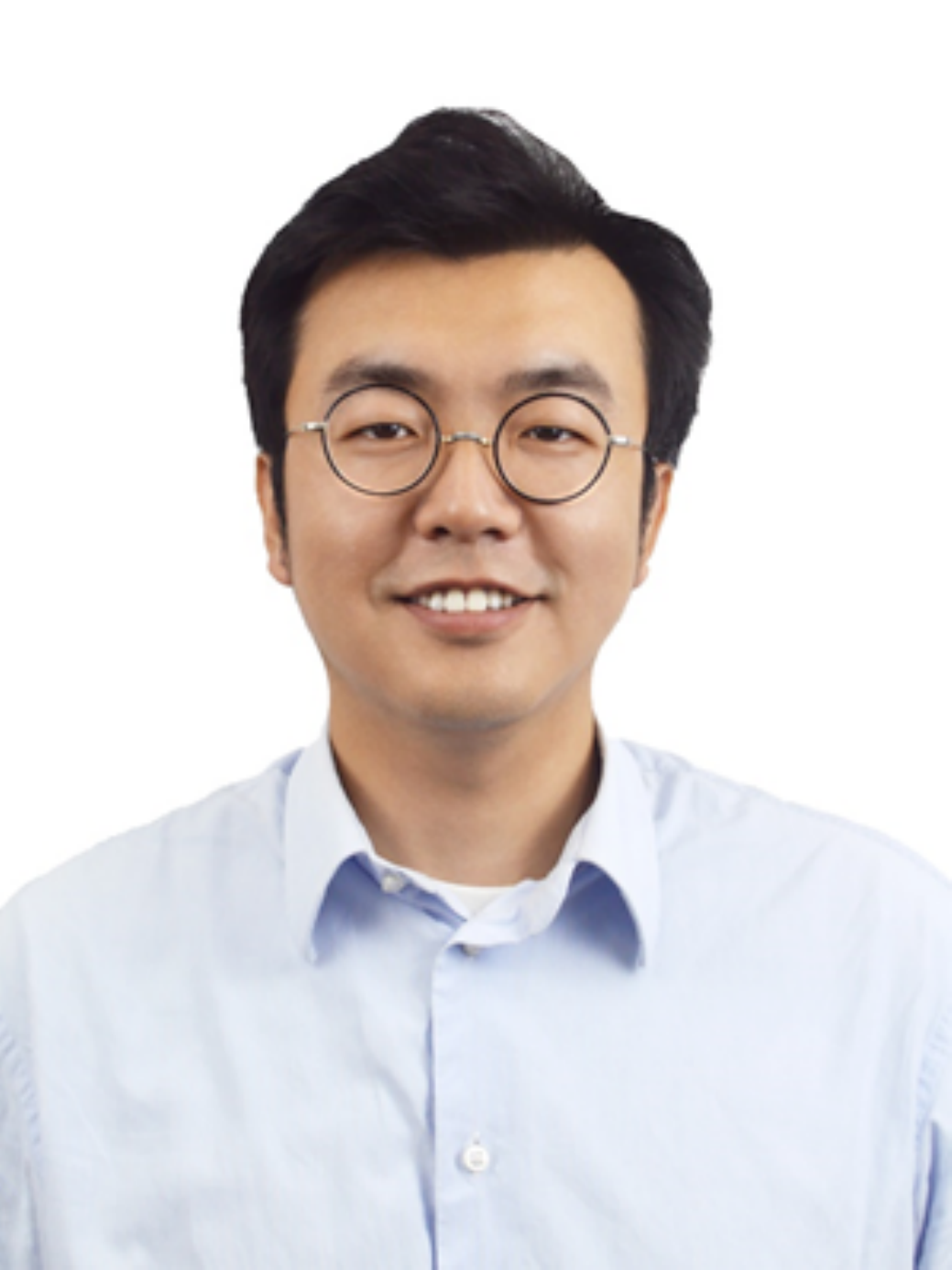}}]{Xiaoqiang Ji}
	received his Ph.D. degree from Columbia University, New York, USA, in 2019 with the department of mechanical engineering. He is currently a research assistant professor in School of Science and Engineering, The Chinese University of Hong Kong, Shenzhen. His research interests include control and stability of intelligent and complex systems, robotic systems, etc.
\end{IEEEbiography}
\vspace{-50 mm} 
\begin{IEEEbiography}[{\includegraphics[width=1in,height=1.25in,clip,keepaspectratio]{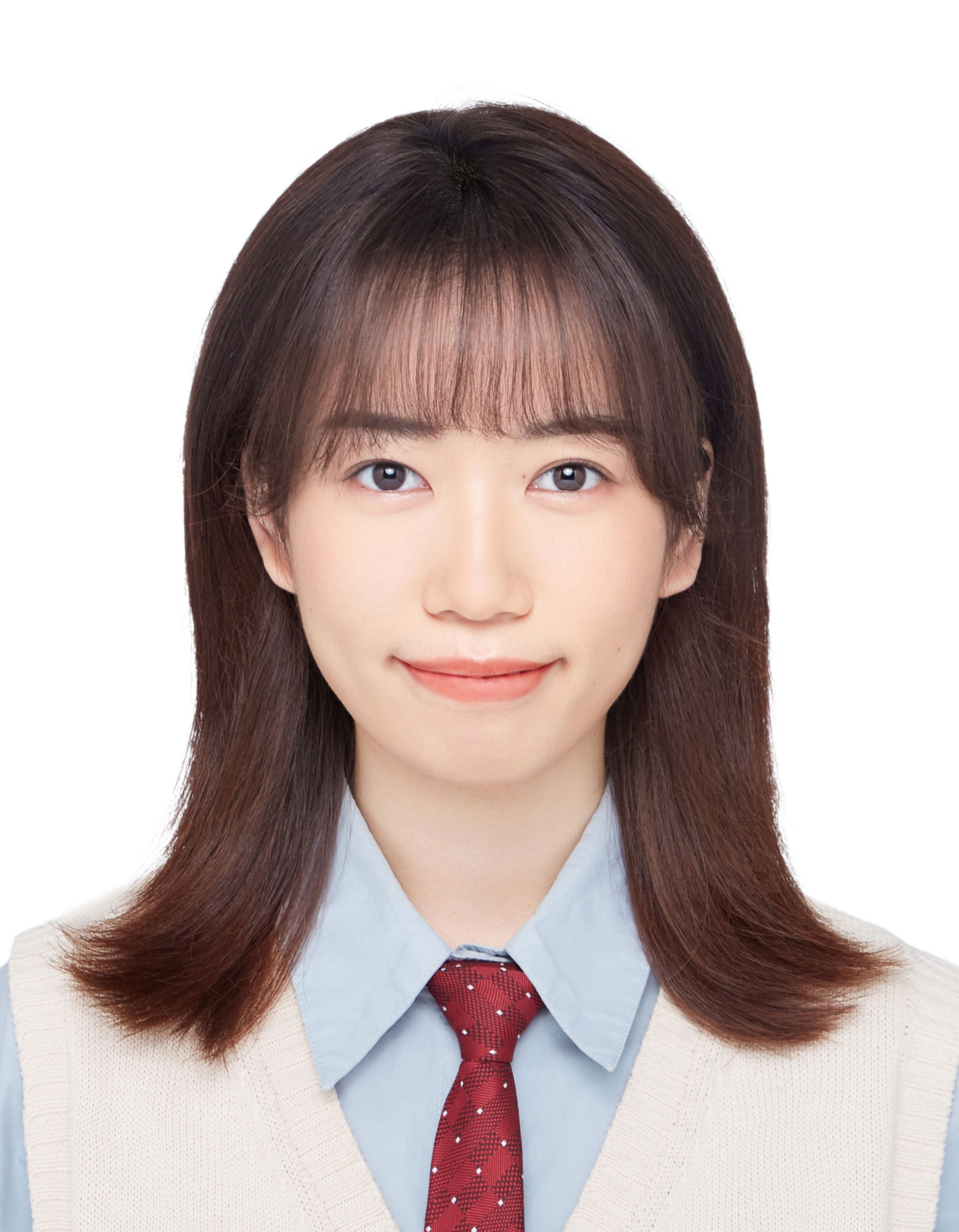}}]{Yang Jiao}
	received the B.E. degree in electronic information engineering from The Chinese University of Hong Kong, Shenzhen, Guangdong, China, in 2022. 
	She is currently pursuing the M.S. degree in intelligent systems, robotics \& control with the Department of Electrical and Computer Engineering, University of California San Diego, San Diego, CA, USA.
\end{IEEEbiography}
\vspace{-50 mm} 
\begin{IEEEbiography}[{\includegraphics[width=1in,height=1.25in,clip,keepaspectratio]{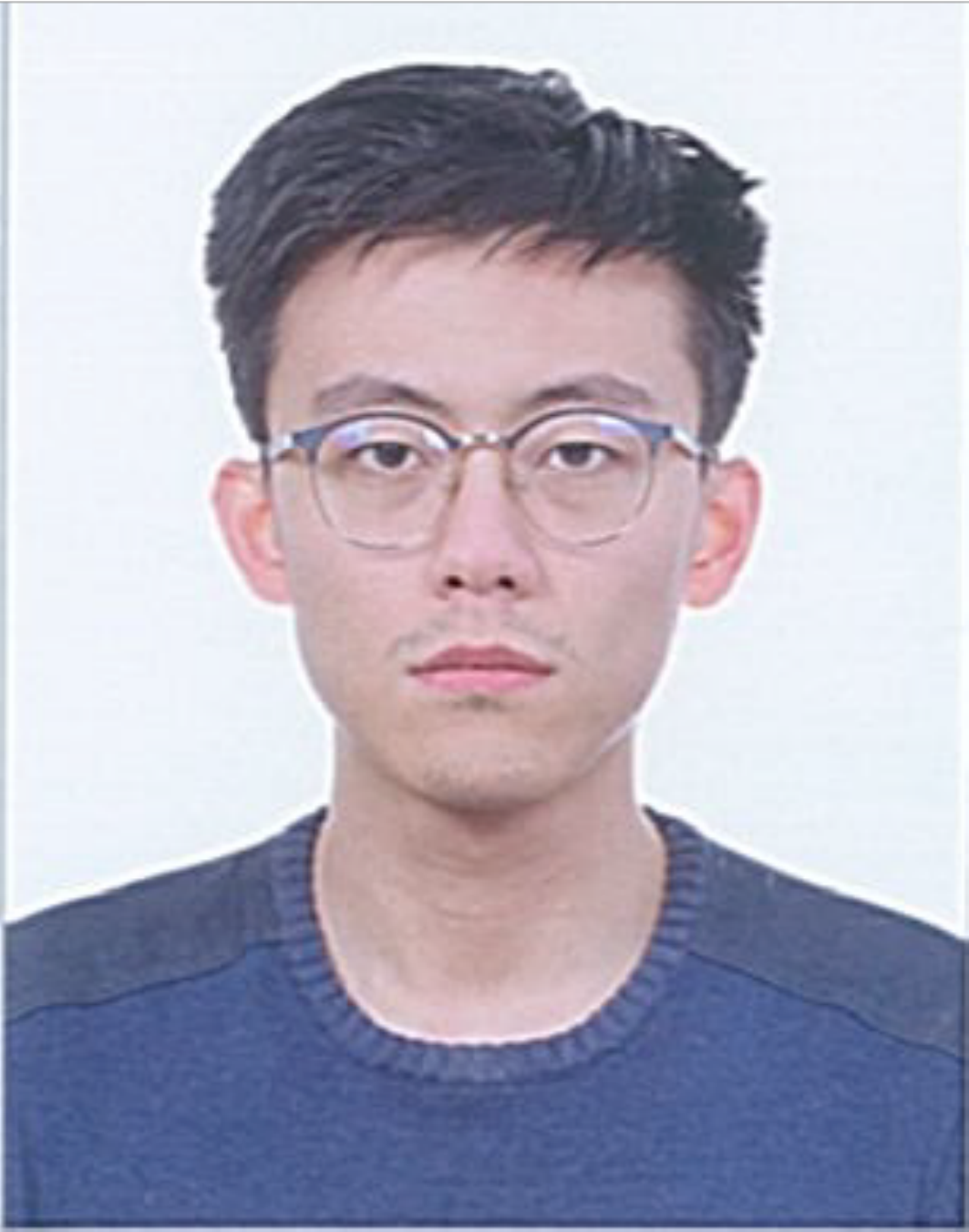}}]{Yihan Huang}
	received the B.E. degree in electronic information engineering from the Chinese University of Hong Kong, Shenzhen, China, in 2021. 
	He is now working in the Robotics and Artificial Intelligence Laboratory of the Chinese University of Hong Kong, Shenzhen, China. 
\end{IEEEbiography}
\vspace{-50 mm} 
\begin{IEEEbiography}[{\includegraphics[width=1in,height=1.25in,clip,keepaspectratio]{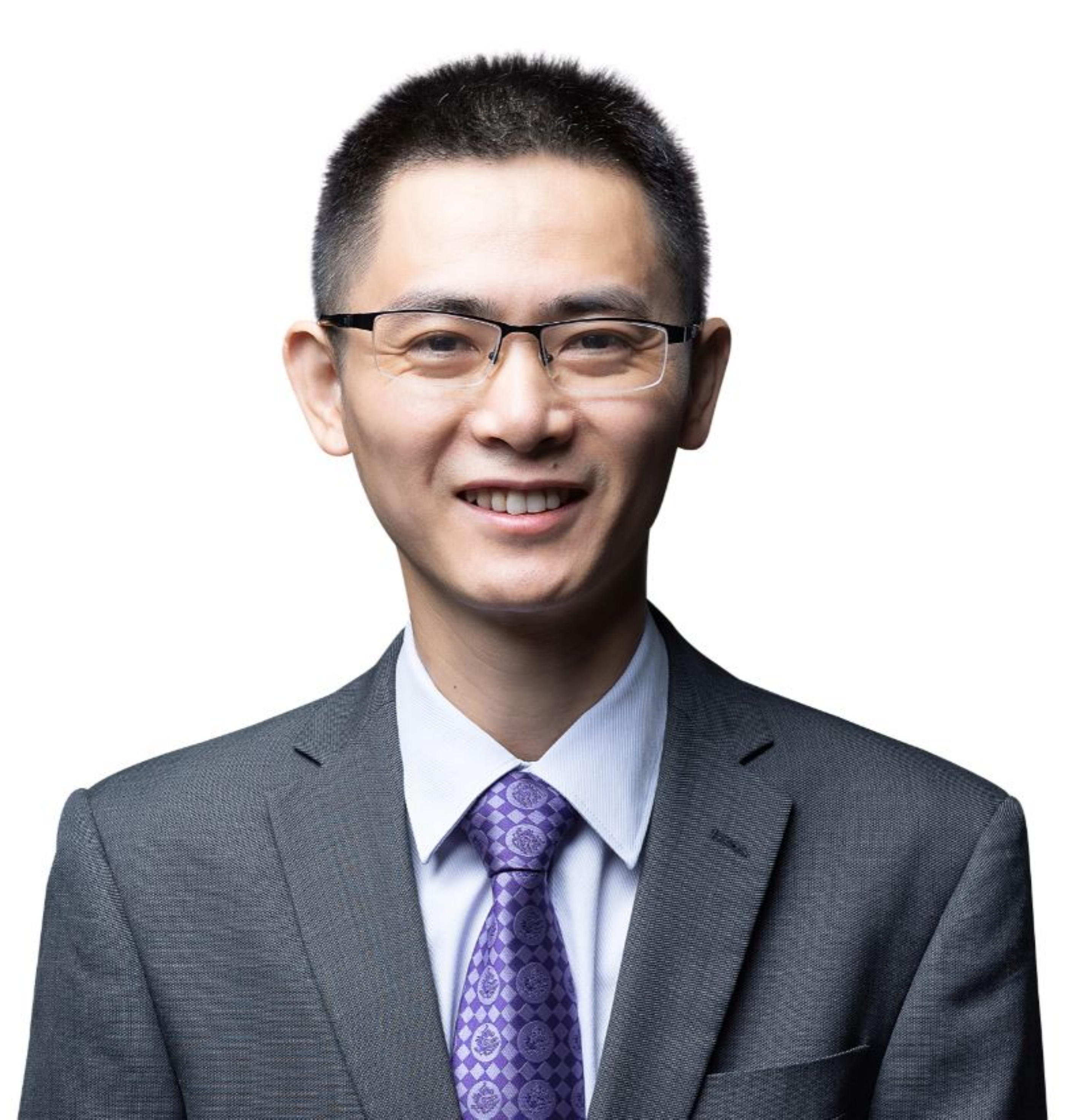}}]{Huihuan Qian}
	(Member, IEEE) received the B.E. degree from the Department of Automation, University of Science and Technology of China, Hefei, China, in 2004, and the Ph.D. degree from the Department of Mechanical and Automation Engineering, The Chinese University of Hong Kong, Hong Kong, SAR, China, in 2010. He is currently an Assistant Professor with the School of Science and Engineering, The Chinese University of Hong Kong, Shenzhen, Guangdong, China, and the Associate Director of the Shenzhen Institute of Artificial Intelligence and Robotics for Society, Shenzhen. His current research interests include robotics and intelligent systems, especially in marine environment.
\end{IEEEbiography}

\end{document}